\pdfoutput=1
\documentclass{article}

\PassOptionsToPackage{numbers, compress}{natbib}
\PassOptionsToPackage{dvipsnames}{xcolor}
\usepackage{iclr2023_conference,times}
\iclrfinalcopy

\usepackage[utf8]{inputenc} % allow utf-8 input
\usepackage[T1]{fontenc}    % use 8-bit T1 fonts
\usepackage{hyperref}
\usepackage{url}            % simple URL typesetting
\usepackage{booktabs}       % professional-quality tables
\usepackage{amsfonts}       % blackboard math symbols
\usepackage{nicefrac}       % compact symbols for 1/2, etc.
\usepackage{microtype}      % microtypography
\usepackage{xcolor}         % colors

\usepackage{caption}
\usepackage{subcaption}
\usepackage{multirow} % For hyperparameter table
\usepackage{tikz}
\usepackage{pgfplots}
\DeclareUnicodeCharacter{2212}{−}
\usepgfplotslibrary{groupplots,dateplot}
\usetikzlibrary{patterns,shapes.arrows}
\pgfplotsset{compat=newest}
\usetikzlibrary{shapes,decorations,arrows,calc,arrows.meta,fit,positioning}
\usetikzlibrary{positioning,calc}
\usetikzlibrary{backgrounds,fit}
\usetikzlibrary{arrows.meta}
\usetikzlibrary{arrows}

\pgfplotsset{
    compat=1.15,
    mystylepalette/.style={
        mystyle/.style={mark=*, mark size=2, mark options={solid, draw=white, line width=.5pt}, forget plot},
    label style={font=\footnotesize}, tick label style={font=\footnotesize},
    },
}

\usepackage{verbatim}
\usepackage{svg}
\usepackage{listings}
\usepackage{amsmath,amssymb}
\DeclareMathOperator*{\E}{\mathbb{E}}
\usepackage[ruled,vlined]{algorithm2e}
\usepackage[shortlabels]{enumitem}

\usepackage{multirow}

\usepackage{todonotes}

\newcommand{\slimparagraph}[1]{\textbf{#1~~~}}

\title{TabPFN: A Transformer That Solves Small\\Tabular Classification Problems in a Second}

\makeatletter
\newcommand{\printfnsymbol}[1]{%
  \textsuperscript{\@fnsymbol{#1}}%
}
\makeatother

\author{Noah Hollmann$^{*, 1, 2}$~\; Samuel M\"uller$^{*, 1}$~\; Katharina Eggensperger$^{1}$~\; Frank Hutter$^{1, 3}$ \\
$^{1}$ University of Freiburg, $^{2}$ Charit\'e University Medicine Berlin \\ $^{3}$ Bosch Center for Artificial Intelligence
\printfnsymbol{1} Equal contribution. \\ Correspondence to \texttt{noah.hollmann@charite.de} \& \texttt{muellesa@cs.uni-freiburg.de}
}

\begin{document}
%\nolinenumbers
\maketitle

\newcommand{\methodname}[0]{TabPFN}
\newcommand{\priortuning}[0]{differentiable hyperparameter tuning}
\newcommand{\openmlcc}[0]{OpenML-CC18}
\newcommand{\openmlautoml}[0]{OpenML-AutoML}
\begin{abstract}
We present \methodname{}, a trained Transformer that can do supervised classification for small tabular datasets in \textit{less than a second}, needs no hyperparameter tuning and is competitive with state-of-the-art classification methods. \methodname{} performs in-context learning (ICL), it learns to make predictions using sequences of labeled examples (x, f(x)) given in the input, without requiring further parameter updates.
\methodname{} is fully entailed in the weights of our network, which accepts training and test samples as a set-valued input and yields predictions for the entire test set in a single forward pass.
\methodname{} is a Prior-Data Fitted Network (PFN) and is trained offline once, to approximate Bayesian inference on synthetic datasets drawn from our prior.
This prior incorporates ideas from causal reasoning: It entails a large space of structural causal models with a preference for simple structures.
On the $18$ datasets in the \openmlcc{} suite that contain up to 1\,000 training data points, up to 100 purely numerical features without missing values, and up to 10 classes, we show that our method clearly outperforms boosted trees and performs on par with complex state-of-the-art AutoML systems with up to $230\times$ speedup.
This increases to a $5\,700\times$ speedup when using a GPU. We also validate these results on an additional 67 small numerical datasets from OpenML.
We provide all our code, the trained \methodname{}, an interactive browser demo and a Colab notebook at
% Anonymize
%\url{https://github.com/tabpfn-anonym/TabPFNAnonym}.
\url{https://github.com/automl/TabPFN}.
\end{abstract}

% -------------------------------------
\section{Introduction}
\label{sec:introduction}
% -------------------------------------

Tabular data has long been overlooked by deep learning research, despite being the most common data type in real-world machine learning (ML) applications \citep{chui-mckinsey18a}.  
While deep learning methods excel on many ML applications, tabular data classification problems are still dominated by Gradient-Boosted Decision Trees (GBDT;~\citealp{friedman-as01a}), largely due to their short training time and robustness \citep{nnsfortabular2}.

We propose a radical change to how tabular classification is done. We do \emph{not} fit a new model from scratch to the training portion of a new dataset. Instead, we replace this step by performing a single forward pass with a large Transformer that has been pre-trained to solve artificially generated classification tasks from a tabular dataset prior.

Our method builds on Prior-Data Fitted Networks (PFNs; \citealp{muller-iclr22a}; see Section~\ref{sec:prior_fitted_intro}), which learn the training and prediction algorithm itself. PFNs approximate Bayesian inference given any prior one can sample from and approximate the posterior predictive distribution (PPD) directly. 
While inductive biases in NNs and GBDTs depend on them being efficient to implement (e.g., through $L_2$ regularization, dropout \citep{srivastava-jmlr14a} or limited tree-depth), in PFNs, one can simply design a dataset-generating algorithm that encodes the desired prior. This fundamentally changes the way we can design learning algorithms.

We design a prior (see Section~\ref{sec:tabprior}) based on Bayesian Neural Networks (BNNs; ~\citealt{neal-bayes96a,gal-phdthesis16a}) and Structural Causal Models (SCMs;~\citealt{pearl_2009,causalinferencelearningbook}) to model complex feature dependencies and potential causal mechanisms underlying tabular data. Our prior also takes ideas from Occam\textquotesingle{s} razor: simpler SCMs and BNNs (with fewer parameters) have a higher likelihood. Our prior is defined via parametric distributions, e.g., a log-scaled uniform distribution for the average number of nodes in data-generating SCMs. The resulting PPD implicitly models uncertainty over all possible data-generating mechanisms, weighting them by their likelihood given the data and their prior probability. Thus, the PPD corresponds to an infinitely large ensemble of data-generating mechanisms, i.e., instantiations of SCMs and BNNs. We learn to approximate this complex PPD in a single forward-pass, requiring no cross-validation or model selection.

Our \textbf{key contribution} is to introduce the \emph{\methodname{}} (see Section \ref{sec:tabpfn}), a single Transformer that has been pre-trained to approximate probabilistic inference for the novel prior above (described in more detail in Section \ref{sec:tabprior}) in a single forward pass, and has thus learned to solve novel small tabular classification tasks ($\le$ 1\,000 training examples, $\le$ 100 purely numerical features without missing values and $\le$ 10 classes) in \emph{less than a second} yielding state-of-the-art performance.

To substantiate this claim, we qualitatively and quantitatively analyze the behavior and performance of our \methodname{} on different tasks and compare it to previous approaches for tabular classification on $18$ small, numerical datasets (see Section \ref{sec:experiments}). 
Quantitatively, the \methodname{} yields much better performance than any individual ``base-level'' classification algorithm, such as gradient-boosting via XGBoost~\citep{chen-kdd16a}, LightGBM~\citep{ke-neurips17a} and CatBoost~\citep{prokhorenkova-neurips18a}, and in less than a second yields performance competitive to what the best available AutoML frameworks~\citep{erickson-arxiv20a,feurer-arxiv20v2} achieve in one hour.
Our in-depth qualitative analysis shows the \methodname{}\textquotesingle{s} predictions to be smooth and intuitive. Yet, its errors are quite uncorrelated to the errors of existing approaches, 
allowing additional performance improvements by ensembling. We also validate \methodname{}'s performance on an additional 67 datasets from OpenML~\cite{vanschoren-sigkdd14a}.

We expect the revolutionary character of our claims to be met with initial skepticism and thus open-source all our code and the pre-trained \methodname{} for scrutinization by the community, along with a scikit-learn-like interface, a Colab notebook and two online demos. Please see Section \ref{sec:reproducibility} for the links.

% 30s per small dataset is usual in SOTA, we do 0.5s
% half a minute to half a second 

\section{Background on Prior-Data Fitted Networks (PFNs)}
\label{sec:prior_fitted_intro}

First, we summarize how PFNs work; we refer to~\cite{muller-iclr22a} for the full details.

\slimparagraph{The Posterior Predictive Distribution for Supervised Learning}
In the Bayesian framework for supervised learning, the prior defines a space of hypotheses $\Phi$ on the relationship of a set of inputs $x$ to the output labels $y$.
Each hypothesis $\phi\in\Phi$ can be seen as a mechanism that generates a data distribution from which we can draw samples forming a dataset. For example, given a prior based on structural causal models, $\Phi$ is the space of structural causal models, a hypothesis $\phi$ is one specific SCM, and a dataset comprises samples generated through this SCM. In practice, a dataset comprises training data with observed labels and test data where labels are missing or held out to assess predictive performance.
The PPD for a test sample $x_{test}$ specifies the distribution of its label $p(\cdot|x_{test},D_{train})$, which is conditioned on the set of training samples $D_{train} := \{(x_1,y_1), \dots, (x_n,y_n)\}$.
The PPD can be obtained by integration over the space of hypotheses $\Phi$, where the weight of a hypothesis $\phi\in\Phi$ is determined by its prior probability $p(\phi)$ and the likelihood $p(D|\phi)$ of the data $D$ given $\phi$:
\begin{equation}
    p(y|x,D) \propto \int_{\Phi} p(y|x,\phi) p(D|\phi) p(\phi) d\phi.
    \label{eq:ppd}
\end{equation}
%While obtaining the PPD is generally intractable, PFNs approximate the PPD, given any prior one can sample from.
%A PFN that minimizes its training objective to the global minimum, provably approximates exactly the PPD, as shown in \cite{muller-iclr22a}.

\slimparagraph{Synthetic Prior-fitting} 
Prior-fitting is the training of a PFN to approximate the PPD and thus do Bayesian prediction.
We implement it with a prior which is specified by a prior sampling scheme of the form $p(D)=\E_{\phi \sim p(\phi)}[p(D|\phi)]$, which first samples hypotheses (generating mechanisms) with $\phi \sim p(\phi)$ and then synthetic datasets with $D \sim p(D|\phi)$.
%, with weights $\theta$,
We repeatedly sample such synthetic datasets $D:={(x_i, y_i)}_{i \in \{1,\dots,n\}}$ and optimize the PFN\textquotesingle{}s parameters $\theta$ to make predictions for $D_{test} \subset D$, conditioned on the rest of the dataset $D_{train} = D \setminus D_{test}$.
The loss of the PFN training thus is the cross-entropy on held-out examples of synthetic datasets.
For a single test point $\{(x_{test},y_{test})\} = D_{test}$, the training loss can be written as
\begin{equation}
    \mathcal{L}_{\textit{PFN}} = \E_{(\{(x_{test}, y_{test})\} \cup D_{train}) \sim p(D)} [-\log q_{\theta}(y_{test}|x_{test},D_{train})].
    \label{eq:lpfn}
\end{equation}
As shown by \cite{muller-iclr22a}, minimizing this loss approximates the true Bayesian posterior predictive distribution. We visualize this in Figure~\ref{fig:pfn_usage} and detail the full training setup in Algorithm \ref{alg:prior-fitting} in the appendix.
Crucially, this \textit{synthetic prior-fitting} phase is performed only once for a given prior $p(D)$ as part of algorithm development. 
%, learning to do \textit{real-world inference} on new datasets.

%
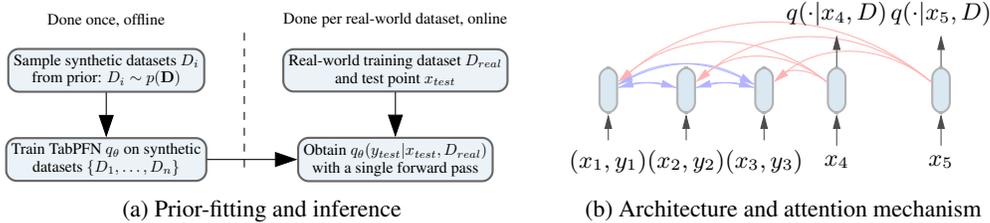
\begin{figure}[t]
     \centering
     \begin{subfigure}[t]{.49\textwidth}
         \centering
         \newcommand{\lastidx}{n+1}
\newcommand{\evalidx}{n+1}
\newcommand{\dataset}{\mathbf{D}}
\newcommand{\setofpairs}[2]{\{(x_1^{#1}, y_1^{#1}),\dots,(x_{#2}^{#1}, y_{#2}^{#1})\}}
\newcommand{\seq}[1]{$\setofpairs{#1}{\lastidx{}}$}
\begin{tikzpicture}[remember picture,
  inner/.style={circle,draw=blue!50,fill=blue!20,thick,inner sep=3pt},
  scale=0.6, every node/.style={transform shape},
]
\tikzstyle{bigbox} = [thick, fill=white, rounded corners, rectangle,draw=black!30]
\tikzstyle{sample} = [thick, minimum size=0.6cm, rounded corners,rectangle, fill=Salmon!10,draw=black!30]
\tikzstyle{loss} = [thick, minimum size=0.6cm, rounded corners,rectangle, fill=PineGreen!10,draw=black!30]
\tikzstyle{pfmbox} = [thick,minimum size=0.6cm, rounded corners,rectangle, fill=MidnightBlue!10,draw=black!30]
\tikzstyle{box} = [thick, minimum size=0.6cm, rounded corners,rectangle,fill=MidnightBlue!5,draw=black!60]

\node[] (desc_offline) {Done once, offline};
\node [below=1em of desc_offline, align=center, box] (sample_datasets) {Sample synthetic datasets $D_i$ \\ from prior: $D_i \sim p(\dataset{})$};
\node [below=of sample_datasets, align=center, box] (train) {Train TabPFN $q_\theta$ on synthetic \\ datasets $\{D_1,\dots,D_n\}$};
\draw [-{Triangle[length=3mm, width=2mm]}] (sample_datasets) -- (train);

\node[right=2em of train] (bottom_of_line) {};
\node [] (top_of_line) at (bottom_of_line|-desc_offline) {};
\draw [dashed] (bottom_of_line)++(0,-.5em) to (top_of_line)++(0,0em);

\node[right=3em of bottom_of_line, anchor=west, align=center, box] (predict) {Obtain $q_\theta(y_{test}|x_{test},D_{real})$\\with a single forward pass};
\node[] (desc_online) at (desc_offline-|predict) {Done per real-world dataset, online};
\draw [-{Triangle[length=3mm, width=2mm]}] (train) -- (predict);
\node[anchor=center, align=center, box] (evalinput) at (sample_datasets-|predict) {Real-world training dataset $D_{real}$\\and test point $x_{test}$};
\draw [-{Triangle[length=3mm, width=2mm]}] (evalinput) -- (predict);

%\node[pfmbox,align=center] (pfm) at (l|-evaldata) {$q_{\theta^*}(y_{\evalidx{}}^{}|x^{}_{\evalidx{}}, \setofpairs{}{n})$ \\ $\approx$ \\ $p(y_{\evalidx{}}^{}|x^{}_{\evalidx{}}, \setofpairs{}{n})$};

%\draw [-{Triangle[length=3mm, width=2mm]}] (evaldata) -- (pfm);
%\draw [-{Triangle[length=3mm, width=2mm]}] (l) -- (pfm) node[midway,right] {$\theta^*$};

\end{tikzpicture}
         \caption{Prior-fitting and inference}
         \label{fig:pfn_usage}
     \end{subfigure}
     \begin{subfigure}[t]{.49\textwidth}
         \centering
         \iffalse
%\documentclass{article}
\usepackage[dvipsnames]{xcolor}
\usepackage{tikz}
\usetikzlibrary{positioning,calc,backgrounds,fit,arrows.meta}
\usetikzlibrary{arrows}
\usetikzlibrary{decorations.pathreplacing,calligraphy}

\fi
%\begin{document}
\newcommand{\lastidx}{n+1}
\newcommand{\evalidx}{n+1}
\newcommand{\setofpairs}[2]{\{(x_1^{#1}, y_1^{#1}),\dots,(x_{#2}^{#1}, y_{#2}^{#1})\}}
\newcommand{\seq}[1]{$\setofpairs{#1}{\lastidx{}}$}
\newcommand{\xypair}[1]{$(x_{#1},y_{#1})$}
\newcommand{\qattbend}{40}
\def\innerdistance{2.2em}
\def\outdistane{1.2em}
\def\indistane{1.em}
\begin{tikzpicture}[remember picture,
  inner/.style={circle,draw=blue!50,fill=blue!20,thick,inner sep=3pt},
]
\tikzstyle{bigbox} = [thick, fill=white, rounded corners, rectangle,draw=black!30]
\tikzstyle{sample} = [thick, minimum size=0.6cm, rounded corners,rectangle, fill=Salmon!10,draw=black!30]
\tikzstyle{loss} = [thick, minimum size=0.6cm, rounded corners,rectangle, fill=PineGreen!10,draw=black!30]
\tikzstyle{timestep} = [thick,minimum width=0.2cm,minimum height=.6cm, rounded corners,rectangle, fill=MidnightBlue!10,draw=black!30]
\tikzstyle{querytimestep} = [thick,minimum width=0.2cm,minimum height=.6cm, rounded corners,rectangle, fill=MidnightBlue!10,draw=black!30]
\tikzstyle{innerattention} = [draw=blue!30,-{Triangle[length=2mm, width=1mm]},]
\tikzstyle{queryattention} = [draw=red!30,-{Triangle[length=2mm, width=1mm]},]
\tikzstyle{output} = [draw=black!70,-{Triangle[length=2mm, width=1mm]},]
\tikzstyle{input} = [draw=black!70,-{Triangle[length=2mm, width=1mm]},]
\tikzstyle{every node} = [font=\footnotesize,]

\node[timestep] (t1) {};
\node[timestep, right=\innerdistance of t1] (t2) {};
\node[timestep, right=\innerdistance of t2] (t3) {};

\draw [ innerattention] (t1) to [bend left=15] (t2);
\draw [ innerattention] (t1) to [bend left=30] (t3);
\draw [ innerattention] (t2) to [bend right=15] (t1);
\draw [ innerattention] (t2) to [bend left=15] (t3);
\draw [ innerattention] (t3) to [bend right=30] (t1);
\draw [ innerattention] (t3) to [bend right=15] (t2);

\node[below=\indistane of t1] (i1) {\xypair{1}};
\draw [ input] (i1) -- (t1);
\node[below=\indistane of t2] (i2) {\xypair{2}};
\draw [ input] (i2) -- (t2);
\node[below=\indistane of t3] (i3) {\xypair{3}};
\draw [ input] (i3) -- (t3);

%\draw [pen colour={black!70},
%    decorate, 
%    decoration = {calligraphic brace, mirror,
%        raise=.7em,
%        amplitude=5pt}] (i1.west) --  (i3.east)
%    node[pos=0.5,below=1.2em,black]{$D$};

\node[querytimestep, right=2em of t3] (q1) {};
\node[querytimestep, right=3.2em of q1] (q2) {};

\draw [ queryattention] (q1) to [bend right=40] (t1);
\draw [ queryattention] (q1) to [bend right=40] (t2);
\draw [ queryattention] (q1) to [bend right=40] (t3);

\draw [ queryattention] (q2) to [bend right=40] (t1);
\draw [ queryattention] (q2) to [bend right=40] (t2);
\draw [ queryattention] (q2) to [bend right=40] (t3);
\node[above=\outdistane of q2] (d2) {$q(\cdot|x_5,D)$};

\node[above=\outdistane of q1] (d1) {$q(\cdot|x_4,D)$};
\draw [ output] (q1) -- (d1);
\node[] at (q1|-i1) (i4) {$x_4$};
\draw [ input] (i4) -- (q1);
\draw [ output] (q2) -- (d2);
\node[] at (q2|-i1) (i5) {$x_5$};
\draw [ input] (i5) -- (q2);
\end{tikzpicture}
%\end{document}
         \caption{Architecture and attention mechanism}
         \label{fig:transformer_visualization}
     \end{subfigure}
     \caption{Left (a): The PFN learns to approximate the PPD of a given prior in the offline stage to yield predictions on a new dataset in a single forward pass in the online stage. Right (b): Training samples $\{(x_1,y_1), \dots, (x_3,y_3)\}$ are transformed to $3$ tokens, which attend to each other; test samples $x_4$ and $x_5$ attend only to the training samples. Plots based on \cite{muller-iclr22a}.}
     \label{fig:pfn_overview}
%     \vspace*{-0.5cm}
\end{figure}
%
%PFNs leverage large-scale ML techniques to enable fast approximation of the Posterior Predictive Distribution (PPD) given any prior that can be sampled from ~\citep{muller-iclr22a}. Given a sufficiently powerful model PFN approximations are provably error-free.
%

\slimparagraph{Real-World Inference} During inference, the trained model is applied to unseen real-world datasets. For a novel dataset with training samples $D_{train}$ and test features $x_{test}$, feeding $\langle{}D_{train},x_{test}\rangle$ as an input to the model trained above yields the PPD %is generated by calculating
$q_{\theta}(y|x_{test},D_{train})$ in a single forward-pass. The PPD class probabilities are then used as predictions for our real-world task. Thus, PFNs perform training and prediction in one step (similar to prediction with Gaussian Processes) and do not use gradient-based learning on data seen at inference time.  This learning capability can be termed as in-context learning (ICL), which has been observed to surprisingly emerge when training large neural sequence models for language tasks \citep{NEURIPS2020_1457c0d6}.
%for a real-world dataset $D$
%using the $\theta$ resulting from \textit{synthetic prior-fitting}.

\textbf{Architecture} 
PFNs rely on a Transformer \citep{vaswani-neurips17a} that encodes each feature vector and label as a token, allowing token representations to attend to each other, as depicted in Figure \ref{fig:transformer_visualization}.
They accept a variable length training set $D_{train}$ of feature and label vectors (treated as a set-valued input to exploit permutation invariance) as well as a variable length query set of feature vectors $x_{test} = \{x_{({test},1)}, \dots, x_{({test},m)}\}$ and return estimates of the PPD for each query.
% PFNs are once trained offline in the \textit{synthetic prior-fitting} phase and are then applied to arbitrary many real-world datasets in the online \textit{real-world inference} phase.
%During meta-training, the PFN, is trained on a ref the objective of PPD approximation as a supervised classification problem:
%A PFN is meta-trained by repeatedly drawing artificial datasets from a prior $p(D) \sim p(\theta)$

\slimparagraph{Tabular Data} Amongst other experiments, \cite{muller-iclr22a} demonstrated PPD approximation with PFNs on binary classification for tiny, balanced tabular datasets with $30$ training examples. Here, we substantially improve performance and scale up to small datasets with up to 1\,000 data points and up to 10 classes, including imbalance.
We discuss detailed changes in Appendix \ref{app:comparison_muller}.

\iffalse
\paragraph{Relation to Prior Works}
Vs Pre-training, Vs Meta-Learning, vs Non parametric transformers

"meta-learning" from the Introduction. Especially para 2-3. Not sure about works around zero-shot meta learning in small tabular datasets but would have liked to see the connection out there with some links. The paper appears well rounded in motivating each design choice or contrasting with what is out there. Zero/One-shot meta learning maybe was something that felt missing as a mention.
Also related to this: Pg 1, Para 3 could do with some references as mentioned above 

Also an interesting point: We do not use gradient for learning, more in line with biologically plausible learning
\fi

%------------------------------------------------------
\section{The TabPFN: A PFN fitted on a new prior for tabular data}\label{sec:tabpfn}
%------------------------------------------------------

Our \methodname{} is a Prior-data Fitted Network (PFN, see Section \ref{sec:prior_fitted_intro}) that is fitted on data sampled from a novel prior for tabular data we introduce in Section \ref{sec:tabprior}. 
%We follow the original PFN architecture~\citep{muller-iclr22a} which uses a Transformer encoder~\citep{vaswani-neurips17a} with a specific attention mask and without positional encodings. 
We modify the original PFN architecture~\citep{muller-iclr22a} in two ways.
i) We make slight modifications to the attention masks, yielding shorter inference times.
ii) Additionally, we enable our model to work on datasets with different numbers of features by zero-padding.
These architectural enhancements alongside our main contributions compared to \cite{muller-iclr22a} are detailed in Appendix \ref{app:pfn architecture change}.

In the prior-fitting phase, we train the \methodname{} once on samples from the prior described in Section~\ref{sec:tabprior}. To be more precise, we trained a 12-layer Transformer for 18\,000 batches of 512 synthetically generated datasets each, which required a total of 20 hours on one machine with 8 GPUs (Nvidia RTX 2080 Ti). This yielded a single network that is used for all our evaluations.
While this training step is moderately expensive, it is done offline, in advance, and only once for the \methodname{}, as part of our algorithm development. The same TabPFN model is used for all experiments in this paper. Full details about our TabPFN training are given in Appendix \ref{sec:our_setup}.
%\ref{app:transformer_hypers,app:training_details}

During inference, the \methodname{} approximates the PPD for our dataset prior, i.e., it approximates the marginal predictions across our spaces of SCMs and BNNs (see Section \ref{sec:tabprior}), including a bias towards simple and causal explanations for the data.
%Looking back at the PPD equation (Equation \ref{eq:ppd}), we can understand our \methodname{}\textquotesingle{}s predictions as follows: The PPD integrates predictions over the space of hypotheses in our prior (which includes SCM graphs, BNN architectures, activation functions, etc.), where the weight of each hypothesis is determined by the product of the likelihood of observing the data given this hypothesis and the probability of this hypothesis in our prior. 
In our experiments, we present predictions for a single forward pass of our \methodname{}, as well as predictions that ensemble 32 forward passes of datasets modified by a power transformation (applied with probability 0.5) and rotating the indices of feature columns and class labels (see Appendix \ref{app:comparison_muller} for details).\footnote{Since we already rotate the indices of feature columns and class labels while training the \methodname{}, it can in principle learn a representation that is invariant to both of these rotations. We expect that, using a longer training time or a larger Transformer, \methodname{} would learn these invariances and make these rotations obsolete.}
%Full details about our training and inference setup can be found in Appendix \ref{sec:our_setup}.

%------------------------------------------------
\section{A Prior for Tabular Data}\label{sec:tabprior}
%------------------------------------------------
The performance of our method crucially depends on the specification of a suitable prior, as the PFN approximates the PPD for this prior.
Section \ref{sec:probabilistic prior} outlines a fundamental technique for our prior: we use distributions instead of point-estimates for almost all of our prior's hyperparameters. Section \ref{sec:modelling simplicity assumptions} motivates simplicity in our prior, while Sections \ref{sec:causal prior} and \ref{sec:bnn prior} describe how we use Structural Causal Models (SCMs) and Bayesian Neural Networks (BNNs) as fundamental mechanisms to generate diverse data in our prior. Since our SCM and BNN priors only yield regression tasks, we show how to convert them to classification tasks in Section \ref{sec:multi-class prediction}. We describe additional refinements to our prior to reflect peculiarities of tabular data (correlated and categorical features, exponentially scaled data and missing values) better in Appendix \ref{sec:tabular data refinement}.
\subsection{Fundamentally Probabilistic Models}
\label{sec:probabilistic prior}
Fitting a model typically requires finding suitable hyperparameters, e.g., the embedding size, number of layers and activation function for NNs. Commonly, resource-intensive searches need to be employed to find suitable hyperparameters \citep{zoph-iclr17a,feurer-automlbook19a}.
The result of these searches, though, is only a point estimate of the hyperparameter choice.
Ensembling over multiple architectures and hyperparameter settings can yield a rough approximation to a distribution over these hyperparameters and has been shown to improve performance \citep{zaidi-neurips21a,hyperensemble}.
This, however, scales linearly in cost with the number of choices considered.

In contrast, PFNs allow us to be fully Bayesian about our prior's hyperparameters. By defining a probability distribution over the space of hyperparameters in the prior, such as BNN architectures, the PPD approximated by our \methodname{} jointly integrates over this space and the respective model weights.
We extend this approach to a mixture not only over hyperparameters but distinct priors: we mix a BNN and an SCM prior,
%(Section \ref{sec:causal prior}) and a BNN prior (Section \ref{sec:bnn prior}), 
each of which again entails a mixture of architectures and hyperparameters.
\subsection{Simplicity}
\label{sec:modelling simplicity assumptions}
We base our priors on a notion of simplicity, such as stated by Occam's Razor or the Speed Prior \citep{speedprior}.
When considering competing hypotheses, the simpler one
%, e.g., the one requiring fewer parameters, 
is to be preferred.
Work in cognitive science has also uncovered this preference for simple explanations in human thinking~\citep{explanatoryvalues}.
Any notion of simplicity, however, depends on choosing a particular criterion that defines simplicity.
In the following, we introduce priors based on SCMs and BNNs, in which we implement simplicity as graphs with few nodes and parameters. %, as well as, small parameter values.
\subsection{SCM Prior}

\label{sec:causal prior}
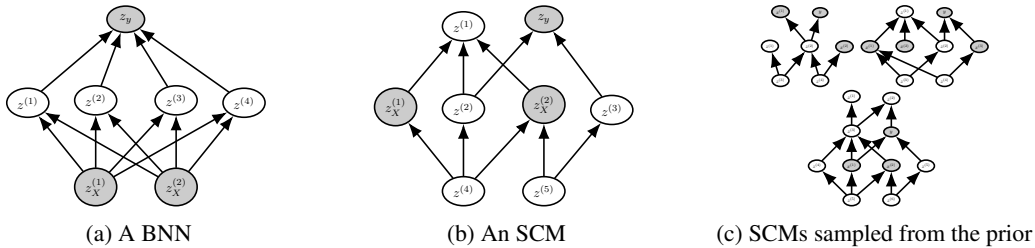
\begin{figure}[t]
     \centering
     \begin{subfigure}[b]{0.3\textwidth}
         \centering
         \begin{tikzpicture}[-Latex,auto,node distance =0.7 cm and 0.5 cm,semithick,
    state/.style ={ellipse, draw, minimum width = 1.0 cm, minimum height = 0.7 cm},every node/.style={scale=0.5},
    point/.style = {circle, draw, inner sep=0.04cm,fill,node contents={}},
    bidirected/.style={Latex-Latex,dashed},
    el/.style = {inner sep=2pt, align=left, sloped}]
    \node[state,fill=gray!40!white] (1) {$z_X^{(1)}$};
    \node[state,fill=gray!40!white] (2) [right =of 1] {$z_X^{(2)}$};
    \node[state] (3) [above left =of 1,yshift=0.2cm] {$z^{(1)}$};
    \node[state] (4) [above =of 1] {$z^{(2)}$};
    \node[state] (5) [above =of 2] {$z^{(3)}$};
    \node[state] (6) [above right =of 2,yshift=0.2cm] {$z^{(4)}$};
    \node[state] (7) [above =of 4,xshift=0.8cm,fill=gray!40!white] {$z_y$};

    \path (1) edge (3);
    \path (1) edge (4);
    \path (1) edge (5);
    \path (1) edge(6);
    \path (2) edge (3);
    \path (2) edge (4);
    \path (2) edge (5);
    \path (2) edge (6);
    
    \path (3) edge (7);
    \path (4) edge (7);
    \path (5) edge (7);
    \path (6) edge (7);
\end{tikzpicture}
         \caption{A BNN\label{fig:bnn}}
     \end{subfigure}
     \hfill
     \begin{subfigure}[b]{0.3\textwidth}
         \centering
         \begin{tikzpicture}[-Latex,auto,node distance =0.7 cm and 0.5 cm,semithick,
    state/.style ={ellipse, draw, minimum width = 1.0 cm, minimum height = 0.7 cm},every node/.style={scale=0.5},
    point/.style = {circle, draw, inner sep=0.04cm,fill,node contents={}},
    bidirected/.style={Latex-Latex,dashed},
    el/.style = {inner sep=2pt, align=left, sloped}]
    \node[state] (1) {$z^{(4)}$};
    \node[state] (2) [right =of 1] {$z^{(5)}$};
    \node[state,fill=gray!40!white] (3) [above left =of 1,yshift=0.2cm] {$z_X^{(1)}$};
    \node[state] (4) [above =of 1] {$z^{(2)}$};
    \node[state,fill=gray!40!white] (5) [above =of 2] {$z_X^{(2)}$};
    \node[state] (6) [above right =of 2,yshift=0.2cm] {$z^{(3)}$};
    \node[state] (7) [above =of 4] {$z^{(1)}$};
    \node[state,fill=gray!40!white] (8) [above =of 5] {$z_y$};

    \path (1) edge (3);
    \path (1) edge (4);
    \path (1) edge (5);
    \path (2) edge (5);
    \path (2) edge (6);
    
    \path (3) edge (7);
    \path (5) edge (7);
    \path (4) edge (7);
    
    \path (4) edge (8);
    \path (6) edge (8);
\end{tikzpicture}
         \caption{An SCM\label{fig:scm}}
     \end{subfigure}
     \hfill
     \begin{subfigure}[b]{0.3\textwidth}
         \centering
         % \documentclass{article}
% \usepackage{tikz}
% \usetikzlibrary{shapes,decorations,arrows,calc,arrows.meta,fit,positioning}
% \begin{document}
\begin{tikzpicture}[-Latex,auto,node distance =0.3 cm and 0.3 cm,semithick,every node/.style={scale=0.2},
    state/.style ={ellipse, draw, minimum width = 1.0 cm},
    point/.style = {circle, draw, inner sep=0.04cm,fill,node contents={}},
    bidirected/.style={Latex-Latex,dashed},
    el/.style = {inner sep=2pt, align=left, sloped}]
    \node[state] (1) {$z^{(3)}$};
    \node[state] (2) [right =of 1] {$z^{(4)}$};
    \node[state] (4) [above =of 1,xshift=-0.7cm] {$z^{(1)}$};
    \node[state] (5) [above =of 2,xshift=-0.7cm] {$z^{(2)}$};
    \node[state,fill=gray!40!white] (6) [above right =of 2,yshift=0.2cm,xshift=-0.7cm] {$x^{(2)}$};
    \node[state,fill=gray!40!white,xshift=0.7cm] (7) [above =of 4] {$x^{(1)}$};
    \node[state,fill=gray!40!white,xshift=0.7cm] (8) [above =of 5] {$y$};

    \path (1) edge (4);
    \path (1) edge (5);
    \path (2) edge (5);
    \path (2) edge (6);
    
    \path (5) edge (7);
    
    \path (5) edge (8);
\end{tikzpicture}
% \documentclass{article}
% \usepackage{tikz}
% \usetikzlibrary{shapes,decorations,arrows,calc,arrows.meta,fit,positioning}
% \begin{document}
\begin{tikzpicture}[-Latex,auto,node distance =0.3 cm and 0.3 cm,semithick,every node/.style={scale=0.2},
    state/.style ={ellipse, draw, minimum width = 1.0 cm},
    point/.style = {circle, draw, inner sep=0.04cm,fill,node contents={}},
    bidirected/.style={Latex-Latex,dashed},
    el/.style = {inner sep=2pt, align=left, sloped}]
    \node[state] (1) {$z^{(3)}$};
    \node[state] (2) [right =of 1] {$z^{(4)}$};
    \node[state,fill=gray!40!white] (3) [above left =of 1,yshift=0.2cm] {$x^{(1)}$};
    \node[state,fill=gray!40!white] (4) [above =of 1] {$x^{(2)}$};
    \node[state] (5) [above =of 2] {$z^{(2)}$};
    \node[state,fill=gray!40!white] (6) [above right =of 2,yshift=0.2cm] {$x^{(3)}$};
    \node[state] (7) [above =of 4] {$z^{(1)}$};
    \node[state,fill=gray!40!white] (8) [above =of 5] {$y$};

    \path (1) edge (3);
    \path (1) edge (5);
    \path (2) edge (3);
    \path (2) edge (6);
    
    \path (3) edge (7);
    \path (5) edge (7);
    \path (4) edge (7);
    
    \path (6) edge (8);
    \path (5) edge (8);
\end{tikzpicture}
% \documentclass{article}
% \usepackage{tikz}
% \usetikzlibrary{shapes,decorations,arrows,calc,arrows.meta,fit,positioning}
% \begin{document}
\begin{tikzpicture}[-Latex,auto,node distance =0.3 cm and 0.3 cm,semithick,every node/.style={scale=0.2},
    state/.style ={ellipse, draw, minimum width = 1.0 cm},
    point/.style = {circle, draw, inner sep=0.04cm,fill,node contents={}},
    bidirected/.style={Latex-Latex,dashed},
    el/.style = {inner sep=2pt, align=left, sloped}]
    \node[state] (1) {$z^{(5)}$};
    \node[state] (2) [right =of 1] {$z^{(6)}$};
    \node[state] (3) [above left =of 1,yshift=0.2cm] {$z^{(4)}$};
    \node[state,fill=gray!40!white] (4) [above =of 1] {$x^{(1)}$};
    \node[state,fill=gray!40!white] (5) [above =of 2] {$x^{(2)}$};
    \node[state] (6) [above right =of 2,yshift=0.2cm] {$z^{(5)}$};
    \node[state] (7) [above =of 4] {$z^{(3)}$};
    \node[state,fill=gray!40!white] (8) [above =of 5] {$y$};
    \node[state] (9) [above =of 7] {$z^{(1)}$};
    \node[state] (10) [above =of 8] {$z^{(2)}$};

    \path (1) edge (3);
    \path (1) edge (4);
    \path (1) edge (5);
    \path (2) edge (5);
    \path (2) edge (6);
    
    \path (3) edge (7);
    \path (5) edge (7);
    \path (4) edge (7);
    
    \path (4) edge (8);
    \path (6) edge (8);
    
    \path (7) edge (9);
    \path (8) edge (10);
    \path (7) edge (10);
\end{tikzpicture}
         \caption{SCMs sampled from the prior}
     \end{subfigure}
        \caption{Overview of graphs generating data in our prior. Inputs $x$ are mapped to the output $y$ through unobserved nodes $z$. Plots based on \cite{muller-iclr22a}.}
        \label{fig:structural_graphs}
\end{figure}
It has been demonstrated that causal knowledge can facilitate various ML tasks, including semi-supervised learning, transfer learning and out-of-distribution generalization \citep{scholkopf2012causal, NEURIPS2019_2172fde4, rotenh2018}.
Tabular data often exhibits causal relationships between columns, and causal mechanisms have been shown to be a strong prior in human reasoning \citep{waldmann2013causal, explanatoryvalues}.
Thus, we base our \methodname{} prior on SCMs that model causal relationships~\citep{pearl_2009, causalinferencelearningbook}.
An SCM %$\mathcal{G} := (Z, \epsilon{})$
consists of a collection $Z := (\{z_1, \dots, z_k\})$ of structural assignments (called mechanisms): $z_i = f_i(z_{\text{PA}_{\mathcal{G}}(i)}, \epsilon_i), \label{scm_eq}$
where $\text{PA}_{\mathcal{G}}(i)$ is the set of parents of the node $i$ (its direct causes) in an underlying DAG $\mathcal{G}$ (the causal graph), $f_i$ is a (potentially non-linear) deterministic function and $\epsilon_i$ is a noise variable.
Causal relationships in $\mathcal{G}$ are represented by directed edges pointing from causes to effects and each mechanism $z_i$ is assigned to a node in $\mathcal{G}$, as visualized in Figure \ref{fig:structural_graphs}.

\slimparagraph{Predictions based on ideas from causal reasoning}
Previous works have applied causal reasoning to predict observations on unseen data by using causal inference, a method which seeks to identify causal relations between the components of a system by the use of interventions and observational data \citep{pearl2010causal, PearlMackenzie18, lin2021scoping}. 
The predicted causal representations are then used to make observational predictions on novel samples or to provide explainability.
Most existing work focuses on determining a single causal graph to use for downstream prediction, which can be problematic since most kinds of SCMs are non-identifiable without interventional data, and the number of compatible DAGs explodes due to the combinatorial nature of the space of DAGs.
Recently \citet{ke2022} and \citet{lorch2022amortized} use transformers to approximate the causal graphs from observational and interventional data.
We skip any explicit graph representation in our inference step and approximate the PPD directly.
Thus, we do not perform causal inference but solve the downstream prediction task directly.
This implicit assumption of SCM-like processes generating our data can be explained in Pearl\textquotesingle{}s ``ladder of causation'', an abstraction of inference categories, where each higher rung represents a more involved notion of inference \citep{PearlMackenzie18}. At the lowest rung lies association, which includes most of ML. The second rung considers predicting the effect of interventions, i.e., what happens when we influence features directly. %lastly, counterfactual inference aims to predict what would have happened if things were different, after the fact.
Our work can be considered as ``rung 1.5'', similar to \cite{NEURIPS2020_1068bceb, SchaarMiracleCDL}: we do not perform causal reasoning, but make association-based predictions on observational data assuming SCMs model common datasets well. In Figure \ref{fig:inductive_bias} in Appendix~\ref{app:ssec:addresults}, we experimentally show that our indeed align with simple SCM hypotheses.
% Similarly, recent work by \cite{NEURIPS2020_1068bceb} uses pointwise DAG representations as a regularizer of NN training, forcing predictions to be regularized by causal structures but not making actual predictions of causal structures during inference. 

\begin{figure}[t]
     \centering
     \begin{subfigure}[b]{0.49\textwidth}
         \centering
         \hspace*{-0.8cm}
            \includegraphics[width=0.5\columnwidth]{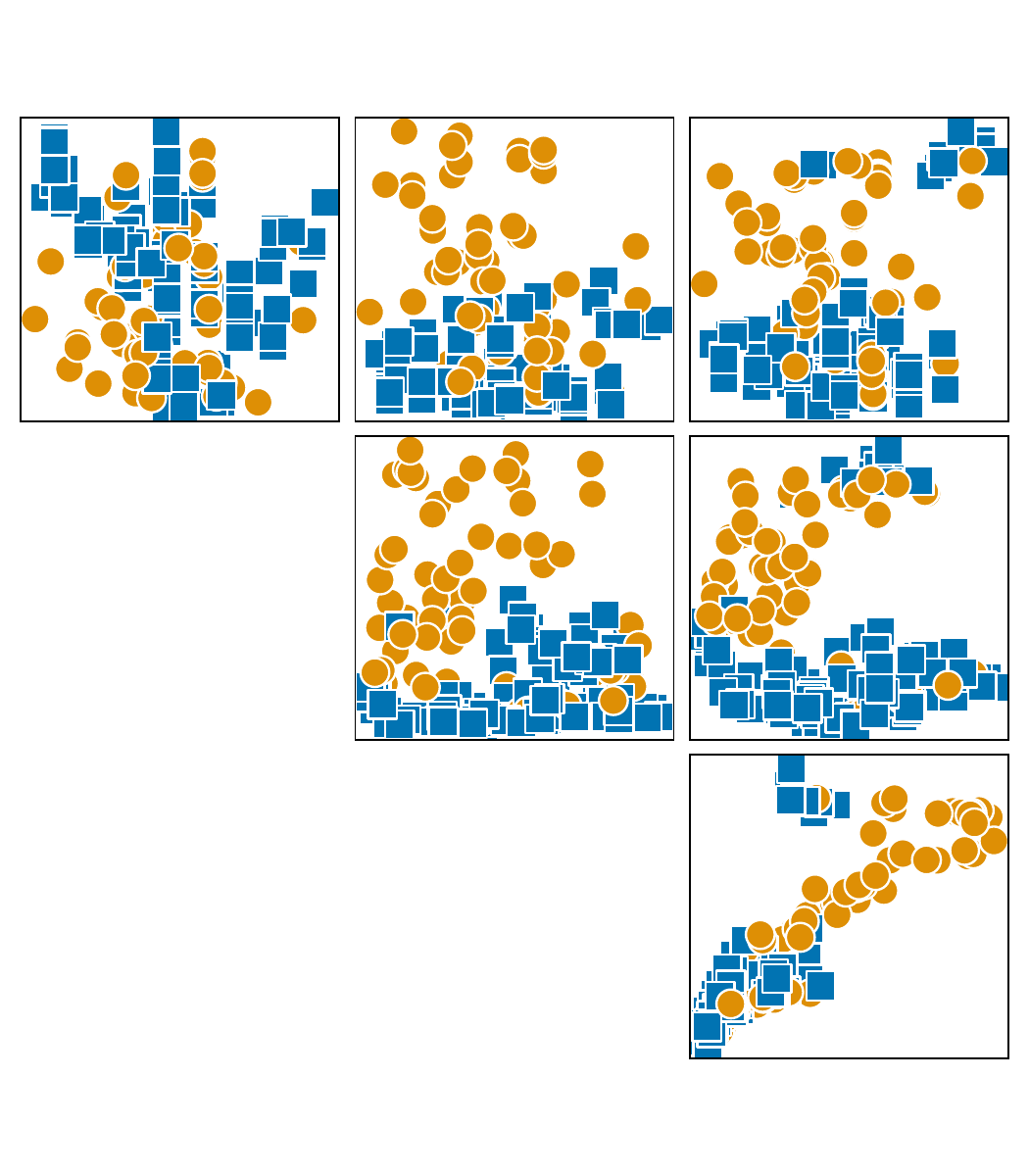}
%         \caption{}
         \centering
         \hspace*{-0.2cm}
            \includegraphics[width=0.5\columnwidth]{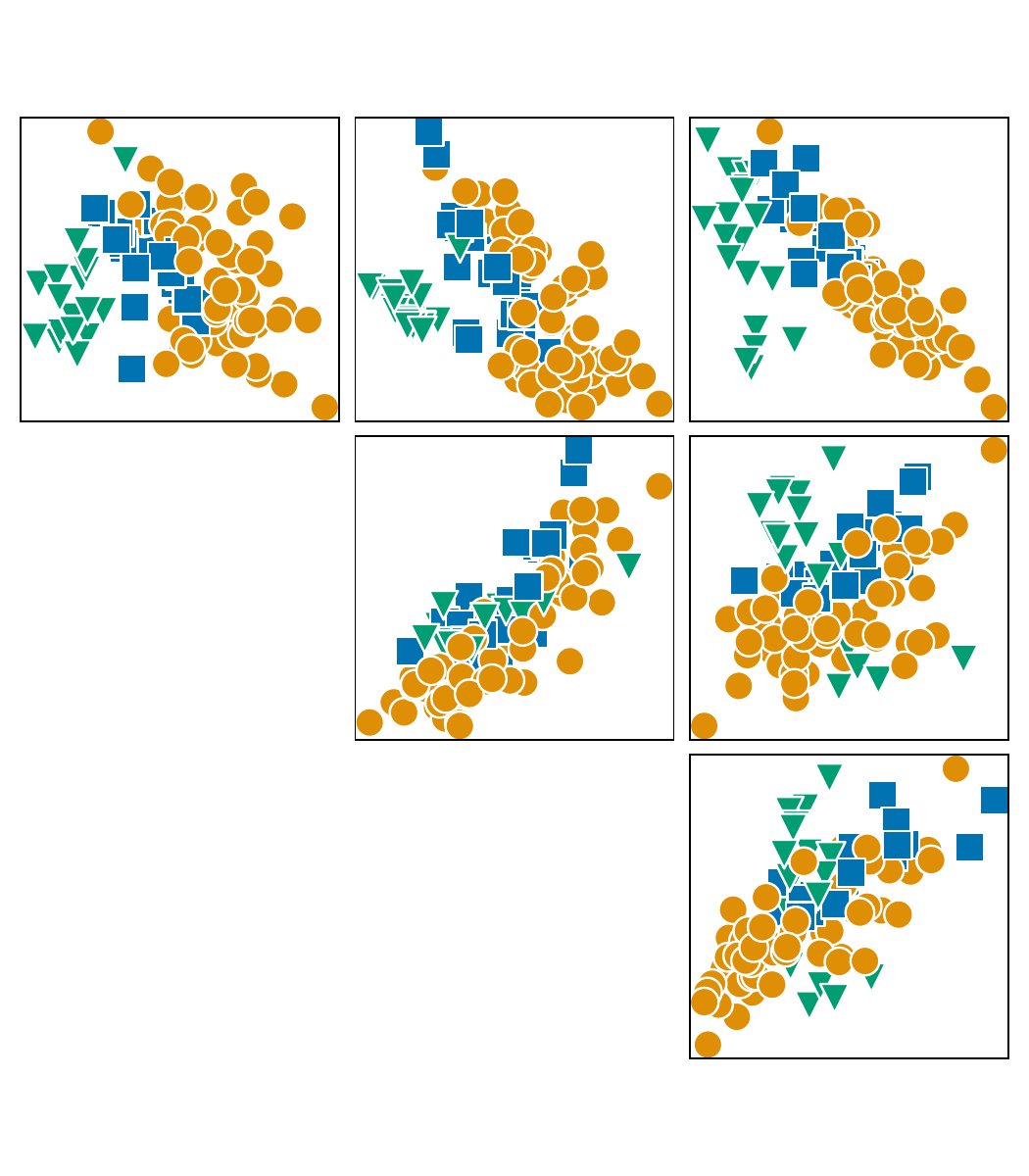}
         \caption{Synthetic datasets\label{fig:samples example}}
     \end{subfigure}
     \begin{subfigure}[b]{0.49\textwidth}
         \centering
         \hspace*{-0.8cm}
            \includegraphics[width=0.5\columnwidth]{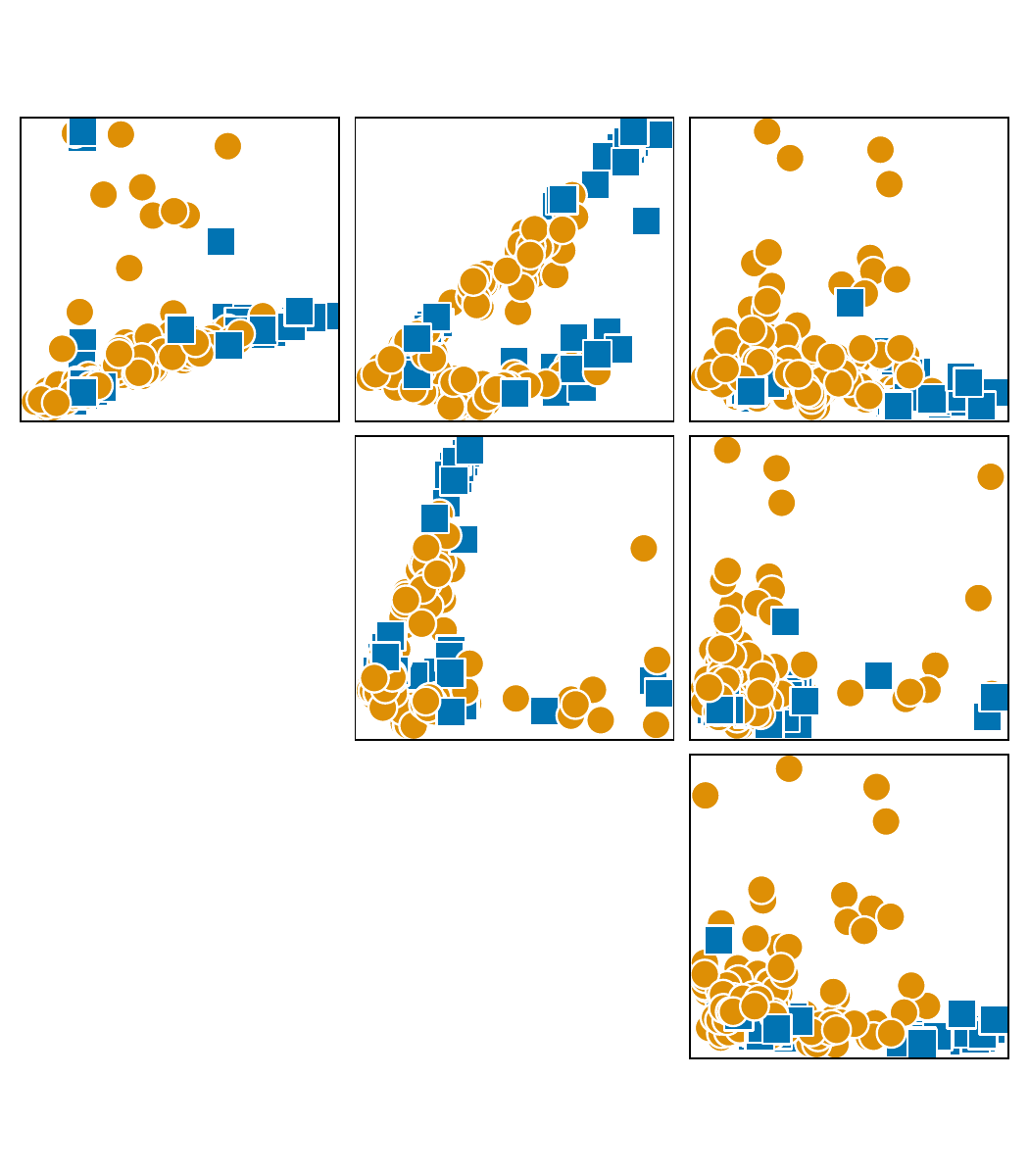}
%         \caption{}
         \centering
         \hspace*{-0.2cm}
            \includegraphics[width=0.5\columnwidth]{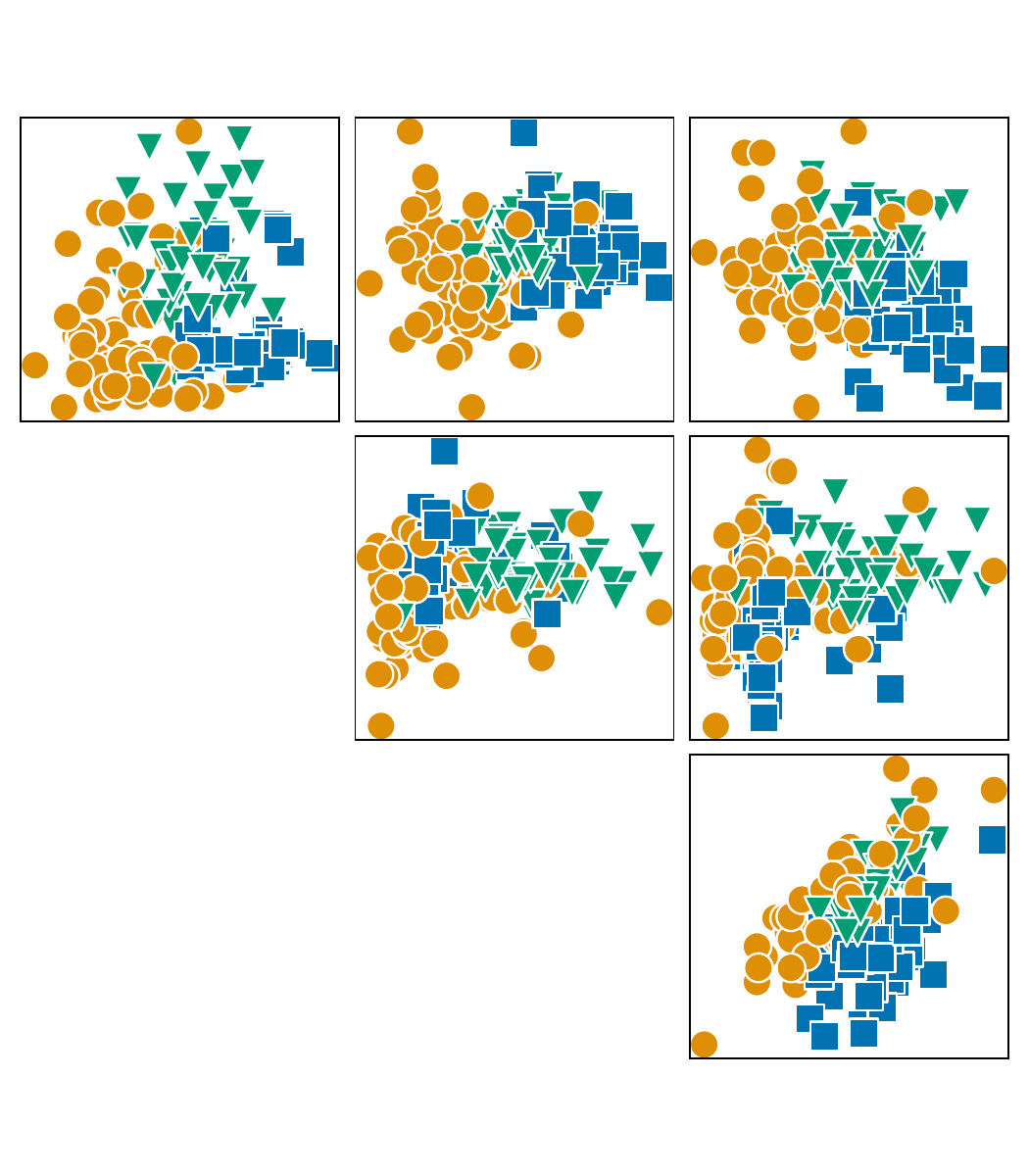}
         \caption{Actual datasets}
         \label{fig:actual example}
     \end{subfigure}
%     \vspace*{-0.3cm}
     \caption{Each sub-plot shows the combination of two features, each dot represents a sample, color indicates the class label. %The diagonal shows histograms of each feature, given a class. 
     (a) Two synthetic datasets generated by our causal tabular data prior. Numeric SCM outputs are mapped to classes as described in Section \ref{sec:multi-class prediction}. (b) Two datasets from our validation datasets: Parkinsons (Left) and Wine (Right).
     }
     \label{fig:samples_and_correlation}
\end{figure}

\slimparagraph{Defining a prior based on causal models}
To create a PFN prior based on SCMs, we have to define a sampling procedure that creates supervised learning tasks (i.e., datasets).
Here, each dataset is based on one randomly-sampled SCM (including the DAG structure and deterministic functions $f_i$).
Given an SCM, we sample a set $z_X$ of nodes in the causal graph $\mathcal{G}$, one for each feature in our synthetic dataset, as well as one node $z_y$ from $\mathcal{G}$.
These nodes are observed nodes: values of $z_X$ will be included in the set of features, while values from $z_y$ will act as targets.
For each such SCM and list of nodes $z_X$ and $z_y$, $n$ samples are generated by sampling all noise variables in the SCM $n$ times, propagating these through the graph and retrieving the values at the nodes $z_X$ and $z_y$ for all $n$ samples.
Figure~\ref{fig:scm} depicts an SCM with observed feature- and target-nodes in grey.
The resulting features and targets are correlated through the generating DAG structure.
This leads to features conditionally dependent through forward and backward causation, i.e., targets might be a cause or an effect of features.
In Figure~\ref{fig:samples_and_correlation}, we compare samples generated by two distinct SCMs to actual datasets, demonstrating the diversity in the space of datasets our prior can model. 
%Making predictions on a dataset generated in this way entails inferring potential graph structures generating the training samples and making predictions based on these graphs.
%The Bayesian PPD approximated by \methodname{} entails all SCM

In this work, we instantiate a large subfamily of DAGs and deterministic functions $f_i$ to build SCMs described in Appendix \ref{sec:details_scm_prior}.
Since efficient sampling is the only requirement we have, the instantiated subfamily is very general, including multiple activation functions and noise distributions.

\subsection{BNN Prior}
\label{sec:bnn prior}
%\vspace*{-0.1cm}

We also consider a BNN prior as introduced by \citet{muller-iclr22a} and mix it with the SCM prior described above by randomly sampling datasets during PFN training from either one or the other prior with equal probability.
%We begin by introducing a prior based on Bayesian Neural Networks (BNNs) from \citep{muller-iclr22a}.
To sample a dataset from the BNN prior, we first sample an
%(noisy) 
NN architecture and its weights.
Then, for each data point in the to-be-generated dataset, we sample an input $x$,
% \sim \mathcal{N}(\mathbf{0},\mathbf{I})$,
%a standard normal, 
feed it through the BNN with sampled noise variables and use the output $y$ as a target (see Figure~\ref{fig:bnn}).
%\footnote{We note that, while tabular data is not always normalized, we apply z-normalization to the columns.}.
An experimental ablation of this BNN prior 
%based on \cite{muller-iclr22a} 
to our final mixture of both SCM-based and BNN-based priors in Appendix \ref{app: prior_ablation} demonstrates the strength of our novel SCM-based prior. 

\subsection{Multi-class Prediction}
\label{sec:multi-class prediction}
%\vspace*{-0.11cm}
So far, the described priors return scalar labels. In order to generate synthetic classification labels for imbalanced multi-class datasets, we need to transform our scalar labels $\hat{y}$ to discrete class labels $y$. We do so by splitting the values of $\hat{y}$ into intervals that map to class labels:
\vspace{-5mm}\\
\begin{enumerate}[i), itemsep=-0.1em]
    \item We sample the number of classes $N_c \sim p(N_c)$, where $p(N_c)$ is a distribution over integers.
    \item  We sample $N_c-1$ class bounds $B_{i}$ randomly from the set of continuous targets $\hat{y}$. % for half of the datasets and from a standard normal distribution for the other half.
    \item We map each scalar label $\hat{y}_i$ to the index of the unique interval that contains it:
    $y_i \gets \sum_{j} [B_j < \hat{y}_i]$, where $[\cdot]$ is the indicator function.
\end{enumerate}
%\hphantom{~~~~}(1) We sample the number of classes $N_c \sim p(N_c)$, where $p(N_c)$ is a distribution over integers.

%\hphantom{~~~~}(2) We sample $N_c-1$ quantile bounds $B_c \sim p(B_c|N_c)$, uniformly between [0, 1].\\
%\hphantom{~~~~}(2) We sample $N_c-1$ bounds $B_c \sim p(B_c|N_c,\hat{y})$. $p(B_c|N_c,\hat{y})$ samples bounds uniformly from $\hat{y}$ for half of the datasets and from a standard normal distribution for the other half.

%\hphantom{~~~~}(3) We map each scalar label $\hat{y}_i$ to the unique range that contains it:\vspace{2mm}\\
%\hphantom{~~~~~~~~~~}$y_i \gets \sum_{j} B_j < \textit{rank}(\hat{y}_i)/n$, where \textit{rank} maps $\bar{y}_i$ to its rank in $\bar{y}$.\vspace{2mm}\\
%\hphantom{~~~~~~~~~~}$y_i \gets \sum_{j} [B_j < \hat{y}_i]$, where $[\cdot]$ is the indicator function.\vspace{2mm}\\
%For example for $N_c = 2$ the quantile bound $B_c = \{0.5\}$ would define two quantile ranges $\{[0.0, 0.5], [0.5, 1.0]\}$. This would map the lower quantile of values in $\hat{y}$ to 0 and the upper half to 1, yielding balanced binary classification. Finally, we shuffle the labelling of classes, i.e. we remove the ordering of class labels w.r.t.\ the quantile range.
%\vspace*{-0.11cm}
For example, with $N_c = 3$ classes the bounds $B_c = \{-0.1, 0.5\}$ would define three intervals $\{(-\infty, -0.1], (-0.1, 0.5], (0.5, \infty)\}$. Any $\hat{y}_i$ would be mapped to the label $0$ if it is smaller than $-0.1$, to $1$ if lies in $(-0.1, 0.5]$ and to $2$ otherwise.
Finally, we shuffle the labels of classes, i.e. we remove the ordering of class labels w.r.t.\ the ranges.
% \hphantom{~~~~}(4) With probability $p_s$, a hyperparameter, we shuffle the labelling of classes, that is, we remove the ordering of class labels wrt.\ the quantile range.
% For the distribution of the number of classes, $P(N_c)$ is a uniform with an extra $50\%$ probability for the binary case (followed by renormalization). $P(B_c)$ controls the class distribution and is chosen to sample bounds uniformly in the range $[0, 1]$, and $p_s$ is a hyperparameter.

%------------------------------------------------------
\section{Experiments}\label{sec:experiments}
%------------------------------------------------------
%------------------------------------------------------

%\vspace*{-0.15cm}

%------------

\begin{figure}[t]
     \centering
     \begin{subfigure}[b]{1.0\textwidth}
         \centering
            \includegraphics[width=0.85\columnwidth]{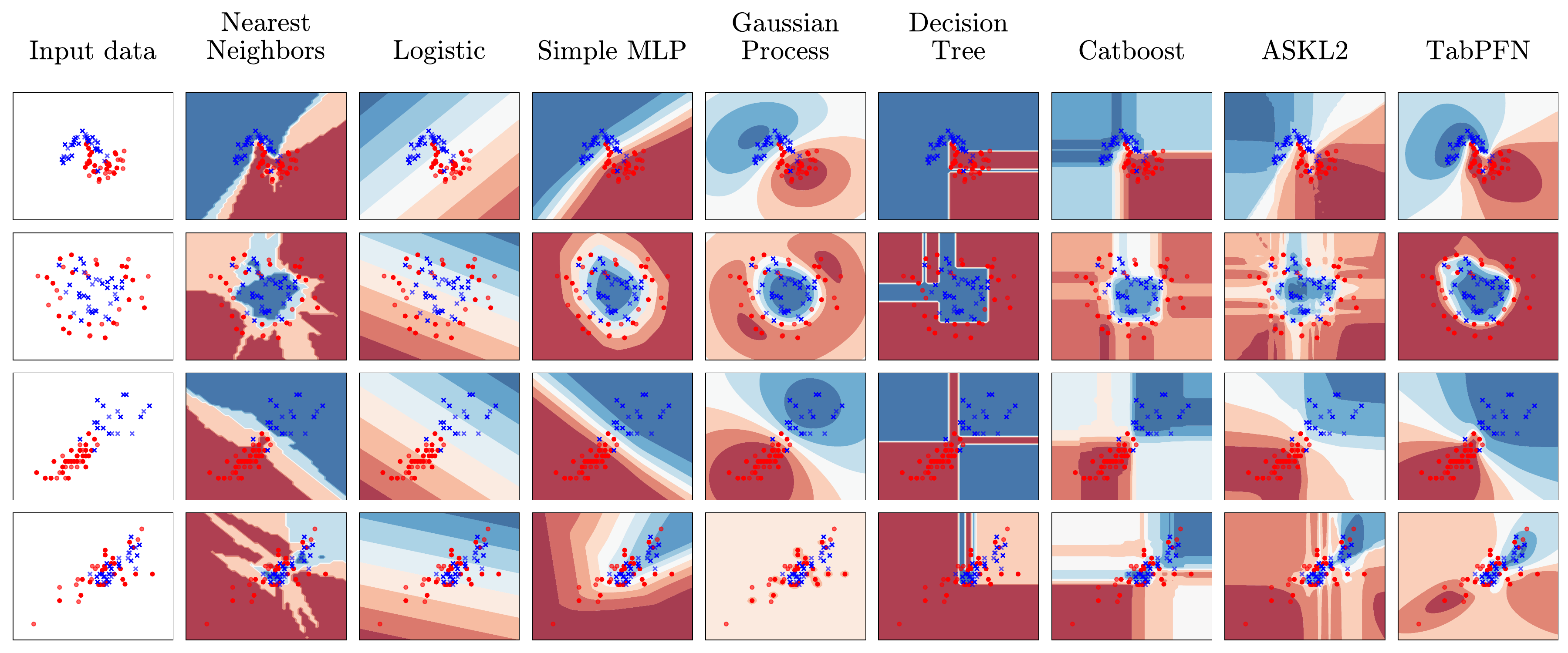}
     \end{subfigure}
     \caption{Decision boundaries on toy datasets generated with \emph{scikit-learn}~\citep{scikit-learn}.}
%     \vspace*{-0.3cm}
     \label{fig:toy_samples}
%     \vspace*{-0.1cm}
\end{figure}
\subsection{Evaluation on Toy Problems}\label{ssec:toyresults}
%\vspace*{-0.11cm}

We first qualitatively compare our \methodname{} to standard classifiers (without hyperparameter tuning) in Figure \ref{fig:toy_samples}.
The top row shows the \emph{moons} dataset with noise. The \methodname{} accurately models the decision boundary between samples; also, similar to Gaussian processes, uncertainties are large for points far from observed samples.
The second row shows the \emph{circles} dataset with noise: the \methodname{} accurately models the circle\textquotesingle{}s shape with high confidence anywhere outside the region where samples are mixed.
The third row shows two classes and features from the iris dataset, while the fourth rows shows two classes and features from the wine dataset (both in \emph{scikit-learn}~\citep{scikit-learn}); in both of these cases, the TabPFN makes intuitive, well-calibrated predictions.
% TODO: Could have some more explanations if there is space

\subsection{Evaluation on Tabular ML Tasks}\label{ssec:tabresults}\vspace*{-0.1cm}
%------------------------------------------------------
Now, we turn to an empirical analysis of our method for real-world classification tasks.
We compare our method against state-of-the-art ML and AutoML methods for tabular classification.
%. For this we use on the \openmlcc{}~\citep{bischl-neuripsdbt21a}, a curated open-source list of medium-sized classification datasets 
% We briefly describe our experimental setup before we turn to discussing the results and provide further details in Appendix~\ref{app:expdetails}.
% I just moved this

\slimparagraph{Datasets} As test datasets, we used all datasets from the curated open-source \openmlcc{} benchmark suite~\citep{bischl-neuripsdbt21a} that contain up to $2\,000$ samples ($1\,000$ for the training split), $100$ features and $10$ classes. The resulting set comprises $30$ datasets. We split these datasets into 18 datasets that contain only numerical features and no missing values, and 12 other datasets that contain categorical features and/or missing values.
In the main paper, in Figure \ref{fig:perfovertime} and Table \ref{table:tabular_results_table_small}, we limit our analysis to the case of numerical datasets without missing values, which we focussed the development of our \methodname{} prior on. Results on all of the 30 datasets are given in Appendix \ref{app:sec:detailed_results} and still show strong aggregate performance for \methodname{}, albeit not as strong as for the purely numerical case, due to generally worse performance of \methodname{} on datasets with categorical features and/or missing values. 
%The results in Figure \ref{fig:perfovertime} and Table \ref{table:tabular_results_table_small} refer to results on the numerical datasets, while we report results on all datasets in the Appendix. 
We focus on small datasets because (1) small datasets are often encountered in real-world applications~\citep{dua_2019}, (2) existing DL methods are most limited in this domain~\citep{grinsztajn2022why} and (3) the \methodname{} would be significantly more expensive to train and evaluate for larger datasets, a limitation detailed in Appendix \ref{sec:limitations}.

\definecolor{color0}{HTML}{0173b2}
\definecolor{color1}{HTML}{de8f05}
\definecolor{color2}{HTML}{029e73}
\definecolor{color3}{HTML}{d55e00}
\definecolor{color4}{HTML}{cc78bc}
\definecolor{color5}{HTML}{ca9161}
\definecolor{color6}{HTML}{fbafe4}
\definecolor{color7}{HTML}{56b4e9} % the only color that is not original palette
\definecolor{color8}{HTML}{ece133}
\pgfdeclareplotmark{mystar}{
    \node[star,star point ratio=2.25,minimum size=10pt,
          inner sep=0pt,draw=none,solid, fill=color3] {};
}
\pgfplotsset{
    mystylepalette3/.style={
        height=6cm,
        width=\textwidth,
    },
}

\definecolor{color0}{rgb}{0.00392156862745098,0.450980392156863,0.698039215686274}
\definecolor{color1}{rgb}{0.870588235294118,0.56078431372549,0.0196078431372549}
\definecolor{color2}{rgb}{0.00784313725490196,0.619607843137255,0.450980392156863}
\definecolor{color3}{rgb}{0.835294117647059,0.368627450980392,0}
\definecolor{color4}{rgb}{0.8,0.470588235294118,0.737254901960784}
\definecolor{color5}{rgb}{0.792156862745098,0.568627450980392,0.380392156862745}
\definecolor{color6}{rgb}{0.984313725490196,0.686274509803922,0.894117647058824}
\definecolor{color7}{rgb}{0.925490196078431,0.882352941176471,0.2}
\definecolor{color8}{rgb}{0.337254901960784,0.705882352941177,0.913725490196078}

\begin{figure}[t]
     \centering
     \begin{subfigure}[t]{0.32\textwidth}
         \centering
         %\hspace*{-1.9cm}
         % This file was created with tikzplotlib v0.9.12.
\begin{tikzpicture}

\begin{axis}[
height=4.5cm,
legend style={fill opacity=0.8, draw opacity=1, text opacity=1, draw=none},
minor xtick={},
minor ytick={},
tick align=outside,
tick pos=left,
width=\textwidth,
x grid style={white!69.0196078431373!black},
xlabel={Given Time Budget},
xmin=0, xmax=3.55630250076729,
xtick style={color=black},
xtick={0,0.698970004336019,1.47712125471966,2.47712125471966,3.55630250076729},
xticklabels={1s,5s,30s,5min,1h},
y grid style={white!69.0196078431373!black},
ylabel={Mean ROC AUC},
ymin=0.88, ymax=0.94,
ytick style={color=black},
ytick={0.88,0.89,0.9,0.91,0.92,0.93,0.94},
mystylepalette
]

\path [draw=color1, fill=color1, opacity=0.1, line width=1pt]
(axis cs:1.77815125038364,0.928967293306433)
--(axis cs:1.77815125038364,0.923337389753079)
--(axis cs:2.47712125471966,0.924974473819512)
--(axis cs:2.95424250943932,0.923486228637966)
--(axis cs:3.55630250076729,0.925483282632585)
--(axis cs:3.55630250076729,0.932361731006138)
--(axis cs:3.55630250076729,0.932361731006138)
--(axis cs:2.95424250943932,0.931484813454741)
--(axis cs:2.47712125471966,0.930962953428816)
--(axis cs:1.77815125038364,0.928967293306433)
--cycle;

\path [draw=color2, fill=color2, opacity=0.1, line width=1pt]
(axis cs:0.698970004336019,0.910737185553028)
--(axis cs:0.698970004336019,0.857391368176483)
--(axis cs:1.17609125905568,0.900277185327172)
--(axis cs:1.47712125471966,0.914434995115592)
--(axis cs:1.77815125038364,0.916531984930485)
--(axis cs:2.47712125471966,0.915434402424111)
--(axis cs:2.95424250943932,0.916463424181783)
--(axis cs:3.55630250076729,0.915574928898626)
--(axis cs:3.55630250076729,0.923038329526335)
--(axis cs:3.55630250076729,0.923038329526335)
--(axis cs:2.95424250943932,0.923267835216241)
--(axis cs:2.47712125471966,0.920598560583346)
--(axis cs:1.77815125038364,0.922079227061932)
--(axis cs:1.47712125471966,0.921698894634802)
--(axis cs:1.17609125905568,0.91713079522044)
--(axis cs:0.698970004336019,0.910737185553028)
--cycle;

\path [draw=color4, fill=color4, opacity=0.1, line width=1pt]
(axis cs:1.77815125038364,0.924413096895011)
--(axis cs:1.77815125038364,0.879662881894458)
--(axis cs:2.47712125471966,0.919095307107579)
--(axis cs:2.95424250943932,0.920182106522037)
--(axis cs:3.55630250076729,0.921184592681795)
--(axis cs:3.55630250076729,0.92647378779733)
--(axis cs:3.55630250076729,0.92647378779733)
--(axis cs:2.95424250943932,0.925459467899862)
--(axis cs:2.47712125471966,0.925276145998611)
--(axis cs:1.77815125038364,0.924413096895011)
--cycle;

\path [draw=color3, fill=color3, opacity=0.1, line width=1pt]
(axis cs:0,0.935690524062943)
--(axis cs:0,0.930382577465569)
--(axis cs:3.55630250076729,0.930382577465569)
--(axis cs:3.55630250076729,0.935690524062943)
--(axis cs:3.55630250076729,0.935690524062943)
--(axis cs:0,0.935690524062943)
--cycle;

\path [draw=color5, fill=color5, opacity=0.1, line width=1pt]
(axis cs:-0.301029995663981,0.917123753945531)
--(axis cs:-0.301029995663981,0.912023492427497)
--(axis cs:0,0.912988760594812)
--(axis cs:0.698970004336019,0.914128233172847)
--(axis cs:1.17609125905568,0.913844596225145)
--(axis cs:1.47712125471966,0.913696793605204)
--(axis cs:1.77815125038364,0.913836881164642)
--(axis cs:2.47712125471966,0.914051704314578)
--(axis cs:2.95424250943932,0.914554168284444)
--(axis cs:3.55630250076729,0.914544969225439)
--(axis cs:4.15836249209525,0.914544969225438)
--(axis cs:4.15836249209525,0.918531545095865)
--(axis cs:4.15836249209525,0.918531545095865)
--(axis cs:3.55630250076729,0.918531545095865)
--(axis cs:2.95424250943932,0.919257594601879)
--(axis cs:2.47712125471966,0.918979481932864)
--(axis cs:1.77815125038364,0.91893347553902)
--(axis cs:1.47712125471966,0.918975195499559)
--(axis cs:1.17609125905568,0.918355227710067)
--(axis cs:0.698970004336019,0.918346742101639)
--(axis cs:0,0.91811893117754)
--(axis cs:-0.301029995663981,0.917123753945531)
--cycle;

\path [draw=color6, fill=color6, opacity=0.1, line width=1pt]
(axis cs:2.47712125471966,0.921030340347165)
--(axis cs:2.47712125471966,0.78397041577221)
--(axis cs:2.95424250943932,0.914602584153709)
--(axis cs:3.55630250076729,0.915846302154584)
--(axis cs:3.55630250076729,0.92173286975299)
--(axis cs:3.55630250076729,0.92173286975299)
--(axis cs:2.95424250943932,0.919351074760211)
--(axis cs:2.47712125471966,0.921030340347165)
--cycle;

\path [draw=white!58.0392156862745!black, fill=white!58.0392156862745!black, opacity=0.1, line width=1pt]
(axis cs:2.47712125471966,0.920836597012775)
--(axis cs:2.47712125471966,0.915699347779978)
--(axis cs:2.95424250943932,0.918970155014227)
--(axis cs:3.55630250076729,0.920512196290061)
--(axis cs:3.55630250076729,0.926250467656733)
--(axis cs:3.55630250076729,0.926250467656733)
--(axis cs:2.95424250943932,0.924816856374335)
--(axis cs:2.47712125471966,0.920836597012775)
--cycle;

\path [draw=color7, fill=color7, opacity=0.1, line width=1pt]
(axis cs:-0.301029995663981,0.891715915267603)
--(axis cs:-0.301029995663981,0.879641369418268)
--(axis cs:0,0.879708937300195)
--(axis cs:0.698970004336019,0.879890351007487)
--(axis cs:1.17609125905568,0.879888713969838)
--(axis cs:1.47712125471966,0.879888713969838)
--(axis cs:1.77815125038364,0.879753836575239)
--(axis cs:2.47712125471966,0.879514044058666)
--(axis cs:2.95424250943932,0.879753836575239)
--(axis cs:3.55630250076729,0.879885342034973)
--(axis cs:4.15836249209525,0.880023591364437)
--(axis cs:4.15836249209525,0.892075653699615)
--(axis cs:4.15836249209525,0.892075653699615)
--(axis cs:3.55630250076729,0.891726106628322)
--(axis cs:2.95424250943932,0.892075653699615)
--(axis cs:2.47712125471966,0.892075653699615)
--(axis cs:1.77815125038364,0.892210531094214)
--(axis cs:1.47712125471966,0.892210531094214)
--(axis cs:1.17609125905568,0.892210531094214)
--(axis cs:0.698970004336019,0.891873024311176)
--(axis cs:0,0.892154899779798)
--(axis cs:-0.301029995663981,0.891715915267603)
--cycle;

\path [draw=color8, fill=color8, opacity=0.1, line width=1pt]
(axis cs:2.47712125471966,0.919799968897112)
--(axis cs:2.47712125471966,0.913832264545252)
--(axis cs:2.95424250943932,0.913218167272791)
--(axis cs:3.55630250076729,0.914737456826579)
--(axis cs:3.55630250076729,0.918112420924641)
--(axis cs:3.55630250076729,0.918112420924641)
--(axis cs:2.95424250943932,0.920391100773614)
--(axis cs:2.47712125471966,0.919799968897112)
--cycle;

\addplot [line width=1.5pt, color0, mystyle]
table {%
1.77815125038364 0.928264620998406
2.47712125471966 0.930171999225587
2.95424250943932 0.929924709103303
3.55630250076729 0.928987462055996
};
\addplot [line width=1.5pt, color1, , mystyle]
table {%
1.77815125038364 0.926758741556529
2.47712125471966 0.928481551643642
2.95424250943932 0.928395117120304
3.55630250076729 0.929668812423112
};
\addplot [line width=1.5pt, color2, , mystyle]
table {%
0.698970004336019 0.885970018027614
1.17609125905568 0.910082814724813
1.47712125471966 0.918285807212785
1.77815125038364 0.919136632093855
2.47712125471966 0.917881620748201
2.95424250943932 0.920201899821771
3.55630250076729 0.919578772245967
};
\addplot [line width=1.5pt, color4, , mystyle]
table {%
1.77815125038364 0.908220706457505
2.47712125471966 0.922532334275284
2.95424250943932 0.923212021965333
3.55630250076729 0.924115174877871
};
\addplot [line width=1.5pt,color3, dashed, mystyle, mark size=5, mark=mystar]
table {%
0 0.933527920918235
};

\addplot[line width=1.5pt, color3, dashed, no marks]
table {
-0.259637310505756 0.933527920918235
4.15836249209525 0.933527920918235
};
\addplot [line width=1.5pt, color5, , mystyle]
table {%
-0.301029995663981 0.914874314683732
0 0.915748260075884
0.698970004336019 0.916342161061946
1.17609125905568 0.916316610171142
1.47712125471966 0.916334893407843
1.77815125038364 0.916436985372796
2.47712125471966 0.916505972822801
2.95424250943932 0.917169810815256
3.55630250076729 0.916653451056247
4.15836249209525 0.916653451056247
};
\addplot [line width=1.5pt, color6, , mystyle]
table {%
2.47712125471966 0.873951134521789
2.95424250943932 0.917131742144791
3.55630250076729 0.918819623218716
};
\addplot [line width=1.5pt, white!58.0392156862745!black, , mystyle]
table {%
2.47712125471966 0.918634815718808
2.95424250943932 0.922078318687442
3.55630250076729 0.923618474066103
};
\addplot [line width=1.5pt, color7, , mystyle]
table {%
-0.301029995663981 0.886783756594055
0 0.886836278154124
0.698970004336019 0.886986162450209
1.17609125905568 0.886981025042502
1.47712125471966 0.886981025042502
1.77815125038364 0.886981025042502
2.47712125471966 0.886981025042502
2.95424250943932 0.886981025042502
3.55630250076729 0.886981025042502
4.15836249209525 0.886981025042502
};
\addplot [line width=1.5pt, color8, , mystyle]
table {%
2.47712125471966 0.917255223599265
2.95424250943932 0.917176008543259
3.55630250076729 0.91656542462931
};
\end{axis}

\end{tikzpicture}
         %\label{fig:aucovertime}
     \end{subfigure}
     \begin{subfigure}[t]{0.32\textwidth}
         \centering
         %\hspace*{-0.2cm}
         % This file was created with tikzplotlib v0.9.12.
\begin{tikzpicture}

\begin{axis}[
height=4.5cm,
legend style={fill opacity=0.8, draw opacity=1, text opacity=1, draw=none},
minor xtick={},
minor ytick={},
tick align=outside,
tick pos=left,
width=\textwidth,
x grid style={white!69.0196078431373!black},
xlabel={Given Time Budget},
xmin=0, xmax=3.55630250076729,
xtick style={color=black},
xtick={0,0.698970004336019,1.47712125471966,2.47712125471966,3.55630250076729},
xticklabels={1s,5s,30s,5min,1h},
y grid style={white!69.0196078431373!black},
ylabel={ROC AUC Wins},
ymin=0, ymax=15,
ytick style={color=black},
ytick={0,2,4,6,8,10,12,14,16},
mystylepalette
]
\path [draw=color0, fill=color0, opacity=0.1, line width=1pt]
(axis cs:1.77815125038364,3.8)
--(axis cs:1.77815125038364,1)
--(axis cs:2.47712125471966,2)
--(axis cs:2.95424250943932,1.2)
--(axis cs:3.55630250076729,0.6)
--(axis cs:3.55630250076729,2)
--(axis cs:3.55630250076729,2)
--(axis cs:2.95424250943932,2.4)
--(axis cs:2.47712125471966,3.4)
--(axis cs:1.77815125038364,3.8)
--cycle;

\path [draw=color1, fill=color1, opacity=0.1, line width=1pt]
(axis cs:1.77815125038364,3.8)
--(axis cs:1.77815125038364,1.8)
--(axis cs:2.47712125471966,2.2)
--(axis cs:2.95424250943932,4)
--(axis cs:3.55630250076729,3.4)
--(axis cs:3.55630250076729,4.6)
--(axis cs:3.55630250076729,4.6)
--(axis cs:2.95424250943932,4.6)
--(axis cs:2.47712125471966,3.8)
--(axis cs:1.77815125038364,3.8)
--cycle;

\path [draw=color2, fill=color2, opacity=0.1, line width=1pt]
(axis cs:0.698970004336019,1.4)
--(axis cs:0.698970004336019,0)
--(axis cs:1.17609125905568,0)
--(axis cs:1.47712125471966,0)
--(axis cs:1.77815125038364,0)
--(axis cs:2.47712125471966,0)
--(axis cs:2.95424250943932,0)
--(axis cs:3.55630250076729,0)
--(axis cs:3.55630250076729,0)
--(axis cs:3.55630250076729,0)
--(axis cs:2.95424250943932,0.6)
--(axis cs:2.47712125471966,0.6)
--(axis cs:1.77815125038364,1.4)
--(axis cs:1.47712125471966,1.2)
--(axis cs:1.17609125905568,0.8)
--(axis cs:0.698970004336019,1.4)
--cycle;

\path [draw=color4, fill=color4, opacity=0.1, line width=1pt]
(axis cs:1.77815125038364,1.6)
--(axis cs:1.77815125038364,0.4)
--(axis cs:2.47712125471966,0.2)
--(axis cs:2.95424250943932,0)
--(axis cs:3.55630250076729,0)
--(axis cs:3.55630250076729,0.6)
--(axis cs:3.55630250076729,0.6)
--(axis cs:2.95424250943932,0.8)
--(axis cs:2.47712125471966,1.4)
--(axis cs:1.77815125038364,1.6)
--cycle;

\path [draw=color3, fill=color3, opacity=0.1, line width=1pt]
(axis cs:0,14.6)
--(axis cs:0,12.2)
--(axis cs:0.698970004336019,12.2)
--(axis cs:1.17609125905568,8.4)
--(axis cs:1.47712125471966,9.8)
--(axis cs:1.77815125038364,5.2)
--(axis cs:2.47712125471966,2.6)
--(axis cs:2.95424250943932,3.4)
--(axis cs:3.55630250076729,4.2)
--(axis cs:3.55630250076729,6.6)
--(axis cs:3.55630250076729,6.6)
--(axis cs:2.95424250943932,5.4)
--(axis cs:2.47712125471966,4.8)
--(axis cs:1.77815125038364,6)
--(axis cs:1.47712125471966,11)
--(axis cs:1.17609125905568,11.6)
--(axis cs:0.698970004336019,14.6)
--(axis cs:0,14.6)
--cycle;

\path [draw=color5, fill=color5, opacity=0.1, line width=1pt]
(axis cs:0,1.8)
--(axis cs:0,0.6)
--(axis cs:0.698970004336019,0.2)
--(axis cs:1.17609125905568,0.2)
--(axis cs:1.47712125471966,0.2)
--(axis cs:1.77815125038364,0.2)
--(axis cs:2.47712125471966,0)
--(axis cs:2.95424250943932,0)
--(axis cs:3.55630250076729,0)
--(axis cs:4.15836249209525,0)
--(axis cs:4.15836249209525,1.2)
--(axis cs:4.15836249209525,1.2)
--(axis cs:3.55630250076729,1.2)
--(axis cs:2.95424250943932,1.8)
--(axis cs:2.47712125471966,1.2)
--(axis cs:1.77815125038364,2.2)
--(axis cs:1.47712125471966,2.2)
--(axis cs:1.17609125905568,2.2)
--(axis cs:0.698970004336019,2.6)
--(axis cs:0,1.8)
--cycle;

\path [draw=color6, fill=color6, opacity=0.1, line width=1pt]
(axis cs:2.47712125471966,0.8)
--(axis cs:2.47712125471966,0)
--(axis cs:2.95424250943932,0)
--(axis cs:3.55630250076729,0)
--(axis cs:3.55630250076729,0.8)
--(axis cs:3.55630250076729,0.8)
--(axis cs:2.95424250943932,0)
--(axis cs:2.47712125471966,0.8)
--cycle;

\path [draw=white!58.0392156862745!black, fill=white!58.0392156862745!black, opacity=0.1, line width=1pt]
(axis cs:2.47712125471966,1.4)
--(axis cs:2.47712125471966,0.2)
--(axis cs:2.95424250943932,0.2)
--(axis cs:3.55630250076729,0.4)
--(axis cs:3.55630250076729,1)
--(axis cs:3.55630250076729,1)
--(axis cs:2.95424250943932,1.6)
--(axis cs:2.47712125471966,1.4)
--cycle;

\path [draw=color7, fill=color7, opacity=0.1, line width=1pt]
(axis cs:0,0.6)
--(axis cs:0,0)
--(axis cs:0.698970004336019,0)
--(axis cs:1.17609125905568,0)
--(axis cs:1.47712125471966,0)
--(axis cs:1.77815125038364,0)
--(axis cs:2.47712125471966,0)
--(axis cs:2.95424250943932,0)
--(axis cs:3.55630250076729,0)
--(axis cs:4.15836249209525,0)
--(axis cs:4.15836249209525,0)
--(axis cs:4.15836249209525,0)
--(axis cs:3.55630250076729,0)
--(axis cs:2.95424250943932,0)
--(axis cs:2.47712125471966,0)
--(axis cs:1.77815125038364,0)
--(axis cs:1.47712125471966,0)
--(axis cs:1.17609125905568,0)
--(axis cs:0.698970004336019,0)
--(axis cs:0,0.6)
--cycle;

\path [draw=color8, fill=color8, opacity=0.1, line width=1pt]
(axis cs:2.47712125471966,0.8)
--(axis cs:2.47712125471966,0)
--(axis cs:2.95424250943932,0.2)
--(axis cs:3.55630250076729,0)
--(axis cs:3.55630250076729,0.8)
--(axis cs:3.55630250076729,0.8)
--(axis cs:2.95424250943932,1.4)
--(axis cs:2.47712125471966,0.8)
--cycle;

\addplot [line width=1.5pt, color0, mystyle]
table {%
1.77815125038364 2.2
2.47712125471966 2.6
2.95424250943932 1.8
3.55630250076729 1.4
};
\addplot [line width=1.5pt, color1, mystyle]
table {%
1.77815125038364 2.8
2.47712125471966 3
2.95424250943932 4.2
3.55630250076729 4
};
\addplot [line width=1.5pt, color2, mystyle]
table {%
0.698970004336019 0.6
1.17609125905568 0.4
1.47712125471966 0.4
1.77815125038364 0.6
2.47712125471966 0.2
2.95424250943932 0.2
3.55630250076729 0
};
\addplot [line width=1.5pt, color4, mystyle]
table {%
1.77815125038364 1
2.47712125471966 0.8
2.95424250943932 0.4
3.55630250076729 0.2
};
\addplot [line width=1.5pt, color3, mystyle]
table {%
0 13.4
0.698970004336019 13.2
1.17609125905568 10
1.47712125471966 10.6
1.77815125038364 5.6
2.47712125471966 3.8
2.95424250943932 4.4
3.55630250076729 5.2
};
\addplot [line width=1.5pt, color5, mystyle]
table {%
0 1.2
0.698970004336019 1.2
1.17609125905568 1.2
1.47712125471966 1.2
1.77815125038364 1.2
2.47712125471966 0.6
2.95424250943932 0.6
3.55630250076729 0.4
4.15836249209525 0.4
};
\addplot [line width=1.5pt, color6, mystyle]
table {%
2.47712125471966 0.4
2.95424250943932 0
3.55630250076729 0.4
};
\addplot [line width=1.5pt, white!58.0392156862745!black, mystyle]
table {%
2.47712125471966 0.8
2.95424250943932 0.8
3.55630250076729 0.8
};
\addplot [line width=1.5pt, color7, mystyle]
table {%
0 0.2
0.698970004336019 0
1.17609125905568 0
1.47712125471966 0
1.77815125038364 0
2.47712125471966 0
2.95424250943932 0
3.55630250076729 0
4.15836249209525 0
};
\addplot [line width=1.5pt, color8, mystyle]
table {%
2.47712125471966 0.4
2.95424250943932 0.8
3.55630250076729 0.4
};
\end{axis}

\end{tikzpicture}
         %\label{fig:winsovertime}
     \end{subfigure}
     \begin{subfigure}[t]{0.32\textwidth}
         \centering
         %\hspace*{1.1cm}
         % This file was created with tikzplotlib v0.9.12.
\begin{tikzpicture}

\begin{axis}[
height=4.5cm,
legend style={fill opacity=0.8, draw opacity=1, text opacity=1, draw=none},
minor xtick={},
minor ytick={},
tick align=outside,
tick pos=left,
width=\textwidth,
x grid style={white!69.0196078431373!black},
xlabel={Given Time Budget},
xmin=0, xmax=3.55630250076729,
xtick style={color=black},
xtick={0,0.698970004336019,1.47712125471966,2.47712125471966,3.55630250076729},
xticklabels={1s,5s,30s,5min,1h},
y grid style={white!69.0196078431373!black},
ylabel={Mean ROC AUC Rank},
ymin=1, ymax=10,
ytick style={color=black},
ytick={1,2,3,4,5,6,7,8,9,10},
mystylepalette
]
\path [draw=color0, fill=color0, opacity=0.1, line width=1pt]
(axis cs:1.77815125038364,3.37777777777778)
--(axis cs:1.77815125038364,2.91111111111111)
--(axis cs:2.47712125471966,3.16097222222222)
--(axis cs:2.95424250943932,2.97222222222222)
--(axis cs:3.55630250076729,3.71111111111111)
--(axis cs:3.55630250076729,3.91111111111111)
--(axis cs:3.55630250076729,3.91111111111111)
--(axis cs:2.95424250943932,3.88888888888889)
--(axis cs:2.47712125471966,3.91111111111111)
--(axis cs:1.77815125038364,3.37777777777778)
--cycle;

\path [draw=color1, fill=color1, opacity=0.1, line width=1pt]
(axis cs:1.77815125038364,3.23902777777778)
--(axis cs:1.77815125038364,2.89444444444444)
--(axis cs:2.47712125471966,3.28888888888889)
--(axis cs:2.95424250943932,3.45555555555556)
--(axis cs:3.55630250076729,3.34444444444444)
--(axis cs:3.55630250076729,3.88458333333333)
--(axis cs:3.55630250076729,3.88458333333333)
--(axis cs:2.95424250943932,3.99569444444444)
--(axis cs:2.47712125471966,4.08888888888889)
--(axis cs:1.77815125038364,3.23902777777778)
--cycle;

\path [draw=color2, fill=color2, opacity=0.1, line width=1pt]
(axis cs:0.698970004336019,2.92222222222222)
--(axis cs:0.698970004336019,2.71666666666667)
--(axis cs:1.17609125905568,3.42208333333333)
--(axis cs:1.47712125471966,3.34444444444444)
--(axis cs:1.77815125038364,4.61666666666667)
--(axis cs:2.47712125471966,6.30555555555556)
--(axis cs:2.95424250943932,6.03888888888889)
--(axis cs:3.55630250076729,6.17777777777778)
--(axis cs:3.55630250076729,6.92777777777778)
--(axis cs:3.55630250076729,6.92777777777778)
--(axis cs:2.95424250943932,6.72777777777778)
--(axis cs:2.47712125471966,6.77777777777778)
--(axis cs:1.77815125038364,4.92777777777778)
--(axis cs:1.47712125471966,3.78333333333333)
--(axis cs:1.17609125905568,3.58888888888889)
--(axis cs:0.698970004336019,2.92222222222222)
--cycle;

\path [draw=color4, fill=color4, opacity=0.1, line width=1pt]
(axis cs:1.77815125038364,4.45)
--(axis cs:1.77815125038364,3.85472222222222)
--(axis cs:2.47712125471966,5.21111111111111)
--(axis cs:2.95424250943932,5.29444444444444)
--(axis cs:3.55630250076729,5.58888888888889)
--(axis cs:3.55630250076729,5.85)
--(axis cs:3.55630250076729,5.85)
--(axis cs:2.95424250943932,6.02236111111111)
--(axis cs:2.47712125471966,5.77263888888889)
--(axis cs:1.77815125038364,4.45)
--cycle;

\path [draw=color3, fill=color3, opacity=0.1, line width=1pt]
(axis cs:0,1.28888888888889)
--(axis cs:0,1.20555555555556)
--(axis cs:0.698970004336019,1.2)
--(axis cs:1.17609125905568,1.39444444444444)
--(axis cs:1.47712125471966,1.5)
--(axis cs:1.77815125038364,2.01111111111111)
--(axis cs:2.47712125471966,2.52777777777778)
--(axis cs:2.95424250943932,2.71666666666667)
--(axis cs:3.55630250076729,2.61111111111111)
--(axis cs:3.55630250076729,3.03347222222222)
--(axis cs:3.55630250076729,3.03347222222222)
--(axis cs:2.95424250943932,3.02777777777778)
--(axis cs:2.47712125471966,2.78333333333333)
--(axis cs:1.77815125038364,2.23333333333333)
--(axis cs:1.47712125471966,1.66666666666667)
--(axis cs:1.17609125905568,1.63888888888889)
--(axis cs:0.698970004336019,1.32777777777778)
--(axis cs:0,1.28888888888889)
--cycle;

\path [draw=color5, fill=color5, opacity=0.1, line width=1pt]
(axis cs:0,1.98333333333333)
--(axis cs:0,1.87777777777778)
--(axis cs:0.698970004336019,2.21666666666667)
--(axis cs:1.17609125905568,2.77222222222222)
--(axis cs:1.47712125471966,3.15555555555556)
--(axis cs:1.77815125038364,4.08333333333333)
--(axis cs:2.47712125471966,5.83333333333333)
--(axis cs:2.95424250943932,5.94444444444444)
--(axis cs:3.55630250076729,5.99416666666667)
--(axis cs:4.15836249209525,5.98305555555555)
--(axis cs:4.15836249209525,6.61666666666667)
--(axis cs:4.15836249209525,6.61666666666667)
--(axis cs:3.55630250076729,6.65)
--(axis cs:2.95424250943932,6.54444444444444)
--(axis cs:2.47712125471966,6.58333333333333)
--(axis cs:1.77815125038364,4.59444444444444)
--(axis cs:1.47712125471966,3.42222222222222)
--(axis cs:1.17609125905568,3.00555555555556)
--(axis cs:0.698970004336019,2.4)
--(axis cs:0,1.98333333333333)
--cycle;

\path [draw=color6, fill=color6, opacity=0.1, line width=1pt]
(axis cs:2.47712125471966,6.71111111111111)
--(axis cs:2.47712125471966,5.32222222222222)
--(axis cs:2.95424250943932,5.86666666666667)
--(axis cs:3.55630250076729,5.53888888888889)
--(axis cs:3.55630250076729,6.12222222222222)
--(axis cs:3.55630250076729,6.12222222222222)
--(axis cs:2.95424250943932,6.39444444444444)
--(axis cs:2.47712125471966,6.71111111111111)
--cycle;

\path [draw=white!58.0392156862745!black, fill=white!58.0392156862745!black, opacity=0.1, line width=1pt]
(axis cs:2.47712125471966,6.07222222222222)
--(axis cs:2.47712125471966,5.37222222222222)
--(axis cs:2.95424250943932,4.85)
--(axis cs:3.55630250076729,4.7)
--(axis cs:3.55630250076729,5.15555555555556)
--(axis cs:3.55630250076729,5.15555555555556)
--(axis cs:2.95424250943932,5.55555555555556)
--(axis cs:2.47712125471966,6.07222222222222)
--cycle;

\path [draw=color7, fill=color7, opacity=0.1, line width=1pt]
(axis cs:0,2.85)
--(axis cs:0,2.8)
--(axis cs:0.698970004336019,3.56111111111111)
--(axis cs:1.17609125905568,4.46111111111111)
--(axis cs:1.47712125471966,4.99444444444444)
--(axis cs:1.77815125038364,6.21666666666667)
--(axis cs:2.47712125471966,8.71111111111111)
--(axis cs:2.95424250943932,8.90555555555556)
--(axis cs:3.55630250076729,8.90555555555556)
--(axis cs:4.15836249209525,8.91097222222222)
--(axis cs:4.15836249209525,9.37222222222222)
--(axis cs:4.15836249209525,9.37222222222222)
--(axis cs:3.55630250076729,9.37222222222222)
--(axis cs:2.95424250943932,9.33361111111111)
--(axis cs:2.47712125471966,9.22791666666667)
--(axis cs:1.77815125038364,6.55)
--(axis cs:1.47712125471966,5.17777777777778)
--(axis cs:1.17609125905568,4.5725)
--(axis cs:0.698970004336019,3.67222222222222)
--(axis cs:0,2.85)
--cycle;

\path [draw=color8, fill=color8, opacity=0.1, line width=1pt]
(axis cs:2.47712125471966,6.48888888888889)
--(axis cs:2.47712125471966,6.00513888888889)
--(axis cs:2.95424250943932,5.95)
--(axis cs:3.55630250076729,6.07222222222222)
--(axis cs:3.55630250076729,6.46111111111111)
--(axis cs:3.55630250076729,6.46111111111111)
--(axis cs:2.95424250943932,6.41666666666667)
--(axis cs:2.47712125471966,6.48888888888889)
--cycle;

\addplot [line width=1.5pt, color0, mystyle]
table {%
1.77815125038364 3.17222222222222
2.47712125471966 3.50555555555556
2.95424250943932 3.39444444444444
3.55630250076729 3.81666666666667
};
\addplot [line width=1.5pt, color1, mystyle]
table {%
1.77815125038364 3.06666666666667
2.47712125471966 3.67777777777778
2.95424250943932 3.73333333333333
3.55630250076729 3.60555555555556
};
\addplot [line width=1.5pt, color2, mystyle]
table {%
0.698970004336019 2.8
1.17609125905568 3.51111111111111
1.47712125471966 3.57777777777778
1.77815125038364 4.76111111111111
2.47712125471966 6.51111111111111
2.95424250943932 6.40555555555556
3.55630250076729 6.57777777777778
};
\addplot [line width=1.5pt, color4, mystyle]
table {%
1.77815125038364 4.15
2.47712125471966 5.50555555555556
2.95424250943932 5.66111111111111
3.55630250076729 5.69444444444444
};
\addplot [line width=1.5pt, color3, mystyle]
table {%
0 1.23888888888889
0.698970004336019 1.26666666666667
1.17609125905568 1.51666666666667
1.47712125471966 1.58333333333333
1.77815125038364 2.12222222222222
2.47712125471966 2.65555555555556
2.95424250943932 2.9
3.55630250076729 2.82777777777778
};
\addplot [line width=1.5pt, color5, mystyle]
table {%
0 1.93333333333333
0.698970004336019 2.32222222222222
1.17609125905568 2.9
1.47712125471966 3.3
1.77815125038364 4.34444444444444
2.47712125471966 6.21111111111111
2.95424250943932 6.25
3.55630250076729 6.28333333333333
4.15836249209525 6.28333333333333
};
\addplot [line width=1.5pt, color6, mystyle]
table {%
2.47712125471966 6.00555555555556
2.95424250943932 6.11111111111111
3.55630250076729 5.83888888888889
};
\addplot [line width=1.5pt, white!58.0392156862745!black, mystyle]
table {%
2.47712125471966 5.72222222222222
2.95424250943932 5.28333333333333
3.55630250076729 4.92777777777778
};
\addplot [line width=1.5pt, color7, mystyle]
table {%
0 2.82777777777778
0.698970004336019 3.61111111111111
1.17609125905568 4.51666666666667
1.47712125471966 5.07777777777778
1.77815125038364 6.38333333333333
2.47712125471966 8.96666666666667
2.95424250943932 9.13888888888889
3.55630250076729 9.13333333333333
4.15836249209525 9.13333333333333
};
\addplot [line width=1.5pt, color8, mystyle]
table {%
2.47712125471966 6.23888888888889
2.95424250943932 6.12222222222222
3.55630250076729 6.29444444444444
};
\end{axis}

\end{tikzpicture}
         %\label{fig:ranksovertime}
     \end{subfigure}
     \vspace{-.6cm}
     \begin{subfigure}[b]{1.0\textwidth}
         \centering
         % This file was created with tikzplotlib v0.9.12.
\begin{tikzpicture}
\begin{axis}[
hide axis,
width=1.\textwidth,
height=.2\textwidth,
legend columns=4,legend style={draw=none,column sep=5ex},
legend cell align={left},
legend style={fill opacity=0.8, draw opacity=1, text opacity=1, draw=white!80!black, /tikz/every even column/.append style={column sep=0.1cm}},
log basis x={10},
tick align=outside,
tick pos=left,
x grid style={white!69.0196078431373!black},
xlabel={Training time per task},
xmin=0.005, xmax=300,
xmode=log,
xtick style={color=black},
y grid style={white!69.0196078431373!black},
ylabel={Negative Log Likelihood},
ymin=-1, ymax=15,
ytick style={color=black}
]

\addlegendimage{ultra thick, color3}
\addlegendentry{\methodname{}}
\addlegendimage{ultra thick, color0}
\addlegendentry{Auto-sklearn 2.0}
\addlegendimage{ultra thick, color1}
\addlegendentry{Autogluon}
\addlegendimage{ultra thick, color2}
\addlegendentry{LightGBM}
\addlegendimage{ultra thick, color4}
\addlegendentry{XGBoost}
\addlegendimage{ultra thick, color5}
\addlegendentry{LogReg}
\addlegendimage{ultra thick, color6}
\addlegendentry{Cocktail}
\addlegendimage{ultra thick, white!49.8039215686275!black}
\addlegendentry{Catboost}
\addlegendimage{ultra thick, color7}
\addlegendentry{KNN}
\addlegendimage{ultra thick, color8}
\addlegendentry{SAINT}
\end{axis}

\end{tikzpicture}
         %\label{fig:legendovertime}
     \end{subfigure}
     \vspace{0cm}
     \caption{ROC AUC as a function of the time allowed to train \& tune methods, on 18 numerical datasets from the \openmlcc{} Benchmark. We report the mean, mean wins and rank and 95\% confidence interval across $5$ splits for increasing budgets. 
     %(Unlabelled ticks: 1min, 15min). 
     The red star indicates performance of \methodname{} with 32 permutations (which requires 0.62s on GPU). We report more results in Table~\ref{table:tabular_results_table_small}.
     }
%     \vspace*{-0.5cm}
     \label{fig:perfovertime}
\end{figure}
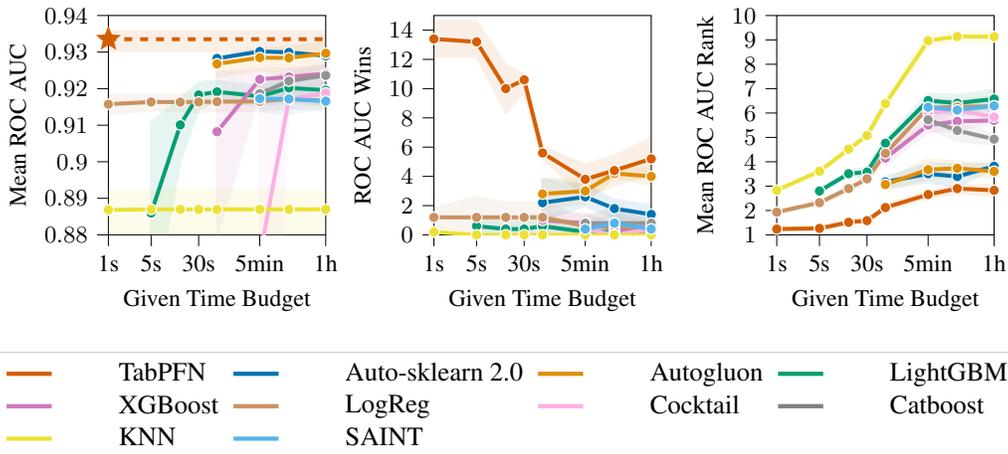

\slimparagraph{Baselines} We compare against five standard ML methods and two state-of-the-art AutoML systems for tabular data.
As ML models we considered two simple and fast baselines, \textit{K-nearest-neighbors (KNN)} and \textit{Logistic Regression (LogReg)}. Additionally, we considered
%\textit{Gaussian Processes (GPs)}~\citep{rasmussen-book06a} and 
three popular tree-based boosting methods, \textit{XGBoost}~\citep{chen-arxiv20a}, \textit{LightGBM}~\citep{ke-neurips17a} and \textit{CatBoost}~\citep{prokhorenkova-neurips18a}.
For each ML model, we used $5$-fold cross-validation to evaluate randomly drawn hyperparameter configurations until a given budget was exhausted or $10\,000$ configurations were evaluated (for the search spaces, see Appendix~\ref{app:ssec:spaces}).
We then chose the best-performing hyperparameter configuration (maximum ROC AUC OVO) and refit on the whole training set. Where necessary, we imputed missing values with the mean, one-hot or ordinal encoded categorical inputs, normalized features and passed categorical feature indicators.
As more complex but powerful baselines, we chose two state-of-the-art AutoML systems: \textit{AutoGluon}~\citep{erickson-arxiv20a}, which combines ML models including neural networks and tree-based models into a stacked ensemble, and \textit{Auto-sklearn 2.0}~\citep{feurer-nips15a,feurer-arxiv20v2} which uses Bayesian Optimization and combines the evaluated models into a weighted ensemble.%and \textit{LightAutoML}~\citep{DBLP:journals/corr/abs-2109-01528}, which we optimize for fast training by restricting models to \textit{LightGBM}~\citep{ke-neurips17a} and linear models.
\footnote{Auto-sklearn 2.0 optimizes ROC AUC for binary classification and cross-entropy for multi-class classification (as multi-class ROC AUC is not implemented).
%otherwise, AutoGluon failed to completely fit on $N$ splits, diminishing its performance across the entire Benchmark (Results of that run can be found in Appendix \ref{app:addresults}).
}
We note that previous works have found that DL baselines do not outperform or match the performance of GBDT or AutoML methods for small to medium-sized tabular data (< 10,000 samples; while matching GBDT performance on larger datasets) \citep{nnsfortabular,grinsztajn2022why,nnsfortabular2}. Furthermore, DL methods such as TabNet, SAINT, Regularization Cocktails, Non-parametric Transformers \citep{arik2021tabnet, somepalli2021saint, regcocktail, kossen2021self} are evaluated on much larger datasets and often use custom parameter tuning and preprocessing.
We still evaluate two prominent DL methods: \textit{Regularization Cocktails}\footnote{The implementation of Regularization Cocktails does not currently support optimization for multi-class AUC ROC. We optimized for cross-entropy, which performed better than balanced accuracy (the default).}~\citep{regcocktail} and \textit{SAINT}~\citep{somepalli2021saint}.
% TODO: We set both AutoML systems to optimize for ROC AUC.

\slimparagraph{Evaluation Protocol} For each dataset and method, we evaluated $5$ repetitions, each with a different random seed and train- and test split (50\% train and 50\% test samples; all methods used the same split given a seed). To aggregate results across datasets, we report the ROC AUC (one-vs-one (OVO) for multi-class classification) average, ranks and wins including the $95\%$ confidence interval and compare to the performance of the baselines with a budget of $\{30, 60, 300, 900, 3600\}$ seconds.\footnote{When comparing methods to each other for a given time budget in Figure \ref{fig:perfovertime}, we drop methods that take more than 200\% of the requested time budget; some methods also do not use their full budget.} Our \methodname{} uses $32$ data permutations for ensembling as described in Section \ref{sec:tabpfn}; we also evaluate \methodname{} without permutations, which we label ``\methodname{}$_{n.e.}$'' in Table \ref{table:tabular_results_table_small}.% and all permutations.
%------------------------------------------------------
%\label{ssec:expresults}
%------------------------------------------------------
%
\begin{table}[t]
    \centering
    \smaller
    \begin{tabular}{@{\hskip 0mm}
l@{\hskip 1mm}
l@{\hskip 1mm}
l@{\hskip 1mm}
l@{\hskip 1mm}
l@{\hskip 1mm}
l@{\hskip 1mm}
l@{\hskip 1mm}
l@{\hskip 1mm}
l@{\hskip 0mm}}
\toprule
{} &        LightGBM &        CatBoost &         XGBoost &          ASKL2.0 &        AutoGluon &  TabPFN$_{n.e.}$ &                    TabPFN &        TabPFN + AutoGluon \\
%\midrule
%Wins AUC OVO
%                                                       &               0 &               0 &               0 &                0 &                1 &                3 &                \textbf{4} &                         3 \\
%Wins Acc.                        &               0 &               1 &               1 &                0 &                1 &                0 &                \textbf{6} &                         4 \\
%Wins CE                          &               0 &               1 &               0 &                1 &                3 &                1 &                         3 &                \textbf{7} \\
\midrule
M. rank AUC OVO
                                                  &          6.9722 &          4.9444 &          6.1944 &           4.4722 &                4 &           3.8056 &                    2.9444 &           \textbf{2.6667} \\
Mean rank Acc.                   &          6.8889 &          4.9722 &          6.0556 &           5.1667 &           3.8889 &           3.8889 &                    2.8889 &             \textbf{2.25} \\
Mean rank CE                     &          5.7778 &          5.4444 &               6 &           6.4167 &           3.1111 &           4.1389 &                    3.0278 &           \textbf{2.0833} \\
\midrule
Mean AUC OVO
                                                     &   0.92$\pm$.013 &  0.924$\pm$.011 &   0.924$\pm$.01 &  0.929$\pm$.0096 &   0.93$\pm$.0091 &  0.932$\pm$.0088 &  \textbf{0.934}$\pm$.0086 &  \textbf{0.934}$\pm$.0084 \\
Mean Acc.                        &  0.862$\pm$.012 &  0.864$\pm$.011 &  0.866$\pm$.011 &    0.87$\pm$.014 &    0.881$\pm$.01 &  0.873$\pm$.0095 &           0.879$\pm$.0089 &  \textbf{0.886}$\pm$.0094 \\
Mean CE                          &   0.75$\pm$.039 &  0.747$\pm$.029 &   0.759$\pm$.04 &   0.813$\pm$.073 &   0.714$\pm$.014 &   0.727$\pm$.021 &            0.716$\pm$.019 &   \textbf{0.711}$\pm$.014 \\
\midrule
Mean time (s) 
                                    &            \multirow{2}{*}{3280} &            \multirow{2}{*}{3746} &            \multirow{2}{*}{3364} &            \multirow{2}{*}{3601} &                    \multirow{2}{*}{3077} &  \textbf{1.301} (CPU) &           37.59 (CPU) &                      3109 (CPU) \\
(Tune + Train + Predict)                                                                        &             &             &             &             &                     &  \textbf{0.0519} (GPU) &           0.6172 (GPU) &                      3077 (GPU) \\
\bottomrule
\end{tabular}
    \caption{Results on the 18 numerical datasets in \openmlcc{} for 60 minutes requested time per data-split. $\pm$ values indicate a metric's standard deviation. If available, all baselines optimize ROC AUC.
    %Each method got a time budget of 150 hours, but not all methods used the full budget. 
    TabPFN n.e. is a faster version of our method, without ensembling, while TabPFN + AutoGluon is an ensemble of TabPFN and AutoGluon. Separate inference times in Table \ref{app:table:tabular_results_table}.}
    \label{table:tabular_results_table_small}
\end{table}

\slimparagraph{Results}
We now present results for the 18 purely numerical datasets without missing values, detailed in Figure~\ref{fig:perfovertime} and Table \ref{table:tabular_results_table_small}. 
%the achieved AUC ROC as a function of budget in Figure~\ref{fig:perfovertime},
% and the final performance wrt.\ different metrics in Table~\ref{tab:tabular_results_table_small}. 
%
%Considering performance-over-time (Figure~\ref{fig:perfovertime}), we
Figure \ref{fig:perfovertime} shows that \methodname{} achieves a dramatically better tradeoff of accuracy and training speed than all the other methods: it makes predictions within less than a second on one GPU that tie with the performance of the best competitors (the AutoML systems) after training one hour, and that dominate the performance of tuned GBDT methods.
Unsurprisingly, the simple baselines (\textit{LogReg}, \textit{KNN}) already yield results with a small budget but perform worst overall.
%(see average rank). 
GBDTs (\textit{XGBoost}, \textit{CatBoost}, \textit{LightGBM}) perform better but are still outperformed by TabPFN and the state-of-the-art AutoML systems (\textit{Auto-sklearn 2.0}, \textit{Autogluon}).
%We also performed statistical tests, showing that \methodname{} statistically significantly outperforms the tuned baselines; it also performs slightly, but insignificantly, better than the AutoML systems (see Figure \ref{fig:cd_plots}).
%Furthermore, already the fastest version of \methodname{} with just a single forward pass, \methodname{}$_{n.e.}$, statistically significantly outperforms all default baselines (see Figure \ref{fig:cd_plots}).

% TODO REDO CALCULATIONS BECAUSE OF NUMERICAL
\methodname{} is much faster than methods with comparable performance.
In the following, we compare the combined times for training and prediction (and tuning if applicable).
\methodname{}$_{n.e.}$ requires 1.30s on a CPU and 0.05s on a GPU on average to predict for one dataset\footnote{This includes overhead from a conservative scikit-learn interface. We assume that our trained model is in (GPU) memory already, which otherwise required an additional 0.2s for us.}, performing comparably to the strongest baselines at five minutes; it thus yields a $230\times$ speedup on CPU and a $5\,700\times$ speedup using a GPU.
We ignore computational development costs of each method, see Appendix \ref{app:time_comparison_discussion} for why.

We would like to emphasize that the discussed results are \emph{aggregate} results across datasets, and that no classification method, including TabPFN, performs best on all individual datasets. Indeed, there do exist datasets for which TabPFN is outperformed even by default baselines.
Generally, \methodname{} is less strong when categorical features or missing values are present. Appendix \ref{app:sec:detailed_results} presents results for all 30 test datasets from the \openmlcc{} Benchmark, including 12 with categorical features and/or missing values. %while \methodname{} is still a strong classification algorithm for these datasets, it does not perform as well as for the purely numerical case without missing values.
Appendix~\ref{app:sec:larger_results} presents a more detailed overview of the results for different kinds of datasets, including on an additional 149 validation datasets from OpenML, confirming \methodname{}'s strong performance for purely numerical datasets. We also show per dataset results for each of the 179 datasets (our 30 test datasets and the 149 validation datasets); these demonstrate that, while TabPFN generally does better for datasets with numerical features, there also exist several purely numerical datasets for which \methodname{} does \emph{not} perform better than the baselines, as well as categorical datasets for which it \emph{does} clearly perform best. %We also provide violin plots at \url{https://github.com/automl/TabPFNResults/blob/main/violin_plots.pdf} to compare TabPFN against each individual other classifier across datasets.

\slimparagraph{\openmlautoml{} Benchmark} 
Additionally, we evaluated our \methodname{} on the small ($\le$ 1\,000 training samples, 100 features, 10 classes) datasets of the \openmlautoml{} Benchmark, for which externally-validated performance numbers are available.  
%(due to the 10-fold cross-validation in the \openmlautoml{} Benchmark this is identical to the setup above, where we used up to 2\,000 samples split 50-50 into training and test).
%
We use the setup provided by the \openmlautoml{} Benchmark,~i.e.: metrics, official evaluation scripts and pre-released evaluations for an even wider range of AutoML baselines. Using only an average of $4.4$ seconds per dataset on a single CPU, compared to 60 minutes for the baselines, \methodname{} outperformed all baselines in terms of mean cross-entropy, accuracy and the OpenML Metric \footnote{The OpenML metric evaluates binary classification using ROC AUC and multiclass using Cross Entropy.}. We provide detailed results in Appendix \ref{app:sec:automl_results}. % [TODO: @Katha]

\slimparagraph{In-depth analysis of \methodname{} predictions} We evaluated our model predictions in a plethora of ways to provide extra insights, such as the inductive biases of our method, see Appendix \ref{sec:app:in_depth}. We confirm that \methodname{} learns to make predictions biased towards simple causal explanations, as detailed in \ref{sec:app:inductive_bias}, while GBDT methods do not share this inductive bias.
We also evaluate our method\textquotesingle{}s invariance to feature rotations and robustness to uninformative features in Appendix \ref{sec:app:robustness_uninformative} and \ref{sec:app:rotation_invariance}.
In Figure \ref{fig:results_dataset_characteristics}, we observe that \methodname{} is especially strong compared to baselines when datasets do not contain categorical features or missing values.

\slimparagraph{Ensembling} We observe that \methodname{} performs best on different datasets than our baselines, i.e., the per-dataset correlation of normalized ROC AUC scores between \methodname{} and the strong baselines is lower than between the strong baselines, visualized in Figure \ref{fig:model correlation} in our Appendix.
This is likely due to the novel inductive biases of our approach, which lead to distinct predictions.
Ensembling predictions is more effective when the considered methods make fewer correlated errors~\citep{breimann-mlj01a}, which encourages the use of more diverse strategies, making \methodname{} an ideal candidate for ensembling with baseline methods \citep{ensemblediversity}.
Also, \methodname{} is evaluated so quickly, that predictions are almost free, compared to the long runtimes of the baselines. To demonstrate its potential for ensembling, we include an entry \textit{"TabPFN + AutoGluon"} in Table \ref{table:tabular_results_table_small}, generated by averaging the predictions of \methodname{} and \textit{AutoGluon}; this strongly outperforms all other methods.

\slimparagraph{Model Generalization}
The PFN architecture accepts datasets of any length as input. However, when training our model in the \textit{synthetic prior-fitting} phase, we limited synthetic data to a maximum size of $1\,024$, as prior-fitting becomes more expensive with larger datasets.
Nevertheless, we wondered: Would \methodname{} generalize to larger training set sizes that were never seen during training?
% on our prior we limit the sequence length of datasets to 1024 of which a random fraction as used as a training set and the remaining samples as evaluation data.
To test this, we used a collection of $18$ datasets from the \openmlautoml{} Benchmark (see Table \ref{table:openml_datasets_table}) from which we selected $10\,000$ samples each. We evaluated \methodname{} on up to $5\,000$ training samples, using 5 random data splits and $5000$ test samples. %That is a $5$-fold increase compared to the dataset sizes seen during prior-fitting of the PFN.
Surprisingly, our models generalize beyond sample sizes seen during training, as shown in Figure \ref{fig:model generalization} in Appendix~\ref{app:sec:detailed_results}.
%That is, average ROC AUC improves when more training data is provided.

%\section{Uncertainty estimation}

%\vspace*{-0.1cm}
\section{Conclusions \& Future Work}
%\vspace*{-0.1cm}
%Categorical features
%Mixture of more models
%Plausible causal structures
%Spped allows for new applications, here we show aleatoric, but also interpretability etc.
%Sequence length, but goes hand in hand with strong priors
%Custom applications on top of causal prior
%Diffable allows to encode more real world knowledge
We have shown how a single Transformer, the \methodname{}, can be trained to do the work of a full AutoML framework for tabular data and can yield predictions in 0.4 seconds that are competitive with the performance that the best available AutoML frameworks achieve in $5$ to $60$ minutes.
%This result could disrupt the field of data science since it allows real-time high-quality fitting and prediction. 
%It also
This
slashes the computational expense of AutoML, enabling affordable, green, AutoML.

The \methodname{} still has important limitations: the underlying Transformer architecture only scales to small datasets as detailed in Appendix \ref{sec:limitations}; our evaluations focused on classification datasets with only up to 1\,000 training samples, 100 purely numerical features without missing values and 10 classes, which motivates work on (1)~scaling up to large datasets. Our in-depth analysis of the inductive biases of \methodname{} (see Appendix \ref{sec:app:in_depth}), points towards extensions for (2)~improved handling of categorical features, (3) missing values and (4) robustness to unimportant features. Also, our work motivates a multitude of exciting follow-ups regarding (5)~integration of \methodname{} into existing AutoML frameworks; (6)~ensembling to continue making improvements given more time; (7)~dataset-dependent choices of the prior; (8)~generalizations to non-tabular data and (9)~regression tasks, as well as studies of the \methodname{} regarding the dimensions of trustworthy AI (e.g., (10)~out-of-distribution robustness; (11)~algorithmic fairness, (12)~robustness to adversarial examples, and (13)~explainability). The almost instant state-of-the-art predictions of \methodname{} are also likely to give rise to (14)~novel exploratory data analysis methods, (15)~novel feature engineering methods and (16)~novel active learning methods. Finally, our advances in causal reasoning warrant follow-ups on (17)~approximating the effects of interventions and counterfactuals considering a distribution of SCMs.

\newpage

\section{Ethics Statements}
\label{sec:impacts}
In terms of broader societal impact of this work, we do not see any foreseeable strongly negative impacts.
However, this paper could positively impact the carbon footprint and accessibility of learning algorithms. The computations required for machine learning research have been doubling every few months, resulting in a large carbon footprint \citep{schwartz2020green}. Moreover, the financial cost of the computations can make it difficult for academics, students, and researchers to apply these methods.
The decreased computational time shown by \methodname{} translates to reductions in CO2 emissions and cost, making it available to an audience that does not have access to larger scale computing. 

As the \methodname{} provides a highly portable and convenient way of building new classifiers that work in real-time, it is likely to increase the pervasiveness of machine learning even further. While this can have many positive effects on society, such as better personalized healthcare, increased customer satisfaction, efficiency of processes, etc, it will also be crucial to study and improve the \methodname{} under the lens of the many dimensions of trustworthy AI other than computational sustainability, such as algorithmic fairness, robustness to adversarial examples, explainability, and auditability. We hope that its foundation in causal models and simplicity will allow possible avenues for work along these lines.
% ---------------------------------------------

\section{Reproducibility}\label{sec:reproducibility}

\paragraph{Code release} In an effort to ensure reproducibility, we release code alongside our pre-trained \methodname{} and notebooks to reproduce our experiments at 
% Anonymize
%\url{https://github.com/tabpfn-anonym/TabPFNAnonym}
\url{https://github.com/automl/TabPFN}
.

\textbf{Application to public benchmarks} In our work, we evaluate \methodname{} to publicly available benchmarks: the \openmlcc{} Benchmark and the \openmlautoml{} Benchmark. This ensures, that dataset choices are not cherry-picked to our method. Also, for the \openmlautoml{} Benchmark, we use official baseline results and evaluate our method using the evaluation scripts published for this benchmark.\footnote{Available at https://github.com/openml/automlbenchmark}

\textbf{Availability of datasets} All datasets used in our experiments are freely available at \url{OpenML.org}~\citep{vanschoren-sigkdd14a}, with downloading procedures included in the submission. Further details on the datasets used can be found in Section \ref{app:ssec:datasets}.

\paragraph{Online resources}We created a Colab notebook, that lets you interact with our scikit-learn interface at \url{https://colab.research.google.com/drive/1J0l1AtMV_H1KQ7IRbgJje5hMhKHczH7-?usp=sharing}.

We created another Colab, where our evaluation and plots on 179 test and validation datasets can be reproduced easily. \url{https://colab.research.google.com/drive/1yUGaAf3D7RSyO5Jc4PYXSVbtaUAozh6S}

We also created two demos. One to experiment with the TabPFNs predictions (\url{https://huggingface.co/spaces/TabPFN/TabPFNPrediction}) and one to check cross-validation ROC AUC scores on new datasets (\url{https://huggingface.co/spaces/TabPFN/TabPFNEvaluation}). Both of them run on a weak CPU, thus it can require a little bit of time. 

\textbf{Details of training procedures for \methodname{} and baselines} Details shared in the training procedure of all our Transformer models can be found in Appendix \ref{app:expdetails}.
An overview of the hyperparameters used for running the \methodname{} and our baselines can be found in Tables \ref{tab:diffable_hps} and \ref{app:tab:spaces}, respectively.

 % shouldnt be part of submission, because of double-blind review
\subsubsection*{Acknowledgments}
Robert Bosch GmbH is acknowledged for financial support. 
This research was supported by the Deutsche Forschungsgemeinschaft (DFG, German Research Foundation) under grant number 417962828, the state of Baden-W\"{u}rttemberg through bwHPC and the German Research Foundation (DFG) through grant no INST 39/963-1 FUGG, and TAILOR, a project funded by EU Horizon 2020 research, and innovation programme under GA No 952215. 
We acknowledge funding through the European Research Council (ERC) Consolidator Grant ``Deep Learning 2.0'' (grant no.\ 101045765). Funded by the European Union. Views and opinions expressed are however those of the author(s) only and do not necessarily reflect those of the European Union or the ERC. Neither the European Union nor the ERC can be held responsible for them. 
\begin{center}\includegraphics[width=0.3\textwidth]{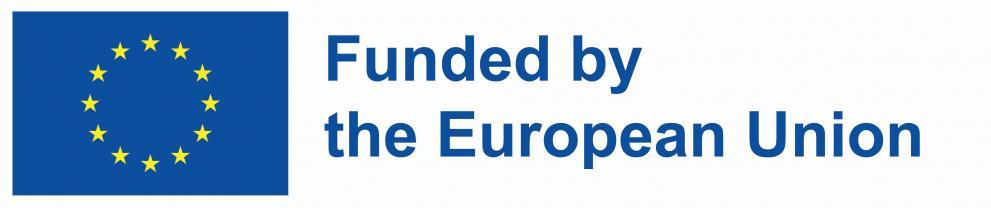}\end{center}
%\newpage

%with competitive performance in a fraction of time. 
%Our model, \methodname{}, is likely to enable new applications of AutoML on small tabular datasets, as it yields a $1000\times$ speedup over other comparably performing AutoML methods.
%We have shown how crucial the right choice of prior is to build such a powerful method based on PFNs.
%Future work can of course entail increasing the viable dataset size even further.
%There are multiple viable directions for future work. i) One can try to improve performance with other priors based on the SCMs or e.g.\ based on decision trees. ii) Also, scaling this method up even further, to even larger tables can be an interesting challenge to consider. iii) Lastly, work on regression for tabular data is a viable path for future work.

\bibliographystyle{plainnat}
\bibliography{strings,lib,main,proc}

\newpage
\appendix

\section{Limitations}
\label{sec:limitations}
The runtime and memory usage of the Transformer-based PFN architecture used in this work scales quadratically with the number of inputs, i.e., training samples passed. Thus, inference on larger sequences (> $100\,000$) is hard on current consumer GPUs. A growing number of methods seek to tackle this issue and report similar performances while scaling linearly with the number of inputs \citep{zaheer2020big, beltagy2020longformer}. These methods can be integrated into the PFN architecture and thus into the \methodname{}.
Furthermore, in our experiments we limit the number of features to 100 and the number of classes to 10 as described in Section \ref{sec:experiments}. While this choice is flexible, the precise \methodname{} that we fitted cannot work with datasets that go beyond these limits.
We also focused the development of \methodname{} to purely numerical datasets without missing values, and while they \emph{can} be applied to datasets with categorical features and/or missing values, their performance is generally worse. We hope to tackle this problem in future versions of \methodname{} by a modified architecture and prior. Finally, we did not consider the existence of many uninformative features in our prior, leading to performance degradation when such features are added; we hope to address this issue in future versions of \methodname{}.
While baseline models (except Gaussian Processes) fit and predict in separate steps, \methodname{} performs both at the same time. Thus, baseline models are often faster than \methodname{} in terms of pure inference time (see Table \ref{app:table:tabular_results_table}).

\section{Additional Results}\label{app:ssec:addresults}
% ---------------------------------------------
\subsection{Detailed Tabular Results}
\label{app:sec:detailed_results}
\begin{figure}[h]
 \centering
 \hspace*{-0.9cm}
 % This file was created with tikzplotlib v0.9.12.
\begin{tikzpicture}

\definecolor{color0}{rgb}{0.194607843137255,0.453431372549019,0.632843137254902}
\definecolor{color1}{rgb}{0.881862745098039,0.505392156862745,0.173039215686275}

\begin{groupplot}[group style={group size=3 by 1}]
\nextgroupplot[
height=5cm,
legend cell align={left},
legend style={
  fill opacity=0.8,
  draw opacity=1,
  text opacity=1,
  at={(0.97,0.03)},
  anchor=south east,
  draw=white!80!black
},
minor xtick={},
minor ytick={},
tick align=outside,
tick pos=left,
title={Has Categorical Features},
width=5cm,
x grid style={white!69.0196078431373!black},
xlabel={Normalized ROC AUC},
xmin=0, xmax=1,
xtick style={color=black},
xtick={0,0.25,0.5,0.75,1},
y dir=reverse,
y grid style={white!69.0196078431373!black},
ymin=-0.5, ymax=6.5,
ytick style={color=black},
ytick={0,1,2,3,4,5,6},
yticklabels={TabPFN,Autosklearn2,Autogluon,Catboost,XGB,LGBM,Reg. Cocktail}
]
\draw[draw=none,fill=color0,line width=1pt] (axis cs:0,-0.4) rectangle (axis cs:0.88985322045472,0);
\addlegendimage{ybar,ybar legend,draw=none,fill=color0,line width=1pt}
\addlegendentry{False}

\draw[draw=none,fill=color0,line width=1pt] (axis cs:0,0.6) rectangle (axis cs:0.76890888576935,1);
\draw[draw=none,fill=color0,line width=1pt] (axis cs:0,1.6) rectangle (axis cs:0.798489991236987,2);
\draw[draw=none,fill=color0,line width=1pt] (axis cs:0,2.6) rectangle (axis cs:0.636875934457407,3);
\draw[draw=none,fill=color0,line width=1pt] (axis cs:0,3.6) rectangle (axis cs:0.443650918529101,4);
\draw[draw=none,fill=color0,line width=1pt] (axis cs:0,4.6) rectangle (axis cs:0.299675428955759,5);
\draw[draw=none,fill=color0,line width=1pt] (axis cs:0,5.6) rectangle (axis cs:0.449646221628231,6);
\draw[draw=none,fill=color1,line width=1pt] (axis cs:0,-2.77555756156289e-17) rectangle (axis cs:0.582815422126868,0.4);
\addlegendimage{ybar,ybar legend,draw=none,fill=color1,line width=1pt}
\addlegendentry{True}

\draw[draw=none,fill=color1,line width=1pt] (axis cs:0,1) rectangle (axis cs:0.82859759137097,1.4);
\draw[draw=none,fill=color1,line width=1pt] (axis cs:0,2) rectangle (axis cs:0.81345721390492,2.4);
\draw[draw=none,fill=color1,line width=1pt] (axis cs:0,3) rectangle (axis cs:0.67499397174847,3.4);
\draw[draw=none,fill=color1,line width=1pt] (axis cs:0,4) rectangle (axis cs:0.7464450252186,4.4);
\draw[draw=none,fill=color1,line width=1pt] (axis cs:0,5) rectangle (axis cs:0.387279199764912,5.4);
\draw[draw=none,fill=color1,line width=1pt] (axis cs:0,6) rectangle (axis cs:0.325021787440162,6.4);

\nextgroupplot[
height=5cm,
legend cell align={left},
legend style={
  fill opacity=0.8,
  draw opacity=1,
  text opacity=1,
  at={(0.97,0.03)},
  anchor=south east,
  draw=white!80!black
},
minor xtick={},
minor ytick={},
scaled y ticks=manual:{}{\pgfmathparse{#1}},
tick align=outside,
tick pos=left,
title={Has Nans in Features},
width=5cm,
x grid style={white!69.0196078431373!black},
xlabel={Normalized ROC AUC},
xmin=0, xmax=1,
xtick style={color=black},
xtick={0,0.25,0.5,0.75,1},
y dir=reverse,
y grid style={white!69.0196078431373!black},
ymin=-0.5, ymax=6.5,
ytick style={color=black},
ytick={0,1,2,3,4,5,6},
yticklabels={}
]
\draw[draw=none,fill=color0,line width=1pt] (axis cs:0,-0.4) rectangle (axis cs:0.842801832138622,0);
\draw[draw=none,fill=color0,line width=1pt] (axis cs:0,0.6) rectangle (axis cs:0.782480423273864,1);
\draw[draw=none,fill=color0,line width=1pt] (axis cs:0,1.6) rectangle (axis cs:0.797219654492428,2);
\draw[draw=none,fill=color0,line width=1pt] (axis cs:0,2.6) rectangle (axis cs:0.640144165931972,3);
\draw[draw=none,fill=color0,line width=1pt] (axis cs:0,3.6) rectangle (axis cs:0.537812758653069,4);
\draw[draw=none,fill=color0,line width=1pt] (axis cs:0,4.6) rectangle (axis cs:0.320371196903669,5);
\draw[draw=none,fill=color0,line width=1pt] (axis cs:0,5.6) rectangle (axis cs:0.371827031122947,6);
\draw[draw=none,fill=color1,line width=1pt] (axis cs:0,-2.77555756156289e-17) rectangle (axis cs:0.566329109839361,0.4);
\draw[draw=none,fill=color1,line width=1pt] (axis cs:0,1) rectangle (axis cs:0.814103911753995,1.4);
\draw[draw=none,fill=color1,line width=1pt] (axis cs:0,2) rectangle (axis cs:0.828516709328443,2.4);
\draw[draw=none,fill=color1,line width=1pt] (axis cs:0,3) rectangle (axis cs:0.687333070710921,3.4);
\draw[draw=none,fill=color1,line width=1pt] (axis cs:0,4) rectangle (axis cs:0.571660402515729,4.4);
\draw[draw=none,fill=color1,line width=1pt] (axis cs:0,5) rectangle (axis cs:0.362898641846041,5.4);
\draw[draw=none,fill=color1,line width=1pt] (axis cs:0,6) rectangle (axis cs:0.553215593335918,6.4);

\nextgroupplot[
height=5cm,
legend cell align={left},
legend style={
  fill opacity=0.8,
  draw opacity=1,
  text opacity=1,
  at={(0.97,0.03)},
  anchor=south east,
  draw=white!80!black
},
minor xtick={},
minor ytick={},
scaled y ticks=manual:{}{\pgfmathparse{#1}},
tick align=outside,
tick pos=left,
title={Multiclass},
width=5cm,
x grid style={white!69.0196078431373!black},
xlabel={Normalized ROC AUC},
xmin=0, xmax=1,
xtick style={color=black},
xtick={0,0.25,0.5,0.75,1},
y dir=reverse,
y grid style={white!69.0196078431373!black},
ymin=-0.5, ymax=6.5,
ytick style={color=black},
ytick={0,1,2,3,4,5,6},
yticklabels={}
]
\draw[draw=none,fill=color0,line width=1pt] (axis cs:0,-0.4) rectangle (axis cs:0.781792011758519,0);
\draw[draw=none,fill=color0,line width=1pt] (axis cs:0,0.6) rectangle (axis cs:0.765607260538324,1);
\draw[draw=none,fill=color0,line width=1pt] (axis cs:0,1.6) rectangle (axis cs:0.80606966333272,2);
\draw[draw=none,fill=color0,line width=1pt] (axis cs:0,2.6) rectangle (axis cs:0.756229121422002,3);
\draw[draw=none,fill=color0,line width=1pt] (axis cs:0,3.6) rectangle (axis cs:0.560185023926927,4);
\draw[draw=none,fill=color0,line width=1pt] (axis cs:0,4.6) rectangle (axis cs:0.331119845477845,5);
\draw[draw=none,fill=color0,line width=1pt] (axis cs:0,5.6) rectangle (axis cs:0.421152131844848,6);
\draw[draw=none,fill=color1,line width=1pt] (axis cs:0,-2.77555756156289e-17) rectangle (axis cs:0.79498111003602,0.4);
\draw[draw=none,fill=color1,line width=1pt] (axis cs:0,1) rectangle (axis cs:0.819140784611169,1.4);
\draw[draw=none,fill=color1,line width=1pt] (axis cs:0,2) rectangle (axis cs:0.800091360548668,2.4);
\draw[draw=none,fill=color1,line width=1pt] (axis cs:0,3) rectangle (axis cs:0.510120257112216,3.4);
\draw[draw=none,fill=color1,line width=1pt] (axis cs:0,4) rectangle (axis cs:0.524178708923867,4.4);
\draw[draw=none,fill=color1,line width=1pt] (axis cs:0,5) rectangle (axis cs:0.325943323356995,5.4);
\draw[draw=none,fill=color1,line width=1pt] (axis cs:0,6) rectangle (axis cs:0.391042774277216,6.4);
\end{groupplot}

\end{tikzpicture}
      \caption{Normalized ROC AUC performance on datasets from the \openmlcc{} Benchmark, divided by dataset characteristics. For each plot, we split the datasets into two groups. Left: Orange bars indicate the performance on datasets that have categorical features. Middle: Orange bars indicate datasets that contain missing values. Right: Orange bars indicate multiclass datasets, while others are binary.}
 \label{fig:results_dataset_characteristics}
\end{figure}
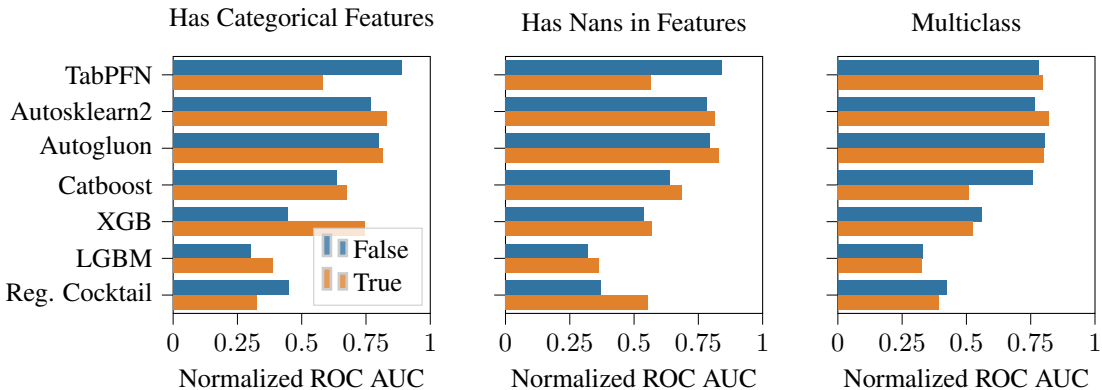
     
%In addition to the results in the main paper in Section~\ref{ssec:tabresults}, we report a wide range of performance values in Table \ref{app:table:tabular_results_table}. We show quantitative results for comparing all methods when given up to $60min$ time. We see that our model is competitive with strong AutoML baselines while requiring a fraction of that time. %Furthermore, we report ROC AUC performance over time, similarly to Figure~\ref{fig:perfovertime}, except that AutoGluon optimizes for ROC AUC. While the mean rank and number of wins remains similar, the average AUC ROC performance decreases in our experiments, due to failure to train Catboost models on $2$ datasets (MiceProtein, mfeat-zernike).
In Figure \ref{fig:results_dataset_characteristics}, we explore how the kind of dataset evaluated affects the performance of \methodname{}, compared to our baselines. We find that \methodname{} performs much better when no categorical features are present. 
We also find that \methodname{} performs better, when no missing values are present in the data.
This warrants an extension of our prior in future work, to make it more customized towards categorical and missing data.
%This can be attributed to \methodname{} assuming that missing values are simply missing by chance in the prior, which could be alleviated by incorporating a reason for values to be missing values in the prior (e.g. when another node surpasses some value limit).
Our method seems to work comparably well for binary and multi-class problems.

\begin{figure}[t]
     \centering
     \begin{subfigure}[t]{0.32\textwidth}
         \centering
         %\hspace*{-1.9cm}
         \input{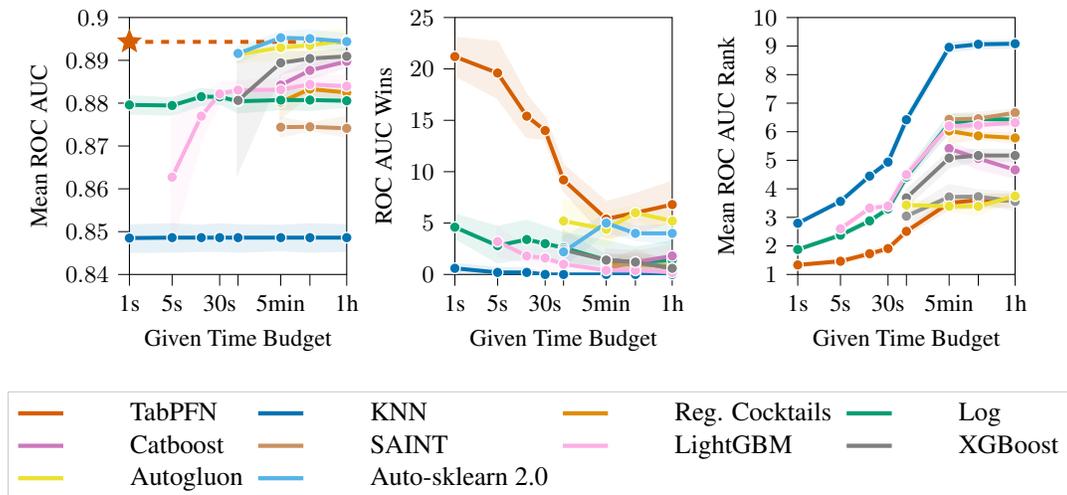}
         %\label{fig:aucovertime}
     \end{subfigure}
     \begin{subfigure}[t]{0.32\textwidth}
         \centering
         %\hspace*{-0.2cm}
         % This file was created with tikzplotlib v0.9.12.
\begin{tikzpicture}

\begin{axis}[
height=5cm,
legend style={fill opacity=0.8, draw opacity=1, text opacity=1, draw=none},
minor xtick={},
minor ytick={},
tick align=outside,
tick pos=left,
width=\textwidth,
x grid style={white!69.0196078431373!black},
xlabel={Given Time Budget},
xmin=0, xmax=3.55630250076729,
xtick style={color=black},
xtick={0,0.698970004336019,1.47712125471966,1.77815125038364,2.47712125471966,2.95424250943932,3.55630250076729},
xticklabels={1s,5s,30s,,5min,,1h},
y grid style={white!69.0196078431373!black},
ylabel={ROC AUC Wins},
ymin=0, ymax=25,
ytick style={color=black},
ytick={0,5,10,15,20,25},
mystylepalette
]
\path [draw=color0, fill=color0, opacity=0.1, line width=1pt]
(axis cs:0,1)
--(axis cs:0,0.2)
--(axis cs:0.698970004336019,0)
--(axis cs:1.17609125905568,0)
--(axis cs:1.47712125471966,0)
--(axis cs:1.77815125038364,0)
--(axis cs:2.47712125471966,0)
--(axis cs:2.95424250943932,0)
--(axis cs:3.55630250076729,0)
--(axis cs:4.15836249209525,0)
--(axis cs:4.15836249209525,0)
--(axis cs:4.15836249209525,0)
--(axis cs:3.55630250076729,0)
--(axis cs:2.95424250943932,0)
--(axis cs:2.47712125471966,0)
--(axis cs:1.77815125038364,0)
--(axis cs:1.47712125471966,0)
--(axis cs:1.17609125905568,0.6)
--(axis cs:0.698970004336019,0.6)
--(axis cs:0,1)
--cycle;

\path [draw=color1, fill=color1, opacity=0.1, line width=1pt]
(axis cs:2.47712125471966,2)
--(axis cs:2.47712125471966,0.2)
--(axis cs:2.95424250943932,0.2)
--(axis cs:3.55630250076729,0.2)
--(axis cs:3.55630250076729,1.4)
--(axis cs:3.55630250076729,1.4)
--(axis cs:2.95424250943932,1.605)
--(axis cs:2.47712125471966,2)
--cycle;

\path [draw=color2, fill=color2, opacity=0.1, line width=1pt]
(axis cs:0,5.8)
--(axis cs:0,3.6)
--(axis cs:0.698970004336019,1.2)
--(axis cs:1.17609125905568,1.6)
--(axis cs:1.47712125471966,1.4)
--(axis cs:1.77815125038364,1.595)
--(axis cs:2.47712125471966,0.2)
--(axis cs:2.95424250943932,0.2)
--(axis cs:3.55630250076729,0.2)
--(axis cs:4.15836249209525,0.2)
--(axis cs:4.15836249209525,3.2)
--(axis cs:4.15836249209525,3.2)
--(axis cs:3.55630250076729,3.2)
--(axis cs:2.95424250943932,2.6)
--(axis cs:2.47712125471966,3.2)
--(axis cs:1.77815125038364,4)
--(axis cs:1.47712125471966,4.6)
--(axis cs:1.17609125905568,5.2)
--(axis cs:0.698970004336019,4.6)
--(axis cs:0,5.8)
--cycle;

\path [draw=color3, fill=color3, opacity=0.1, line width=1pt]
(axis cs:0,23)
--(axis cs:0,19.4)
--(axis cs:0.698970004336019,17.2)
--(axis cs:1.17609125905568,13.4)
--(axis cs:1.47712125471966,13.2)
--(axis cs:1.77815125038364,8.2)
--(axis cs:2.47712125471966,3.6)
--(axis cs:2.95424250943932,4)
--(axis cs:3.55630250076729,5.2)
--(axis cs:3.55630250076729,9)
--(axis cs:3.55630250076729,9)
--(axis cs:2.95424250943932,7.8)
--(axis cs:2.47712125471966,7.005)
--(axis cs:1.77815125038364,10.6)
--(axis cs:1.47712125471966,15.2)
--(axis cs:1.17609125905568,17.4)
--(axis cs:0.698970004336019,22.6)
--(axis cs:0,23)
--cycle;

\path [draw=color4, fill=color4, opacity=0.1, line width=1pt]
(axis cs:2.47712125471966,2.4)
--(axis cs:2.47712125471966,0.6)
--(axis cs:2.95424250943932,0.2)
--(axis cs:3.55630250076729,0.8)
--(axis cs:3.55630250076729,2.8)
--(axis cs:3.55630250076729,2.8)
--(axis cs:2.95424250943932,2.2)
--(axis cs:2.47712125471966,2.4)
--cycle;

\path [draw=color5, fill=color5, opacity=0.1, line width=1pt]
(axis cs:2.47712125471966,1)
--(axis cs:2.47712125471966,0.2)
--(axis cs:2.95424250943932,0.2)
--(axis cs:3.55630250076729,0.2)
--(axis cs:3.55630250076729,1)
--(axis cs:3.55630250076729,1)
--(axis cs:2.95424250943932,2.4)
--(axis cs:2.47712125471966,1)
--cycle;

\path [draw=color6, fill=color6, opacity=0.1, line width=1pt]
(axis cs:0.698970004336019,4.6)
--(axis cs:0.698970004336019,1.8)
--(axis cs:1.17609125905568,1.2)
--(axis cs:1.47712125471966,0.6)
--(axis cs:1.77815125038364,0.2)
--(axis cs:2.47712125471966,0)
--(axis cs:2.95424250943932,0)
--(axis cs:3.55630250076729,0)
--(axis cs:3.55630250076729,0.6)
--(axis cs:3.55630250076729,0.6)
--(axis cs:2.95424250943932,0.8)
--(axis cs:2.47712125471966,0.8)
--(axis cs:1.77815125038364,2)
--(axis cs:1.47712125471966,3)
--(axis cs:1.17609125905568,2.4)
--(axis cs:0.698970004336019,4.6)
--cycle;

\path [draw=white!49.8039215686275!black, fill=white!49.8039215686275!black, opacity=0.1, line width=1pt]
(axis cs:1.77815125038364,4.2)
--(axis cs:1.77815125038364,1.2)
--(axis cs:2.47712125471966,1)
--(axis cs:2.95424250943932,0.4)
--(axis cs:3.55630250076729,0)
--(axis cs:3.55630250076729,1.4)
--(axis cs:3.55630250076729,1.4)
--(axis cs:2.95424250943932,2)
--(axis cs:2.47712125471966,1.8)
--(axis cs:1.77815125038364,4.2)
--cycle;

\path [draw=color7, fill=color7, opacity=0.1, line width=1pt]
(axis cs:1.77815125038364,7.2)
--(axis cs:1.77815125038364,3.8)
--(axis cs:2.47712125471966,2.8)
--(axis cs:2.95424250943932,5.4)
--(axis cs:3.55630250076729,4)
--(axis cs:3.55630250076729,6)
--(axis cs:3.55630250076729,6)
--(axis cs:2.95424250943932,6.6)
--(axis cs:2.47712125471966,6)
--(axis cs:1.77815125038364,7.2)
--cycle;

\path [draw=color8, fill=color8, opacity=0.1, line width=1pt]
(axis cs:1.77815125038364,3.605)
--(axis cs:1.77815125038364,0.8)
--(axis cs:2.47712125471966,4.2)
--(axis cs:2.95424250943932,3.4)
--(axis cs:3.55630250076729,3)
--(axis cs:3.55630250076729,4.8)
--(axis cs:3.55630250076729,4.8)
--(axis cs:2.95424250943932,4.6)
--(axis cs:2.47712125471966,5.8)
--(axis cs:1.77815125038364,3.605)
--cycle;

\addplot [line width=1.5pt, color0, mystyle, forget plot]
table {%
0 0.6
0.698970004336019 0.2
1.17609125905568 0.2
1.47712125471966 0
1.77815125038364 0
2.47712125471966 0
2.95424250943932 0
3.55630250076729 0
4.15836249209525 0
};
\addplot [line width=1.5pt, color1, mystyle, forget plot]
table {%
2.47712125471966 1
2.95424250943932 1
3.55630250076729 0.8
};
\addplot [line width=1.5pt, color2, mystyle, forget plot]
table {%
0 4.6
0.698970004336019 2.8
1.17609125905568 3.4
1.47712125471966 3
1.77815125038364 2.6
2.47712125471966 1.4
2.95424250943932 1.2
3.55630250076729 1.4
4.15836249209525 1.4
};
\addplot [line width=1.5pt, color3, mystyle, forget plot]
table {%
0 21.2
0.698970004336019 19.6
1.17609125905568 15.4
1.47712125471966 14
1.77815125038364 9.2
2.47712125471966 5.4
2.95424250943932 6
3.55630250076729 6.8
};
\addplot [line width=1.5pt, color4, mystyle, forget plot]
table {%
2.47712125471966 1.4
2.95424250943932 1.2
3.55630250076729 1.8
};
\addplot [line width=1.5pt, color5, mystyle, forget plot]
table {%
2.47712125471966 0.6
2.95424250943932 1.2
3.55630250076729 0.6
};
\addplot [line width=1.5pt, color6, mystyle, forget plot]
table {%
0.698970004336019 3.2
1.17609125905568 1.8
1.47712125471966 1.6
1.77815125038364 1
2.47712125471966 0.4
2.95424250943932 0.4
3.55630250076729 0.2
};
\addplot [line width=1.5pt, white!49.8039215686275!black, mystyle, forget plot]
table {%
1.77815125038364 2.4
2.47712125471966 1.4
2.95424250943932 1.2
3.55630250076729 0.6
};
\addplot [line width=1.5pt, color7, mystyle, forget plot]
table {%
1.77815125038364 5.2
2.47712125471966 4.4
2.95424250943932 6
3.55630250076729 5.2
};
\addplot [line width=1.5pt, color8, mystyle, forget plot]
table {%
1.77815125038364 2.2
2.47712125471966 5
2.95424250943932 4
3.55630250076729 4
};
\end{axis}

\end{tikzpicture}
         %\label{fig:winsovertime}
     \end{subfigure}
     \begin{subfigure}[t]{0.32\textwidth}
         \centering
         %\hspace*{1.1cm}
         % This file was created with tikzplotlib v0.9.12.
\begin{tikzpicture}

\definecolor{color0}{HTML}{0173b2}
\definecolor{color1}{HTML}{de8f05}
\definecolor{color2}{HTML}{029e73}
\definecolor{color3}{HTML}{d55e00}
\definecolor{color4}{HTML}{cc78bc}
\definecolor{color5}{HTML}{ca9161}
\definecolor{color6}{HTML}{fbafe4}
\definecolor{color7}{HTML}{949494}
\definecolor{color8}{HTML}{ece133}
\definecolor{color9}{HTML}{56b4e9}

\begin{axis}[
height=5cm,
legend style={fill opacity=0.8, draw opacity=1, text opacity=1, draw=none},
minor xtick={},
minor ytick={},
tick align=outside,
tick pos=left,
width=\textwidth,
x grid style={white!69.0196078431373!black},
xlabel={Given Time Budget},
xmin=0, xmax=3.55630250076729,
xtick style={color=black},
xtick={0,0.698970004336019,1.47712125471966,1.77815125038364,2.47712125471966,2.95424250943932,3.55630250076729},
xticklabels={1s,5s,30s,,5min,,1h},
y grid style={white!69.0196078431373!black},
ylabel={Mean ROC AUC Rank},
ymin=1, ymax=10,
ytick style={color=black},
ytick={1,2,3,4,5,6,7,8,9,10},
mystylepalette
]
\path [draw=color0, fill=color0, opacity=0.1, line width=1pt]
(axis cs:0,2.80666666666667)
--(axis cs:0,2.78)
--(axis cs:0.698970004336019,3.51666666666667)
--(axis cs:1.17609125905568,4.40333333333333)
--(axis cs:1.47712125471966,4.83666666666667)
--(axis cs:1.77815125038364,6.27)
--(axis cs:2.47712125471966,8.81333333333333)
--(axis cs:2.95424250943932,8.94333333333333)
--(axis cs:3.55630250076729,8.95)
--(axis cs:4.15836249209525,8.95333333333333)
--(axis cs:4.15836249209525,9.17)
--(axis cs:4.15836249209525,9.17)
--(axis cs:3.55630250076729,9.17333333333333)
--(axis cs:2.95424250943932,9.17)
--(axis cs:2.47712125471966,9.10333333333333)
--(axis cs:1.77815125038364,6.54)
--(axis cs:1.47712125471966,5)
--(axis cs:1.17609125905568,4.49666666666667)
--(axis cs:0.698970004336019,3.59666666666667)
--(axis cs:0,2.80666666666667)
--cycle;

\path [draw=color1, fill=color1, opacity=0.1, line width=1pt]
(axis cs:2.47712125471966,6.48675)
--(axis cs:2.47712125471966,5.59666666666667)
--(axis cs:2.95424250943932,5.76666666666667)
--(axis cs:3.55630250076729,5.62333333333333)
--(axis cs:3.55630250076729,5.93333333333333)
--(axis cs:3.55630250076729,5.93333333333333)
--(axis cs:2.95424250943932,5.93333333333333)
--(axis cs:2.47712125471966,6.48675)
--cycle;

\path [draw=color2, fill=color2, opacity=0.1, line width=1pt]
(axis cs:0,1.91666666666667)
--(axis cs:0,1.83333333333333)
--(axis cs:0.698970004336019,2.32)
--(axis cs:1.17609125905568,2.79666666666667)
--(axis cs:1.47712125471966,3.19325)
--(axis cs:1.77815125038364,4.26666666666667)
--(axis cs:2.47712125471966,5.98966666666667)
--(axis cs:2.95424250943932,6.25325)
--(axis cs:3.55630250076729,6.22666666666667)
--(axis cs:4.15836249209525,6.23666666666667)
--(axis cs:4.15836249209525,6.53333333333333)
--(axis cs:4.15836249209525,6.53333333333333)
--(axis cs:3.55630250076729,6.53333333333333)
--(axis cs:2.95424250943932,6.6)
--(axis cs:2.47712125471966,6.59)
--(axis cs:1.77815125038364,4.56675)
--(axis cs:1.47712125471966,3.37)
--(axis cs:1.17609125905568,2.93666666666667)
--(axis cs:0.698970004336019,2.42666666666667)
--(axis cs:0,1.91666666666667)
--cycle;

\path [draw=color3, fill=color3, opacity=0.1, line width=1pt]
(axis cs:0,1.37666666666667)
--(axis cs:0,1.3)
--(axis cs:0.698970004336019,1.38)
--(axis cs:1.17609125905568,1.63333333333333)
--(axis cs:1.47712125471966,1.83333333333333)
--(axis cs:1.77815125038364,2.38658333333333)
--(axis cs:2.47712125471966,3.29666666666667)
--(axis cs:2.95424250943932,3.36)
--(axis cs:3.55630250076729,3.39666666666667)
--(axis cs:3.55630250076729,3.79)
--(axis cs:3.55630250076729,3.79)
--(axis cs:2.95424250943932,3.79666666666667)
--(axis cs:2.47712125471966,3.67666666666667)
--(axis cs:1.77815125038364,2.61666666666667)
--(axis cs:1.47712125471966,1.97666666666667)
--(axis cs:1.17609125905568,1.81666666666667)
--(axis cs:0.698970004336019,1.51666666666667)
--(axis cs:0,1.37666666666667)
--cycle;

\path [draw=color4, fill=color4, opacity=0.1, line width=1pt]
(axis cs:2.47712125471966,5.58)
--(axis cs:2.47712125471966,5.11333333333333)
--(axis cs:2.95424250943932,4.70333333333333)
--(axis cs:3.55630250076729,4.40666666666667)
--(axis cs:3.55630250076729,4.98025)
--(axis cs:3.55630250076729,4.98025)
--(axis cs:2.95424250943932,5.28)
--(axis cs:2.47712125471966,5.58)
--cycle;

\path [draw=color5, fill=color5, opacity=0.1, line width=1pt]
(axis cs:2.47712125471966,6.59333333333333)
--(axis cs:2.47712125471966,6.27333333333333)
--(axis cs:2.95424250943932,6.26)
--(axis cs:3.55630250076729,6.58)
--(axis cs:3.55630250076729,6.76)
--(axis cs:3.55630250076729,6.76)
--(axis cs:2.95424250943932,6.66666666666667)
--(axis cs:2.47712125471966,6.59333333333333)
--cycle;

\path [draw=color6, fill=color6, opacity=0.1, line width=1pt]
(axis cs:0.698970004336019,2.66)
--(axis cs:0.698970004336019,2.54)
--(axis cs:1.17609125905568,3.24)
--(axis cs:1.47712125471966,3.21)
--(axis cs:1.77815125038364,4.34666666666667)
--(axis cs:2.47712125471966,6.00333333333333)
--(axis cs:2.95424250943932,5.94983333333333)
--(axis cs:3.55630250076729,6.01666666666667)
--(axis cs:3.55630250076729,6.57666666666667)
--(axis cs:3.55630250076729,6.57666666666667)
--(axis cs:2.95424250943932,6.50333333333333)
--(axis cs:2.47712125471966,6.40333333333333)
--(axis cs:1.77815125038364,4.64666666666667)
--(axis cs:1.47712125471966,3.55333333333333)
--(axis cs:1.17609125905568,3.40666666666667)
--(axis cs:0.698970004336019,2.66)
--cycle;

\path [draw=white!49.8039215686275!black, fill=white!49.8039215686275!black, opacity=0.1, line width=1pt]
(axis cs:1.77815125038364,3.94)
--(axis cs:1.77815125038364,3.39)
--(axis cs:2.47712125471966,4.77)
--(axis cs:2.95424250943932,5.00325)
--(axis cs:3.55630250076729,4.99666666666667)
--(axis cs:3.55630250076729,5.33666666666667)
--(axis cs:3.55630250076729,5.33666666666667)
--(axis cs:2.95424250943932,5.35)
--(axis cs:2.47712125471966,5.29)
--(axis cs:1.77815125038364,3.94)
--cycle;

\path [draw=color7, fill=color7, opacity=0.1, line width=1pt]
(axis cs:1.77815125038364,3.22666666666667)
--(axis cs:1.77815125038364,2.84333333333333)
--(axis cs:2.47712125471966,3.36)
--(axis cs:2.95424250943932,3.5)
--(axis cs:3.55630250076729,3.27)
--(axis cs:3.55630250076729,3.86333333333333)
--(axis cs:3.55630250076729,3.86333333333333)
--(axis cs:2.95424250943932,3.94008333333333)
--(axis cs:2.47712125471966,4.12333333333333)
--(axis cs:1.77815125038364,3.22666666666667)
--cycle;

\path [draw=color8, fill=color8, opacity=0.1, line width=1pt]
(axis cs:1.77815125038364,3.58)
--(axis cs:1.77815125038364,3.28666666666667)
--(axis cs:2.47712125471966,3.22)
--(axis cs:2.95424250943932,3.08325)
--(axis cs:3.55630250076729,3.66)
--(axis cs:3.55630250076729,3.82)
--(axis cs:3.55630250076729,3.82)
--(axis cs:2.95424250943932,3.71333333333333)
--(axis cs:2.47712125471966,3.57)
--(axis cs:1.77815125038364,3.58)
--cycle;

\addplot [line width=1.5pt, color0, mystyle, forget plot]
table {%
0 2.79333333333333
0.698970004336019 3.56
1.17609125905568 4.44666666666667
1.47712125471966 4.93666666666667
1.77815125038364 6.42
2.47712125471966 8.95666666666667
2.95424250943932 9.06333333333333
3.55630250076729 9.08333333333333
4.15836249209525 9.08333333333333
};
\addplot [line width=1.5pt, color1, mystyle, forget plot]
table {%
2.47712125471966 6.02666666666667
2.95424250943932 5.85666666666667
3.55630250076729 5.78333333333333
};
\addplot [line width=1.5pt, color2, mystyle, forget plot]
table {%
0 1.87333333333333
0.698970004336019 2.37666666666667
1.17609125905568 2.87666666666667
1.47712125471966 3.29666666666667
1.77815125038364 4.41333333333333
2.47712125471966 6.28333333333333
2.95424250943932 6.44666666666667
3.55630250076729 6.42333333333333
4.15836249209525 6.42333333333333
};
\addplot [line width=1.5pt, color3, mystyle, forget plot]
table {%
0 1.33333333333333
0.698970004336019 1.46333333333333
1.17609125905568 1.72666666666667
1.47712125471966 1.90666666666667
1.77815125038364 2.51
2.47712125471966 3.50333333333333
2.95424250943932 3.60666666666667
3.55630250076729 3.59333333333333
};
\addplot [line width=1.5pt, color4, mystyle, forget plot]
table {%
2.47712125471966 5.41
2.95424250943932 5.06333333333333
3.55630250076729 4.66333333333333
};
\addplot [line width=1.5pt, color5, mystyle, forget plot]
table {%
2.47712125471966 6.43333333333333
2.95424250943932 6.45666666666667
3.55630250076729 6.66666666666667
};
\addplot [line width=1.5pt, color6, mystyle, forget plot]
table {%
0.698970004336019 2.6
1.17609125905568 3.32333333333333
1.47712125471966 3.39666666666667
1.77815125038364 4.49666666666667
2.47712125471966 6.20333333333333
2.95424250943932 6.22666666666667
3.55630250076729 6.31666666666667
};
\addplot [line width=1.5pt, white!49.8039215686275!black, mystyle, forget plot]
table {%
1.77815125038364 3.68333333333333
2.47712125471966 5.07666666666667
2.95424250943932 5.16666666666667
3.55630250076729 5.16666666666667
};
\addplot [line width=1.5pt, color7, mystyle, forget plot]
table {%
1.77815125038364 3.04333333333333
2.47712125471966 3.71666666666667
2.95424250943932 3.72666666666667
3.55630250076729 3.55666666666667
};
\addplot [line width=1.5pt, color8, mystyle, forget plot]
table {%
1.77815125038364 3.43333333333333
2.47712125471966 3.39
2.95424250943932 3.38666666666667
3.55630250076729 3.74666666666667
};
\end{axis}

\end{tikzpicture}
         %\label{fig:ranksovertime}
     \end{subfigure}
     \vspace{-.6cm}
     \begin{subfigure}[b]{1.0\textwidth}
         \centering
         % This file was created with tikzplotlib v0.9.12.
\begin{tikzpicture}
\begin{axis}[
hide axis,
width=1.\textwidth,
height=.2\textwidth,
legend columns=4,legend style={draw=none,column sep=5ex},
legend cell align={left},
legend style={fill opacity=0.8, draw opacity=1, text opacity=1, draw=white!80!black, /tikz/every even column/.append style={column sep=0.1cm}},
log basis x={10},
tick align=outside,
tick pos=left,
x grid style={white!69.0196078431373!black},
xlabel={Training time per task},
xmin=0.005, xmax=300,
xmode=log,
xtick style={color=black},
y grid style={white!69.0196078431373!black},
ylabel={Negative Log Likelihood},
ymin=-1, ymax=15,
ytick style={color=black}
]

\addlegendimage{ultra thick, color3}
\addlegendentry{\methodname{}}
\addlegendimage{ultra thick, color0}
\addlegendentry{KNN}
\addlegendimage{ultra thick, color1}
\addlegendentry{Reg. Cocktails}
\addlegendimage{ultra thick, color2}
\addlegendentry{Log}
\addlegendimage{ultra thick, color4}
\addlegendentry{Catboost}
\addlegendimage{ultra thick, color5}
\addlegendentry{SAINT}
\addlegendimage{ultra thick, color6}
\addlegendentry{LightGBM}
\addlegendimage{ultra thick, white!49.8039215686275!black}
\addlegendentry{XGBoost}
\addlegendimage{ultra thick, color7}
\addlegendentry{Autogluon}
\addlegendimage{ultra thick, color8}
\addlegendentry{Auto-sklearn 2.0}
\end{axis}

\end{tikzpicture}
         %\label{fig:legendovertime}
     \end{subfigure}
     \vspace{0cm}
     \caption{ROC AUC performance over time on the $30$ small \openmlcc{} including datasets with categorical features and missing values. We report the mean, mean wins and rank along with the 95\% confidence interval across $5$ splits for increasing training and tuning time budgets (Unlabelled ticks: 1min, 15min). The red star indicates performance of \methodname{} with 32 data permutations (which requires 0.62s on GPU). We report detailed results with a 60 min budget in Table~\ref{table:tabular_results_table_small}
     %The 95\% confidence intervals over splits are shaded. Our method finishes after 1.0 seconds and does not use any further time budget; KNN and Logistic Regression finish without the full time budget as well, dashed lines indicate no further training. 
     %The taken for \methodname{} is on GPU.
     }
     \vspace*{-0.5cm}
     \label{fig:perfovertime_all}
\end{figure}

In addition to the results in the main paper in Section~\ref{ssec:tabresults}, we report detailed results on all $30$ datasets in \openmlcc{}, which are small enough, but might include categorical features and missing values. We show performance over time in Figure \ref{fig:perfovertime_all} and a wide range of performance values and per dataset results with a 1 hour time limit in Table \ref{app:table:tabular_results_table}.% and also provide  a critical difference diagram using the Nemenyi test to compare all methods against each other~\citep{demsar-06a} in Figure~\ref{app:fig:cd}.

% \begin{table}[h]
%     \caption{Average performance on the test-datasets for $15$ minutes ($900$ s) requested time. We report overall \textit{Wins}, \textit{Wins/Ties/Losses (W/T/L)} against \methodname{}, \textit{mean performance (Mean)} and \textit{mean rank (Mean rank)} wrt. ROC AUC, crossentropy loss (CE), and accuracy (Acc). Additionally, we also report the mean time used to fit a single model and obtain predictions.
%     %If available, all Baselines are given ROC AUC optimization as an objective, others optimize CE. Overall each method got a time budget of 37.5 hours, that \methodname{}, KNN, GP and LogReg did not use fully.
%     }
%     \label{tab:tabular_results_table_small}
%     \centering
%     \tiny
%     \input{figures/tabular_results/tabular_results_table_small}
% \end{table}

\begin{table}[h]
    \caption{ROC AUC OVO results on the $30$ small \openmlcc{} (including datasets with categorical features and missing values) for 60 minutes requested time per dataset and per split. If available, all baselines are given ROC AUC optimization as an objective, others optimize CE. Overall each method got a time budget of 150 hours, but not all methods used the full budget. Times for \methodname{} refer to times on GPU. \methodname{} runs training and prediction in a joint step and does not have any hyperparameter tuning, so only a single aggregate time is shown.}
    \label{app:table:tabular_results_table}
    \centering
    \tiny
    \begin{tabular}{@{\hskip 0mm}
l@{\hskip 1mm}
l@{\hskip 1mm}
l@{\hskip 1mm}
l@{\hskip 1mm}
l@{\hskip 1mm}
l@{\hskip 1mm}
l@{\hskip 1mm}
l@{\hskip 1mm}
l@{\hskip 0mm}}
\toprule
{} &        LightGBM &        CatBoost &          XGBoost &          ASKL2.0 &               AutoGluon &  TabPFN$_{n.e.}$ &           TabPFN &        TabPFN + AutoGluon \\
\midrule
balance-scale        &          0.9938 &          0.9245 &           0.9939 &            0.997 &                  0.9919 &           0.9965 &  \textbf{0.9973} &                    0.9958 \\
mfeat-fourier        &          0.9786 &          0.9816 &           0.9803 &           0.9826 &         \textbf{0.9843} &           0.9767 &           0.9811 &                    0.9838 \\
breast-w             &           0.991 &          0.9931 &           0.9896 &           0.9939 &                  0.9933 &           0.9931 &           0.9934 &            \textbf{0.994} \\
mfeat-karhunen       &          0.9979 &          0.9986 &           0.9983 &           0.9975 &         \textbf{0.9987} &           0.9939 &           0.9978 &                    0.9985 \\
mfeat-morphologica.. &          0.9601 &          0.9629 &           0.9612 &           0.9671 &                  0.9698 &           0.9657 &           0.9669 &           \textbf{0.9722} \\
mfeat-zernike        &          0.9716 &          0.9759 &           0.9735 &           0.9812 &         \textbf{0.9908} &           0.9812 &           0.9823 &                    0.9901 \\
cmc                  &          0.7288 &          0.7256 &           0.7299 &  \textbf{0.7378} &                  0.7331 &           0.7233 &           0.7276 &                    0.7336 \\
credit-approval      &          0.9415 &          0.9389 &  \textbf{0.9422} &           0.9406 &                  0.9415 &           0.9253 &           0.9322 &                    0.9394 \\
credit-g             &          0.7684 &          0.7852 &           0.7853 &            0.793 &                  0.7941 &           0.7894 &           0.7894 &           \textbf{0.7948} \\
diabetes             &          0.8247 &          0.8383 &           0.8378 &           0.8343 &                  0.8391 &           0.8412 &            0.841 &           \textbf{0.8427} \\
tic-tac-toe          &          0.9988 &          0.9992 &       \textbf{1} &           0.9943 &              \textbf{1} &           0.9547 &           0.9759 &                    0.9992 \\
vehicle              &          0.9232 &          0.9302 &           0.9282 &           0.9504 &                  0.9416 &           0.9568 &  \textbf{0.9589} &                    0.9538 \\
eucalyptus           &          0.8931 &          0.8979 &           0.9004 &           0.9132 &                  0.9204 &           0.9218 &           0.9245 &           \textbf{0.9278} \\
analcatdata\_author.. &          0.9999 &          0.9999 &           0.9997 &       \textbf{1} &                  0.9993 &       \textbf{1} &       \textbf{1} &                \textbf{1} \\
analcatdata\_dmft     &          0.5461 &          0.5589 &           0.5743 &           0.5752 &                  0.5657 &           0.5643 &   \textbf{0.579} &                    0.5756 \\
pc4                  &          0.9301 &          0.9413 &           0.9291 &           0.9331 &                  0.9428 &           0.9298 &           0.9383 &            \textbf{0.944} \\
pc3                  &          0.8178 &          0.8247 &           0.8288 &           0.8265 &                  0.8282 &           0.8308 &  \textbf{0.8373} &                     0.836 \\
kc2                  &          0.8141 &          0.8323 &           0.8227 &           0.8311 &                  0.8242 &           0.8322 &  \textbf{0.8346} &                    0.8321 \\
pc1                  &          0.8321 &            0.86 &           0.8489 &           0.8527 &                  0.8578 &   \textbf{0.877} &           0.8761 &                    0.8739 \\
banknote-authentic.. &      \textbf{1} &      \textbf{1} &       \textbf{1} &       \textbf{1} &              \textbf{1} &       \textbf{1} &       \textbf{1} &                \textbf{1} \\
blood-transfusion-.. &          0.7144 &          0.7403 &           0.7312 &           0.7504 &                  0.7364 &            0.753 &  \textbf{0.7549} &                    0.7469 \\
ilpd                 &          0.6917 &          0.7279 &           0.7171 &           0.7212 &                   0.723 &  \textbf{0.7412} &           0.7379 &                    0.7326 \\
qsar-biodeg          &          0.9126 &          0.9217 &           0.9191 &           0.9247 &                  0.9276 &  \textbf{0.9345} &           0.9336 &                    0.9336 \\
wdbc                 &          0.9904 &          0.9931 &           0.9904 &           0.9947 &                  0.9956 &            0.996 &  \textbf{0.9964} &                     0.996 \\
cylinder-bands       &          0.8556 &          0.8757 &           0.8782 &           0.8718 &         \textbf{0.8878} &           0.8314 &           0.8336 &                    0.8751 \\
dresses-sales        &          0.5593 &          0.5696 &  \textbf{0.5823} &           0.5705 &                  0.5507 &           0.5333 &           0.5376 &                    0.5509 \\
MiceProtein          &          0.9997 &          0.9999 &           0.9998 &           0.9999 &              \textbf{1} &           0.9997 &           0.9999 &                \textbf{1} \\
car                  &          0.9925 &          0.9955 &           0.9948 &   \textbf{0.998} &                   0.997 &           0.9926 &            0.995 &                    0.9972 \\
steel-plates-fault.. &          0.9626 &          0.9655 &           0.9656 &  \textbf{0.9694} &                  0.9666 &           0.9619 &           0.9655 &                    0.9687 \\
climate-model-simu.. &          0.9286 &          0.9344 &           0.9255 &           0.9291 &                  0.9391 &  \textbf{0.9426} &           0.9415 &                    0.9421 \\
\midrule
Wins AUC OVO
                                         &               0 &               0 &                2 &                2 &                       2 &                4 &       \textbf{5} &                \textbf{5} \\
Wins Acc.            &               0 &               2 &                2 &                3 &                       3 &                0 &                6 &                \textbf{8} \\
Wins CE              &               0 &               1 &                3 &                1 &                       7 &                1 &                6 &                \textbf{9} \\
\midrule
M. rank AUC OVO
                                      &          6.6167 &          4.9667 &           5.4167 &             4.05 &                  3.7833 &             4.65 &              3.7 &           \textbf{2.8167} \\
Mean rank Acc.       &          6.5333 &          4.9833 &           5.1833 &           4.8667 &                  3.8167 &           4.5333 &           3.6167 &           \textbf{2.4667} \\
Mean rank CE         &          5.7333 &             5.6 &           5.4667 &              5.8 &                  2.8667 &           4.6167 &           3.5333 &           \textbf{2.3833} \\
\midrule
Win/T/L AUC
                             vs Tab.. &          5/4/21 &          9/4/17 &           6/5/19 &          10/6/14 &                 13/4/13 &           4/8/18 &            --/--/-- &                    15/7/8 \\
Win/T/L Acc vs Tab.. &          6/0/24 &          9/1/20 &          11/0/19 &          11/2/17 &                 12/0/18 &           6/3/21 &            --/--/-- &                    19/3/8 \\
Win/T/L CE vs TabP.. &          6/0/24 &          8/0/22 &           8/0/22 &           8/0/22 &                 20/0/10 &           1/4/25 &            --/--/-- &                    23/0/7 \\
\midrule
Mean AUC OVO
                                         &  0.884$\pm$.012 &   0.89$\pm$.011 &   0.891$\pm$.011 &    0.894$\pm$.01 &           0.895$\pm$.01 &    0.891$\pm$.01 &    0.894$\pm$.01 &  \textbf{0.898}$\pm$.0097 \\
Mean Acc.            &  0.815$\pm$.014 &  0.818$\pm$.011 &   0.821$\pm$.013 &   0.821$\pm$.016 &           0.83$\pm$.012 &    0.82$\pm$.013 &   0.825$\pm$.012 &   \textbf{0.834}$\pm$.011 \\
Mean CE              &  0.782$\pm$.074 &  0.767$\pm$.061 &   0.758$\pm$.047 &    0.815$\pm$.06 &  \textbf{0.72}$\pm$.015 &   0.742$\pm$.021 &   0.732$\pm$.018 &            0.721$\pm$.015 \\
\midrule
Time Tune + Train (s) 
                                    &            3241 &            3718 &             3304 &             3601 &                    3127 &  \multirow{2}{*}{\textbf{0.0519}} &           \multirow{2}{*}{0.6172} &                      \multirow{2}{*}{3127} \\
Predict (s)  &  0.0815 & 0.0168 & 0.0685 & 1.224 & 21.18 &  &  &  \\
\bottomrule
\end{tabular}
\end{table}
\newpage

\iffalse
\begin{figure}
    \centering
    \includegraphics[width=1\textwidth]{figures/cd.png}
    \caption{Critical difference diagram using the Nemenyi test ($\alpha=0.05$). We compare all methods based on their average rank (\emph{M. rank AUC OVO} in Table~\ref{app:table:tabular_results_table}) against each other. The horizontal bar shows the critical difference and connected groups of methods do not perform significantly different.}
    \label{app:fig:cd}
\end{figure}
\fi
\subsection{Results on the \openmlautoml{} Benchmark}
\label{app:sec:automl_results}
\begin{table}[t]
    \caption{ROC AUC OVO results on the 5 small datasets ($\le$ 1\,111 examples, 100 features and 10 classes) for 60 minutes requested time per dataset and per split. If available, all baselines are given ROC AUC optimization as an objective, others optimize CE. Evaluation on this benchmark was performed with our previously released \methodname{} in order to mitigate test-set overfitting. Mean OpenML-Metric is not shown in the table as averaging over a mixture of Cross Entropy and ROC AUC is problematic (however, \methodname{} had the strongest average as well).}
    \label{app:table:tabular_results_table_small_automl}
    \centering
    \tiny
    \begin{tabular}{llllllll}
\toprule
{} &       AutoGluon &            ASKL &          ASKL2.0 & TunedRandomForest &           FLAML &            TPOT &                   TabPFN \\
\midrule
vehicle                          &         -0.3084 &         -0.3816 &          -0.3412 &           -0.4849 &         -0.4286 &         -0.3433 &         \textbf{-0.2955} \\
eucalyptus                       &         -0.6905 &         -0.7255 &          -0.6967 &           -0.7209 &         -0.7433 &         -0.7123 &          \textbf{-0.665} \\
blood-transfusion-service-center &          0.7532 &           0.749 &           0.7557 &            0.6879 &          0.7332 &          0.7359 &          \textbf{0.7593} \\
Australian                       &           0.941 &          0.9315 &  \textbf{0.9411} &            0.9394 &          0.9356 &          0.9382 &                   0.9395 \\
credit-g                         &          0.7977 &          0.7891 &           0.7984 &   \textbf{0.8017} &          0.7838 &          0.7821 &                   0.7989 \\
\midrule
Wins OpenML Metric
                                                     &               0 &               0 &                0 &                 1 &               0 &               0 &               \textbf{3} \\
Wins Acc.                        &               1 &               0 &                0 &                 1 &               0 &               1 &               \textbf{2} \\
Wins CE                          &      \textbf{3} &               0 &                0 &                 0 &               0 &               0 &                        2 \\
\midrule
M. rank OpenML Metric
                                                  &             2.4 &             5.4 &              2.6 &               4.8 &             6.2 &               5 &             \textbf{1.6} \\
Mean rank Acc.                   &             2.4 &             5.7 &              4.7 &               4.4 &             5.7 &               3 &             \textbf{2.1} \\
Mean rank CE                     &    \textbf{1.4} &             5.8 &              4.4 &               5.4 &             4.4 &             4.7 &                      1.9 \\
\midrule
Mean Acc.                        &  0.793$\pm$.031 &  0.763$\pm$.052 &   0.775$\pm$.052 &    0.763$\pm$.039 &  0.761$\pm$.035 &  0.784$\pm$.031 &  \textbf{0.794}$\pm$.033 \\
Mean CE                          &  0.454$\pm$.039 &  0.537$\pm$.061 &   0.502$\pm$.056 &    0.545$\pm$.079 &  0.499$\pm$.048 &     0.73$\pm$.7 &   \textbf{0.449}$\pm$.05 \\
\midrule
Mean time (s) 
                                                &            3182 &            3611 &             3609 &              2877 &            3600 &            3400 &           \textbf{4.374} (CPU) \\
\bottomrule
\end{tabular}
\end{table}
We evaluate \methodname{} using the official benchmarking scripts,\footnote{Available at https://github.com/openml/automlbenchmark} datasets, splits and baselines results of the \openmlautoml{} Benchmark. The full list of 5 datasets can be found in Table \ref{table:automl_small_datasets}. We note that these datasets are not disjoint from our evaluation datasets; in fact 3 of these (``credit-g'', ``vehicle'', and ``blood-transfusion-service-center'') 
%evaluated datasets from the \openmlautoml{} Benchmark 
were included in our evaluation datasets from the \openmlcc{} Benchmark as well, while one dataset (``Australian'') was included in our list of 150 meta-validation datasets. The evaluation using another set of benchmarking scripts with previously published train-test splits and baseline results, however, helps to confirm that: (1) Our baselines are well tuned and not outperformed by another method in the extensive \openmlautoml{} Benchmark; (2) the runtimes of \methodname{} are reproducible in a controlled environment provided by the \openmlautoml{} Benchmark; and (3) \methodname{} is not overfit to datasplits or our evaluation metric. While we used a 50-50 train-test split with 5 iterations for our experiments in Table \ref{app:table:tabular_results_table_small_automl}, the \openmlautoml{} Benchmark uses a 10-fold cross-validation, which results in a 90-10 splits with 10 iterations. Thus in the \openmlautoml{} Benchmark, all methods use more training samples than in our experiments on the \openmlcc{} Benchmark, which leads to slightly stronger results.
\subsection{In-depth analysis of model biases}
\label{sec:app:in_depth}
Previous work by \cite{grinsztajn2022why} empirically investigates the inductive biases of tree-based and deep-learning models. They identify three challenges in developing tabular-specific models. Models must be (1)~able to fit irregular functions; (2)~robust to uninformative features; and (3)~preserve the orientation of the data. We perform and extend this analysis in the following section. We consider three model types in our analyses: \methodname{}, GBDTs (LightGBM) and NNs (Standard Sklearn Multi Layer Perceptron with a hidden dimensionality of 100). We do not seek to make absolute comparisons between these model types, which would warrant tuning of our baselines, but only seek to investigate their qualitative behavior in the following experiments.

\subsubsection{Fitting irregular patterns}
\label{sec:app:inductive_bias}
\begin{figure}[h]
     \centering
     \hspace*{-1.5cm}
     \begin{subfigure}[b]{1.2\textwidth}
         \centering
            \includegraphics[width=1.0\columnwidth]{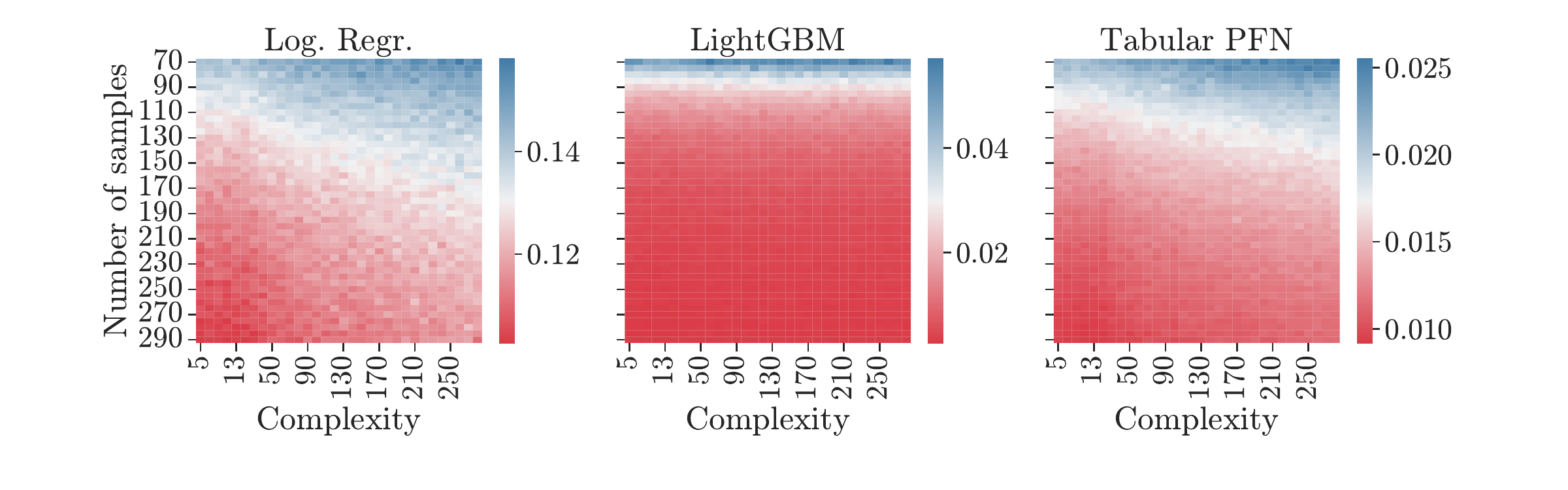}
     \end{subfigure}
     \caption{Mean training set uncertainty (cross-entropy loss) on synthetic data generated from random SCMs. The number of training samples is varied on the y-axis and the complexity of the generated data (number of hidden units in the data generating graph) on the x-axis. The cross-entropy mean is averaged across the 100 samples and 1\,000 SCMs for each point.}
     \label{fig:inductive_bias}
\end{figure}
We believe GBDT methods tend to learn non-smooth and irregular patterns in the targets, while MLPs learn smooth, low-frequency functions, as can be seen in Figure \ref{fig:toy_samples}.
%The work by \citet{grinsztajn2022why} suggests that GBDT methods perform well exactly because they are able to learn these irregular patterns.
%Prior works on regularization and an intuitive exploration of the target decision boundary of GBDT methods in Figure \ref{fig:toy_samples}, suggests, however, that smooth target functions are desirable.
TabPFN learns target functions that are rather smooth, as suggested in Figure \ref{fig:toy_samples}. This is due to our model's prior, which prefers simple SCMs as explanations and thus less irregular decision planes. We note, however, that when many training samples are provided, \methodname{} also fits more complex functions.

The tradeoff between complexity and number of training samples is explored in Figure \ref{fig:inductive_bias}. Here, we evaluate the training set cross-entropy loss on synthetic data generated from random SCMs. The number of training samples and the complexity of the generated data (number of hidden units in the data generating graph) is varied. 

\subsubsection{Robustness to uninformative features}
\label{sec:app:robustness_uninformative}
\begin{figure}[h]
     \centering
     \hspace*{-1cm}
     \begin{subfigure}[t]{0.31\textwidth}
         \centering
            % This file was created with tikzplotlib v0.9.12.
\begin{tikzpicture}

\definecolor{color0}{rgb}{0.194607843137255,0.453431372549019,0.632843137254902}
\definecolor{color1}{rgb}{0.881862745098039,0.505392156862745,0.173039215686275}

\begin{axis}[
height=5cm,
legend cell align={left},
legend style={
  fill opacity=0.8,
  draw opacity=1,
  text opacity=1,
  at={(0.03,0.97)},
  anchor=north west,
  draw=white!80!black
},
minor xtick={},
minor ytick={},
tick align=outside,
tick pos=left,
width=5cm,
x grid style={white!69.0196078431373!black},
xmin=-0.5, xmax=2.5,
xtick style={color=black},
xtick={0,1,2},
xticklabels={L-GBM,MLP,TabPFN},
y grid style={white!69.0196078431373!black},
ylabel={ROC AUC},
xlabel={Method},
ymin=0.820631257415432, ymax=0.880344032994095,
ytick style={color=black},
ytick={0.82,0.83,0.84,0.85,0.86,0.87,0.88,0.89}
]
\path [draw=white!24.7058823529412!black, fill=color0, line width=1.5pt]
(axis cs:-0.396,0.839283206834253)
--(axis cs:-0.004,0.839283206834253)
--(axis cs:-0.004,0.845459069299713)
--(axis cs:-0.396,0.845459069299713)
--(axis cs:-0.396,0.839283206834253)
--cycle;
\path [draw=white!24.7058823529412!black, fill=color1, line width=1.5pt]
(axis cs:0.004,0.827779879893938)
--(axis cs:0.396,0.827779879893938)
--(axis cs:0.396,0.829535075307709)
--(axis cs:0.004,0.829535075307709)
--(axis cs:0.004,0.827779879893938)
--cycle;
\path [draw=white!24.7058823529412!black, fill=color0, line width=1.5pt]
(axis cs:0.604,0.846067932833289)
--(axis cs:0.996,0.846067932833289)
--(axis cs:0.996,0.854047678610379)
--(axis cs:0.604,0.854047678610379)
--(axis cs:0.604,0.846067932833289)
--cycle;
\path [draw=white!24.7058823529412!black, fill=color1, line width=1.5pt]
(axis cs:1.004,0.847822966830471)
--(axis cs:1.396,0.847822966830471)
--(axis cs:1.396,0.855228245839174)
--(axis cs:1.004,0.855228245839174)
--(axis cs:1.004,0.847822966830471)
--cycle;
\path [draw=white!24.7058823529412!black, fill=color0, line width=1.5pt]
(axis cs:1.604,0.865144415653096)
--(axis cs:1.996,0.865144415653096)
--(axis cs:1.996,0.871109246693655)
--(axis cs:1.604,0.871109246693655)
--(axis cs:1.604,0.865144415653096)
--cycle;
\path [draw=white!24.7058823529412!black, fill=color1, line width=1.5pt]
(axis cs:2.004,0.861240150457158)
--(axis cs:2.396,0.861240150457158)
--(axis cs:2.396,0.867869814802835)
--(axis cs:2.004,0.867869814802835)
--(axis cs:2.004,0.861240150457158)
--cycle;
\draw[draw=white!24.7058823529412!black,fill=color0,line width=0.75pt] (axis cs:0,0) rectangle (axis cs:0,0);
\addlegendimage{ybar,ybar legend,draw=white!24.7058823529412!black,fill=color0,line width=0.75pt}
\addlegendentry{False}

\draw[draw=white!24.7058823529412!black,fill=color1,line width=0.75pt] (axis cs:0,0) rectangle (axis cs:0,0);
\addlegendimage{ybar,ybar legend,draw=white!24.7058823529412!black,fill=color1,line width=0.75pt}
\addlegendentry{True}

\addplot [line width=1.5pt, white!24.7058823529412!black, forget plot]
table {%
-0.2 0.839283206834253
-0.2 0.836526243835299
};
\addplot [line width=1.5pt, white!24.7058823529412!black, forget plot]
table {%
-0.2 0.845459069299713
-0.2 0.846313241132851
};
\addplot [line width=1.5pt, white!24.7058823529412!black, forget plot]
table {%
-0.298 0.836526243835299
-0.102 0.836526243835299
};
\addplot [line width=1.5pt, white!24.7058823529412!black, forget plot]
table {%
-0.298 0.846313241132851
-0.102 0.846313241132851
};
\addplot [line width=1.5pt, white!24.7058823529412!black, forget plot]
table {%
0.2 0.827779879893938
0.2 0.827779879893938
};
\addplot [line width=1.5pt, white!24.7058823529412!black, forget plot]
table {%
0.2 0.829535075307709
0.2 0.830423097443697
};
\addplot [line width=1.5pt, white!24.7058823529412!black, forget plot]
table {%
0.102 0.827779879893938
0.298 0.827779879893938
};
\addplot [line width=1.5pt, white!24.7058823529412!black, forget plot]
table {%
0.102 0.830423097443697
0.298 0.830423097443697
};
\addplot [line width=1pt, black, mark=diamond*, mark size=2.5, mark options={solid,fill=white!24.7058823529412!black}, only marks, forget plot]
table {%
0.2 0.823345474487189
};
\addplot [line width=1.5pt, white!24.7058823529412!black, forget plot]
table {%
0.8 0.846067932833289
0.8 0.845016687254403
};
\addplot [line width=1.5pt, white!24.7058823529412!black, forget plot]
table {%
0.8 0.854047678610379
0.8 0.854244748486864
};
\addplot [line width=1.5pt, white!24.7058823529412!black, forget plot]
table {%
0.702 0.845016687254403
0.898 0.845016687254403
};
\addplot [line width=1.5pt, white!24.7058823529412!black, forget plot]
table {%
0.702 0.854244748486864
0.898 0.854244748486864
};
\addplot [line width=1.5pt, white!24.7058823529412!black, forget plot]
table {%
1.2 0.847822966830471
1.2 0.846719189439211
};
\addplot [line width=1.5pt, white!24.7058823529412!black, forget plot]
table {%
1.2 0.855228245839174
1.2 0.855467189325167
};
\addplot [line width=1.5pt, white!24.7058823529412!black, forget plot]
table {%
1.102 0.846719189439211
1.298 0.846719189439211
};
\addplot [line width=1.5pt, white!24.7058823529412!black, forget plot]
table {%
1.102 0.855467189325167
1.298 0.855467189325167
};
\addplot [line width=1.5pt, white!24.7058823529412!black, forget plot]
table {%
1.8 0.865144415653096
1.8 0.86284723505677
};
\addplot [line width=1.5pt, white!24.7058823529412!black, forget plot]
table {%
1.8 0.871109246693655
1.8 0.877629815922338
};
\addplot [line width=1.5pt, white!24.7058823529412!black, forget plot]
table {%
1.702 0.86284723505677
1.898 0.86284723505677
};
\addplot [line width=1.5pt, white!24.7058823529412!black, forget plot]
table {%
1.702 0.877629815922338
1.898 0.877629815922338
};
\addplot [line width=1.5pt, white!24.7058823529412!black, forget plot]
table {%
2.2 0.861240150457158
2.2 0.860362845071157
};
\addplot [line width=1.5pt, white!24.7058823529412!black, forget plot]
table {%
2.2 0.867869814802835
2.2 0.872200071925264
};
\addplot [line width=1.5pt, white!24.7058823529412!black, forget plot]
table {%
2.102 0.860362845071157
2.298 0.860362845071157
};
\addplot [line width=1.5pt, white!24.7058823529412!black, forget plot]
table {%
2.102 0.872200071925264
2.298 0.872200071925264
};
\addplot [line width=1.5pt, white!24.7058823529412!black, forget plot]
table {%
-0.396 0.844532517954297
-0.004 0.844532517954297
};
\addplot [line width=1.5pt, white!24.7058823529412!black, forget plot]
table {%
0.004 0.828256451440518
0.396 0.828256451440518
};
\addplot [line width=1.5pt, white!24.7058823529412!black, forget plot]
table {%
0.604 0.850425642353404
0.996 0.850425642353404
};
\addplot [line width=1.5pt, white!24.7058823529412!black, forget plot]
table {%
1.004 0.851128656163645
1.396 0.851128656163645
};
\addplot [line width=1.5pt, white!24.7058823529412!black, forget plot]
table {%
1.604 0.868623148741197
1.996 0.868623148741197
};
\addplot [line width=1.5pt, white!24.7058823529412!black, forget plot]
table {%
2.004 0.864306233963193
2.396 0.864306233963193
};
\end{axis}

\end{tikzpicture}
            \label{fig:inductive_bias_graphs:rotate}
            \vspace*{0.5cm}
     \end{subfigure}
     \hspace*{0.5cm}
     \begin{subfigure}[t]{0.31\textwidth}
         \centering
            % This file was created with tikzplotlib v0.9.12.
\begin{tikzpicture}

\definecolor{color0}{rgb}{0.12156862745098,0.466666666666667,0.705882352941177}
\definecolor{color1}{rgb}{1,0.498039215686275,0.0549019607843137}
\definecolor{color2}{rgb}{0.172549019607843,0.627450980392157,0.172549019607843}

\begin{axis}[
height=5cm,
minor xtick={},
minor ytick={},
tick align=outside,
tick pos=left,
width=5cm,
x grid style={white!69.0196078431373!black},
xlabel={Percentage of features dropped},
xmin=-5, xmax=105,
xtick style={color=black},
xtick={-20,0,20,40,60,80,100,120},
xticklabels={-20,0,20,40,60,80,100,120},
y grid style={white!69.0196078431373!black},
ylabel={},
ymin=0.675275521046433, ymax=0.888449808631818,
ytick style={color=black},
ytick={0.675,0.7,0.725,0.75,0.775,0.8,0.825,0.85,0.875,0.9},
yticklabels={0.675,0.7,0.725,0.75,0.775,0.8,0.825,0.85,0.875,0.9}
]
\path [draw=color0, fill=color0, opacity=0.2, line width=1pt]
(axis cs:0,0.856902428931779)
--(axis cs:0,0.84729100536605)
--(axis cs:5,0.847200234134778)
--(axis cs:10,0.847665608477598)
--(axis cs:15,0.847119512691201)
--(axis cs:20,0.839394085379111)
--(axis cs:25,0.839910928617127)
--(axis cs:30,0.836339733943386)
--(axis cs:35,0.835072037006976)
--(axis cs:40,0.830284044409429)
--(axis cs:45,0.82976114980455)
--(axis cs:50,0.824929096139139)
--(axis cs:55,0.824302957553855)
--(axis cs:60,0.823228432901778)
--(axis cs:65,0.80715942278694)
--(axis cs:70,0.803929162537415)
--(axis cs:75,0.793761570789738)
--(axis cs:80,0.786647504292885)
--(axis cs:85,0.778418270825223)
--(axis cs:90,0.75936859152184)
--(axis cs:95,0.726765522643453)
--(axis cs:100,0.684965261391223)
--(axis cs:100,0.689852175641733)
--(axis cs:100,0.689852175641733)
--(axis cs:95,0.73841058970971)
--(axis cs:90,0.765543993471725)
--(axis cs:85,0.781337484471042)
--(axis cs:80,0.795215583955737)
--(axis cs:75,0.804655930770704)
--(axis cs:70,0.810256037015294)
--(axis cs:65,0.816203851959224)
--(axis cs:60,0.829024814773027)
--(axis cs:55,0.831200882824506)
--(axis cs:50,0.834012817196972)
--(axis cs:45,0.83913029400619)
--(axis cs:40,0.840634160118441)
--(axis cs:35,0.844110948399072)
--(axis cs:30,0.846188402575128)
--(axis cs:25,0.850918764078066)
--(axis cs:20,0.852814903966143)
--(axis cs:15,0.857918162613545)
--(axis cs:10,0.857448810067436)
--(axis cs:5,0.857008546558717)
--(axis cs:0,0.856902428931779)
--cycle;

\path [draw=color1, fill=color1, opacity=0.2, line width=1pt]
(axis cs:0,0.850339215432427)
--(axis cs:0,0.843520382865156)
--(axis cs:5,0.844082692114669)
--(axis cs:10,0.843632489306317)
--(axis cs:15,0.844695988604911)
--(axis cs:20,0.841047623882437)
--(axis cs:25,0.839869174439388)
--(axis cs:30,0.840835805310361)
--(axis cs:35,0.838883622109217)
--(axis cs:40,0.835884446733253)
--(axis cs:45,0.836517917294521)
--(axis cs:50,0.83198794830688)
--(axis cs:55,0.828033979328791)
--(axis cs:60,0.82425992154173)
--(axis cs:65,0.808574291519257)
--(axis cs:70,0.805166917210463)
--(axis cs:75,0.797758476548147)
--(axis cs:80,0.794193339850734)
--(axis cs:85,0.78377423498523)
--(axis cs:90,0.76356124623194)
--(axis cs:95,0.730602973355038)
--(axis cs:100,0.685879107544817)
--(axis cs:100,0.705824815211724)
--(axis cs:100,0.705824815211724)
--(axis cs:95,0.744643347206717)
--(axis cs:90,0.772201454003929)
--(axis cs:85,0.790544195100513)
--(axis cs:80,0.803751134989285)
--(axis cs:75,0.809004770794495)
--(axis cs:70,0.815769410861127)
--(axis cs:65,0.820222452940233)
--(axis cs:60,0.833926803361926)
--(axis cs:55,0.835823934320861)
--(axis cs:50,0.837719288034716)
--(axis cs:45,0.838033823205144)
--(axis cs:40,0.838371975046907)
--(axis cs:35,0.841477116165042)
--(axis cs:30,0.843741257306726)
--(axis cs:25,0.845886683098284)
--(axis cs:20,0.847518361652638)
--(axis cs:15,0.851060557005599)
--(axis cs:10,0.851961785724521)
--(axis cs:5,0.851349926274876)
--(axis cs:0,0.850339215432427)
--cycle;

\path [draw=color2, fill=color2, opacity=0.2, line width=1pt]
(axis cs:0,0.878490670892505)
--(axis cs:0,0.867998719586815)
--(axis cs:5,0.868456764814183)
--(axis cs:10,0.868644159563406)
--(axis cs:15,0.868262825191936)
--(axis cs:20,0.861051528564019)
--(axis cs:25,0.861571426585242)
--(axis cs:30,0.857720030054025)
--(axis cs:35,0.85635471034646)
--(axis cs:40,0.855072527486541)
--(axis cs:45,0.853882165712742)
--(axis cs:50,0.847703065300619)
--(axis cs:55,0.844410630552543)
--(axis cs:60,0.841627652340724)
--(axis cs:65,0.827518661691187)
--(axis cs:70,0.824642643042446)
--(axis cs:75,0.814991649643608)
--(axis cs:80,0.808827967587808)
--(axis cs:85,0.797398706236222)
--(axis cs:90,0.777858725568905)
--(axis cs:95,0.738774031730206)
--(axis cs:100,0.69157053616534)
--(axis cs:100,0.706213956865073)
--(axis cs:100,0.706213956865073)
--(axis cs:95,0.751214527599549)
--(axis cs:90,0.781444055574186)
--(axis cs:85,0.803579975652772)
--(axis cs:80,0.819291348541781)
--(axis cs:75,0.824976781719742)
--(axis cs:70,0.832181424173162)
--(axis cs:65,0.836073687581906)
--(axis cs:60,0.847872568542848)
--(axis cs:55,0.849410636528009)
--(axis cs:50,0.851055545233346)
--(axis cs:45,0.855206772070129)
--(axis cs:40,0.85698460478678)
--(axis cs:35,0.860055056495843)
--(axis cs:30,0.861061830499053)
--(axis cs:25,0.870704598375953)
--(axis cs:20,0.871469552146714)
--(axis cs:15,0.878175901861056)
--(axis cs:10,0.878760068287028)
--(axis cs:5,0.878735269348342)
--(axis cs:0,0.878490670892505)
--cycle;

\addplot [line width=1.5pt, color0]
table {%
0 0.852088927735841
5 0.852104390346748
10 0.852579018236213
15 0.852589916128979
20 0.846152221623859
25 0.845414846347596
30 0.841293245820308
35 0.839597114702238
40 0.835880255856965
45 0.834881986855459
50 0.829909385198176
55 0.828135967731404
60 0.826897245041326
65 0.812270819540884
70 0.807096608924323
75 0.799563011585109
80 0.791610083830478
85 0.779739220167059
90 0.762393248773244
95 0.73162305424915
100 0.687404643673194
};
\addplot [line width=1.5pt, color1]
table {%
0 0.846929799148791
5 0.847690527845026
10 0.847797137515419
15 0.847979659318739
20 0.844282992767538
25 0.842802064916555
30 0.84234533003077
35 0.840046260812856
40 0.837082982036658
45 0.83721560683692
50 0.835157676804564
55 0.831622738825043
60 0.828927717677472
65 0.814515681754767
70 0.810479113631191
75 0.804225676488146
80 0.798992860204137
85 0.787166962046706
90 0.767890121382196
95 0.737906591168442
100 0.695261616726309
};
\addplot [line width=1.5pt, color2]
table {%
0 0.873376745131013
5 0.873696324950249
10 0.873799024894695
15 0.873280568568337
20 0.865735432663437
25 0.865483365718803
30 0.859391834248346
35 0.858186813618279
40 0.85601169463139
45 0.85437573723066
50 0.849430303195626
55 0.846912996349784
60 0.844750110441786
65 0.831662133111193
70 0.828739759886641
75 0.820659303795298
80 0.813899451628657
85 0.800443974205158
90 0.779665579237574
95 0.745168454639547
100 0.698986221496302
};
\draw (axis cs:50,0.828780386148321) node[
  scale=0.6,
  anchor=base west,
  text=color0,
  rotate=0.0
]{lgb};
\draw (axis cs:50,0.839283966784877) node[
  scale=0.6,
  anchor=base west,
  text=color1,
  rotate=0.0
]{mlp};
\draw (axis cs:50,0.850130067245646) node[
  scale=0.6,
  anchor=base west,
  text=color2,
  rotate=0.0
]{tabpfn};
\end{axis}

\end{tikzpicture}
            \label{fig:inductive_bias_graphs:remove}
     \end{subfigure}
     \hspace*{0.5cm}
     \begin{subfigure}[t]{0.31\textwidth}
         \centering
            % This file was created with tikzplotlib v0.9.12.
\begin{tikzpicture}

\definecolor{color0}{rgb}{0.12156862745098,0.466666666666667,0.705882352941177}
\definecolor{color1}{rgb}{1,0.498039215686275,0.0549019607843137}
\definecolor{color2}{rgb}{0.172549019607843,0.627450980392157,0.172549019607843}

\begin{axis}[
height=5cm,
minor xtick={},
minor ytick={},
tick align=outside,
tick pos=left,
width=5cm,
x grid style={white!69.0196078431373!black},
xlabel={Percentage of features added},
xmin=-4.75, xmax=99.75,
xtick style={color=black},
xtick={-20,0,20,40,60,80,100},
xticklabels={-20,0,20,40,60,80,100},
y grid style={white!69.0196078431373!black},
ylabel={},
ymin=0.81387641849892, ymax=0.881589875196131,
ytick style={color=black},
ytick={0.81,0.82,0.83,0.84,0.85,0.86,0.87,0.88,0.89},
yticklabels={0.81,0.82,0.83,0.84,0.85,0.86,0.87,0.88,0.89}
]
\path [draw=color0, fill=color0, opacity=0.2, line width=1pt]
(axis cs:0,0.85743889864232)
--(axis cs:0,0.847954982867466)
--(axis cs:5,0.849093221800157)
--(axis cs:10,0.849396748456105)
--(axis cs:15,0.848320078633253)
--(axis cs:20,0.847723703909668)
--(axis cs:25,0.846059911984594)
--(axis cs:30,0.84631489259038)
--(axis cs:35,0.8473937331532)
--(axis cs:40,0.844947748511561)
--(axis cs:45,0.843703159272816)
--(axis cs:50,0.842964031768516)
--(axis cs:55,0.843033232987134)
--(axis cs:60,0.84239217060226)
--(axis cs:65,0.842013229320782)
--(axis cs:70,0.844530021648885)
--(axis cs:75,0.844193024581713)
--(axis cs:80,0.846294985750166)
--(axis cs:85,0.841248360033511)
--(axis cs:90,0.840652395425494)
--(axis cs:95,0.841853601302782)
--(axis cs:95,0.847321879742029)
--(axis cs:95,0.847321879742029)
--(axis cs:90,0.850320810816097)
--(axis cs:85,0.851518882329551)
--(axis cs:80,0.851964531041243)
--(axis cs:75,0.85141260870065)
--(axis cs:70,0.849081495163476)
--(axis cs:65,0.853743748761011)
--(axis cs:60,0.852224704292783)
--(axis cs:55,0.853364425498842)
--(axis cs:50,0.85251297059695)
--(axis cs:45,0.85105258362285)
--(axis cs:40,0.852876057131561)
--(axis cs:35,0.855076186971007)
--(axis cs:30,0.855793408797835)
--(axis cs:25,0.855319445402335)
--(axis cs:20,0.85453386927309)
--(axis cs:15,0.85524604942836)
--(axis cs:10,0.858474066337711)
--(axis cs:5,0.858329558254911)
--(axis cs:0,0.85743889864232)
--cycle;

\path [draw=color1, fill=color1, opacity=0.2, line width=1pt]
(axis cs:0,0.852096062034849)
--(axis cs:0,0.844416935777793)
--(axis cs:5,0.840956643123248)
--(axis cs:10,0.839748141489893)
--(axis cs:15,0.837417861437685)
--(axis cs:20,0.831943520586532)
--(axis cs:25,0.835378244012882)
--(axis cs:30,0.833606940379796)
--(axis cs:35,0.834346107917764)
--(axis cs:40,0.830370602935838)
--(axis cs:45,0.828409791867599)
--(axis cs:50,0.827736083183949)
--(axis cs:55,0.828891494754643)
--(axis cs:60,0.821153209203838)
--(axis cs:65,0.823542370680088)
--(axis cs:70,0.821755456774422)
--(axis cs:75,0.822811288271735)
--(axis cs:80,0.819092456901473)
--(axis cs:85,0.823823147182259)
--(axis cs:90,0.816954302894247)
--(axis cs:95,0.817631373734494)
--(axis cs:95,0.829706512896885)
--(axis cs:95,0.829706512896885)
--(axis cs:90,0.828717112267285)
--(axis cs:85,0.832537872814202)
--(axis cs:80,0.831800147569687)
--(axis cs:75,0.828880130890504)
--(axis cs:70,0.829124179804199)
--(axis cs:65,0.831817190564369)
--(axis cs:60,0.835175537184689)
--(axis cs:55,0.833877088816369)
--(axis cs:50,0.834205273983231)
--(axis cs:45,0.835994881065331)
--(axis cs:40,0.839677968825102)
--(axis cs:35,0.839872872349739)
--(axis cs:30,0.840587147377339)
--(axis cs:25,0.843902332898474)
--(axis cs:20,0.845171684361375)
--(axis cs:15,0.846048544081198)
--(axis cs:10,0.848106817412728)
--(axis cs:5,0.847540381234792)
--(axis cs:0,0.852096062034849)
--cycle;

\path [draw=color2, fill=color2, opacity=0.2, line width=1pt]
(axis cs:0,0.878511990800803)
--(axis cs:0,0.868841062506057)
--(axis cs:5,0.865990595723353)
--(axis cs:10,0.860535537423679)
--(axis cs:15,0.863953891412309)
--(axis cs:20,0.858140890329268)
--(axis cs:25,0.857792407577837)
--(axis cs:30,0.854167039148382)
--(axis cs:35,0.85323240376913)
--(axis cs:40,0.850829766588782)
--(axis cs:45,0.850470752205879)
--(axis cs:50,0.850635796919917)
--(axis cs:55,0.849550882731173)
--(axis cs:60,0.846982017191617)
--(axis cs:65,0.844032694361803)
--(axis cs:70,0.843880369313261)
--(axis cs:75,0.840425851206203)
--(axis cs:80,0.842163470409741)
--(axis cs:85,0.842405230512621)
--(axis cs:90,0.840588228789359)
--(axis cs:95,0.837517973523635)
--(axis cs:95,0.847189740281058)
--(axis cs:95,0.847189740281058)
--(axis cs:90,0.849701234780968)
--(axis cs:85,0.851002790461346)
--(axis cs:80,0.850027566143539)
--(axis cs:75,0.852763814522686)
--(axis cs:70,0.852623375311219)
--(axis cs:65,0.854397216827832)
--(axis cs:60,0.856388159156611)
--(axis cs:55,0.858023378322895)
--(axis cs:50,0.856597701413526)
--(axis cs:45,0.858061192089581)
--(axis cs:40,0.861270201946808)
--(axis cs:35,0.863697799346001)
--(axis cs:30,0.863283188954848)
--(axis cs:25,0.867347951878923)
--(axis cs:20,0.868958547940003)
--(axis cs:15,0.870561531625554)
--(axis cs:10,0.872362460691832)
--(axis cs:5,0.877287816392843)
--(axis cs:0,0.878511990800803)
--cycle;

\addplot [line width=1.5pt, color0]
table {%
0 0.852355878772023
5 0.853897138532055
10 0.854033959881633
15 0.851825023615965
20 0.851223899166909
25 0.850708285314488
30 0.851099514235387
35 0.851113394894131
40 0.848861169019932
45 0.847221680081534
50 0.846830154466095
55 0.848154198850323
60 0.847567356001901
65 0.847821189365747
70 0.846554183015279
75 0.847327400744157
80 0.848684286505081
85 0.846383621181531
90 0.844601549732145
95 0.844722017015143
};
\addplot [line width=1.5pt, color1]
table {%
0 0.848256498906321
5 0.844124743472766
10 0.843943670331431
15 0.841786045548859
20 0.838557602473954
25 0.84003270834902
30 0.837255534615752
35 0.837027321268773
40 0.835555602002078
45 0.832096582357061
50 0.830885530634168
55 0.831215675965427
60 0.828805074470414
65 0.827508972983092
70 0.82532802565184
75 0.826059100965828
80 0.825992877208775
85 0.827917418393206
90 0.822188795236295
95 0.824099884143514
};
\addplot [line width=1.5pt, color2]
table {%
0 0.873736905312108
5 0.871639206058098
10 0.866564551151019
15 0.867257711518932
20 0.863549719134636
25 0.862601016144257
30 0.858821088145691
35 0.859056463337462
40 0.855796589042168
45 0.85426597214773
50 0.853787156891467
55 0.853895376845374
60 0.852310205040308
65 0.848944963263983
70 0.848145709338961
75 0.846594832864445
80 0.845476635108127
85 0.847099070543091
90 0.845036232075495
95 0.842353856902347
};
\draw (axis cs:50,0.846676964308728) node[
  scale=0.6,
  anchor=base west,
  text=color0,
  rotate=0.0
]{lgb};
\draw (axis cs:50,0.836810237216175) node[
  scale=0.6,
  anchor=base west,
  text=color1,
  rotate=0.0
]{mlp};
\draw (axis cs:50,0.85844298280161) node[
  scale=0.6,
  anchor=base west,
  text=color2,
  rotate=0.0
]{tabpfn};
\end{axis}

\end{tikzpicture}
            \label{fig:inductive_bias_graphs:add}
     \end{subfigure}
     \vspace{-.5cm}
     \caption{Left: Performance of LightGBM, MLP and \methodname{} when random rotations are applied to the feature space. LightGBM loses most predictive accuracy when rotations are applied, MLP is unaffected and \methodname{} performs only slightly worse. Center: Iteratively removing features according to their  reversed importance rank obtained from a random forest leads to MLPs initially performing better and overtaking LightGBM performance. \methodname{} looses predictive accuracy and performs similar to baselines, when most features are removed. Right: Adding uninformative features leads to performance degradation of MLPs and \methodname{}, while LightGBM remains relatively constant.}
     \label{fig:inductive_bias_graphs}
\end{figure}
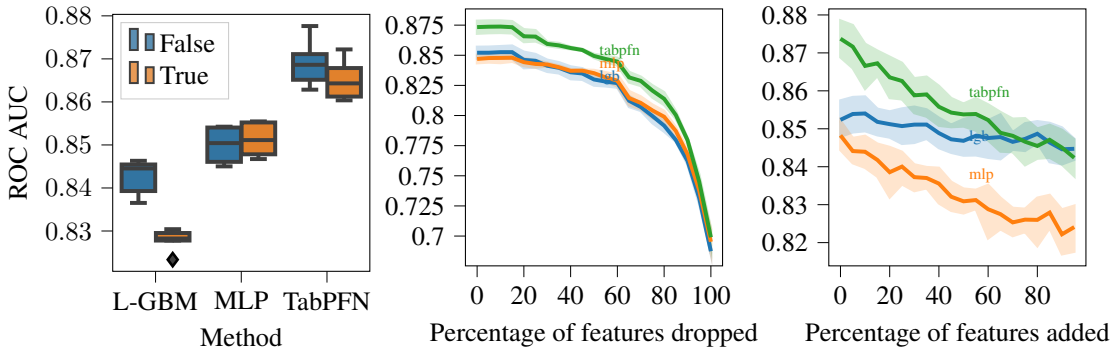
Tabular datasets contain a large fraction of uninformative features \citep{grinsztajn2022why}. To evaluate robustness to uninformative features, we add an increasingly large fraction of uninformative features to our data and show results in Figure \ref{fig:inductive_bias_graphs}. Uninformative features are generated by copying existing features and shuffling their values randomly between samples. We find that \methodname{} and MLPs are less robust to uninformative features than LightGBM. \methodname{} could be adapted by including more uninformative features in the used prior.
In a second experiment we drop an increasingly large fraction of features. We drop these features according to feature importance (ranked by a Random Forest), first removing least informative features. We show results in Figure \ref{fig:inductive_bias_graphs} (middle) and observe that the classification accuracy of a TabPFN and GBDT is not much affected by removing up to 30\% of the features, but constantly diminishes. The MLP, which is less not robust to uninformative features, performs even better when the 20\% least informative features are removed.

We use our testing tasks from the OpenML CC-18 Benchmark, and, to simplify analyses, we drop multiclass datasets and datasets that contain more than 50 features (as adding 100\% more features to a dataset with more than 50 features yields more than 100 features, which \methodname{} cannot handle).

\subsubsection{Invariance to feature rotation}
\label{sec:app:rotation_invariance}
Each feature of a tabular dataset typically carries meaning individually, as expressed by column names, such as age or sex. A learning algorithm is rotationally invariant in the sense of~\cite{ng2004feature}, if it is left unchanged when a rotation (unitary) matrix is applied to the features of both the training and testing set, i.e. when features are mixed. To remove uninformative features under a feature rotation, an algorithm first has to restore the original orientation of the features, and then select informative ones. A rotationally invariant algorithm discards the data orientation and thus has to restore the original orientation internally. Thus, \cite{ng2004feature} shows that any rotationally invariant learning algorithm has a worst-case sample complexity that grows at least linearly in the number of irrelevant features.

Figure \ref{fig:inductive_bias_graphs} (left) shows the change in test ROC AUC when randomly rotating our datasets, and confirms that only MLPs are rotationally invariant. GBDT methods are highly sensitive to rotations, while \methodname{} is less sensitive, but still performs better when no rotations are applied.% \methodname{} is likely able to reconstruct the original data orientation internally.

The theoretical results by~\cite{ng2004feature} and empirical results by~\cite{grinsztajn2022why} imply, that \methodname{}\textquotesingle{}s diminishing performance when uninformative features are added is linked to it\textquotesingle{}s relative rotation invariance. Adjusting the prior to include more uninformative features could address these results.

We use datasets from our testing tasks from the OpenML CC-18 Benchmark with numerical features only, since rotating categorical datasets is problematic, as some GBDT classifiers treat categorical variables distinctly (e.g. generating embeddings per category).

\subsection{Ablation on the selection of prior models}
\label{app: prior_ablation}
We perform ablation experiments for a prior based solely on BNNs, SCMs and a mix of both using a hyperparameter to control the sampling likelihood during prior-fitting.
The PFNs for each prior are fitted using less compute than in our final experiments, thus their scores generally are slightly worse.
The BNN prior, similar to the one used in \citet{muller-iclr22a}, provides diminished performance compared to the priors based on SCMs.
Additionally, we can see that mixing BNN and SCM prior does not seem to make a big difference on our test set compared to a pure SCM prior.

\begin{table}[htbp]
    \centering
    \begin{tabular}{llll}
\toprule
     &            BNN &            SCM &      SCM + BNN \\
\midrule
Mean CE     &  0.811$\pm$0.009 &  \textbf{0.771}$\pm$0.006 &  0.776$\pm$0.009 \\
Mean ROC AUC &  0.865$\pm$0.007 &  0.881$\pm$0.002 & \textbf{0.883}$\pm$0.003 \\
\bottomrule
\end{tabular}
    \vspace{.3cm}
    \caption{An evaluation of the impact of the prior mixing on the final performance. Our final model was trained in the \textit{SCM + BNN} setting (see Section \ref{sec:tabprior} for details on the priors).}
    \label{tab:scm bnn ablation}
\end{table}

% \subsection{Generalization Experiment}

\begin{figure}[t]
     \centering
     % This figure should have a y axis: ROC AUC Mean, I fixed it in the tex, I guess this is the downside of PDF
        %\includegraphics[width=0.5\columnwidth]{figures/model_generalization/generalization.pdf}
        % This file was created with tikzplotlib v0.9.12.
\begin{tikzpicture}

\definecolor{color0}{rgb}{0.12156862745098,0.466666666666667,0.705882352941177}

\begin{axis}[
height=5.2cm,
tick align=outside,
tick pos=left,
width=8.2cm,
x grid style={white!69.0196078431373!black},
xlabel={Training samples},
xmin=250, xmax=5000,
xtick style={color=black},
xtick={0,1000,2000,3000,4000,5000},
xticklabels={0,1000,2000,3000,4000,5000},
y grid style={white!69.0196078431373!black},
ylabel={ROC AUC Mean},
ymin=0.73, ymax=0.79,
ytick style={color=black},
ytick={0.73,0.74,0.75,0.76,0.77,0.78,0.79},
yticklabels={0.73,0.74,0.75,0.76,0.77,0.78,0.79}
]
\path [draw=color0, fill=color0, opacity=0.2]
(axis cs:250,0.734411378643683)
--(axis cs:250,0.721160496150833)
--(axis cs:500,0.738565490403955)
--(axis cs:750,0.749087998290172)
--(axis cs:1000,0.75674439354165)
--(axis cs:1250,0.764766607914957)
--(axis cs:1500,0.769822107928199)
--(axis cs:1750,0.772119384578668)
--(axis cs:2000,0.775173561155393)
--(axis cs:2250,0.776489890343141)
--(axis cs:2500,0.778498658497367)
--(axis cs:2750,0.780439635471814)
--(axis cs:3000,0.780789300209587)
--(axis cs:3250,0.780436762736754)
--(axis cs:3500,0.780919225718728)
--(axis cs:3750,0.78137467404865)
--(axis cs:4000,0.782334867610047)
--(axis cs:4250,0.782345683559777)
--(axis cs:4500,0.783002023841639)
--(axis cs:4750,0.783780506895239)
--(axis cs:5000,0.784329192748813)
--(axis cs:5000,0.786254769613164)
--(axis cs:5000,0.786254769613164)
--(axis cs:4750,0.786051249323151)
--(axis cs:4500,0.785908237410123)
--(axis cs:4250,0.785484134872577)
--(axis cs:4000,0.785210764824056)
--(axis cs:3750,0.784635161829156)
--(axis cs:3500,0.784975362487901)
--(axis cs:3250,0.784612700443988)
--(axis cs:3000,0.783586655922824)
--(axis cs:2750,0.784922822634272)
--(axis cs:2500,0.783539591883913)
--(axis cs:2250,0.781796895924172)
--(axis cs:2000,0.781471597859522)
--(axis cs:1750,0.778533159121003)
--(axis cs:1500,0.7772728233301)
--(axis cs:1250,0.773319295264854)
--(axis cs:1000,0.766533659982931)
--(axis cs:750,0.758546808702069)
--(axis cs:500,0.75145993037192)
--(axis cs:250,0.734411378643683)
--cycle;

\addplot [semithick, color0]
table {%
250 0.728152577503735
500 0.745254819861739
750 0.754210059168914
1000 0.761929119506665
1250 0.769311157817191
1500 0.773799216422216
1750 0.775300897489546
2000 0.778401980467263
2250 0.779042951030354
2500 0.78077531974305
2750 0.782500539628803
3000 0.782285942619729
3250 0.782598795832393
3500 0.782947294103314
3750 0.782963188031858
4000 0.78376420434541
4250 0.783914909216177
4500 0.784455130625881
4750 0.784915873231365
5000 0.785287064126757
};
\addplot [semithick, red, dashed]
table {%
1024 0.73
1024 0.79
};
\end{axis}

\end{tikzpicture}
     \caption{Extrapolation performance of our \methodname{} to dataset sizes never seen during training. Maximum number of samples during training was 1024 (dashed red line). The shading indicates the 95\% confidence interval over random data splits. A method that does not generalize, would be expected to flatten at 1024.}
     % Left: Differentiable Hyperparameter Tuning loss curves of both sets of validation datasets $\V_1$ and $\V_2$: We perform multiple (different colors) gradient-descent optimizations of $\phi$ on $\V_1$ from different random initializations and choose the best performer on $\V_2$. The best configuration is used on test data in our models. Right: 
    \label{fig:model generalization}
\end{figure}
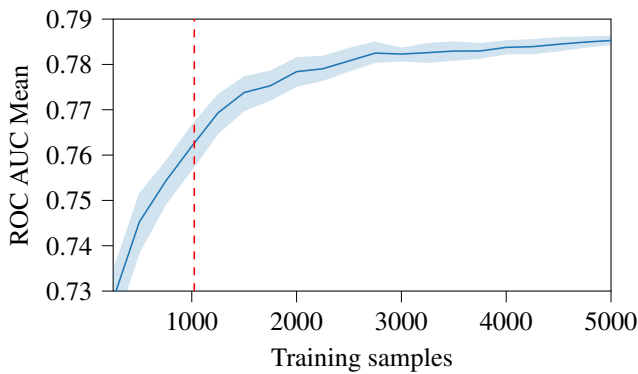

%\subsection{Ensembling Correlations}

\begin{figure}[t]
     \centering
     \hspace*{-1.1cm}
     \begin{subfigure}[b]{.52\textwidth}
         \includegraphics[width=1\columnwidth]{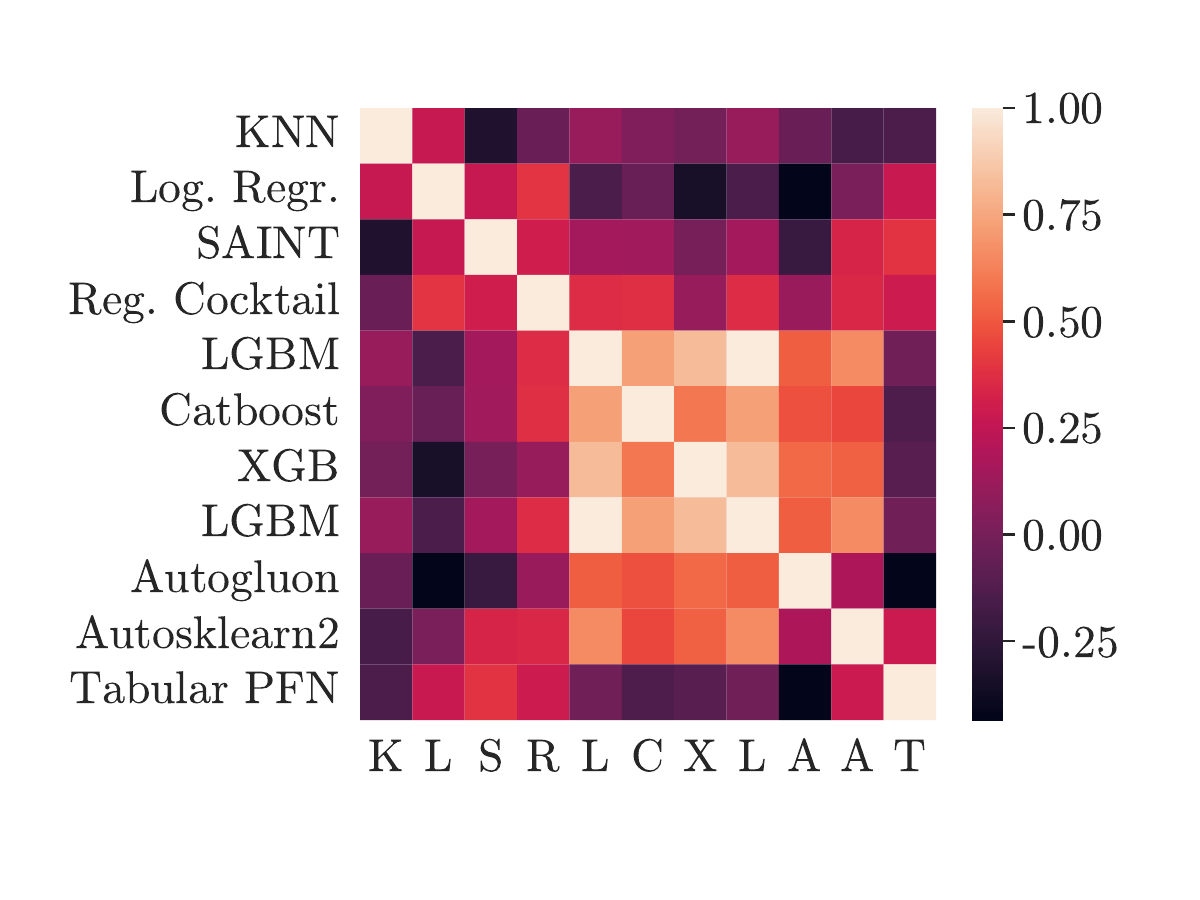}
     \end{subfigure}
     \caption{Spearman correlation of per-dataset normalized ROC AUC performance (i.e. the ranking correlation of per-dataset normalized ROC AUC scores) between the considered methods. Ordering on the x-axis is the same as on the y-axis. The \methodname{} performs well on a different set of datasets than the baselines, i.e. the Spearman correlation with the GBDT methods is low, while GBDT and AutoML methods are highly correlated, and thus perform well on the same datasets. \methodname{} correlates stronger with DL based methods (SAINT and Reg. Cocktail), which, however, do not perform as well as GBDT methods in terms of absolute ROC AUC performance (see \ref{table:tabular_results_table_small})
     }
     % Left: Differentiable Hyperparameter Tuning loss curves of both sets of validation datasets $\V_1$ and $\V_2$: We perform multiple (different colors) gradient-descent optimizations of $\phi$ on $\V_1$ from different random initializations and choose the best performer on $\V_2$. The best configuration is used on test data in our models. Right: 
    \label{fig:model correlation}
\end{figure}

\subsection{Extended analysis on a larger benchmark of datasets}
\label{app:sec:larger_results}

We now provide an extended benchmark where we include an analysis on the additional 149 validation datasets\footnote{The dataset $flags$ was removed as not enough splits could be generated by our code.} (listed in Table~\ref{table:reference_datasets_table}) in order to assess the generality of our results and to better understand its strengths and weaknesses.
We now also include comparisons with default random forests (RFs), support vector machines (SVMs), default XGBoost, etc. , along with their tuned versions after one hour. This evaluation used 5 splits and followed the same experimental setup described in Appendix \ref{app:expdetails}.

\begin{figure}[h]
     \centering
     \begin{subfigure}[h]{1.0\textwidth}
         \includegraphics[width=1\columnwidth]{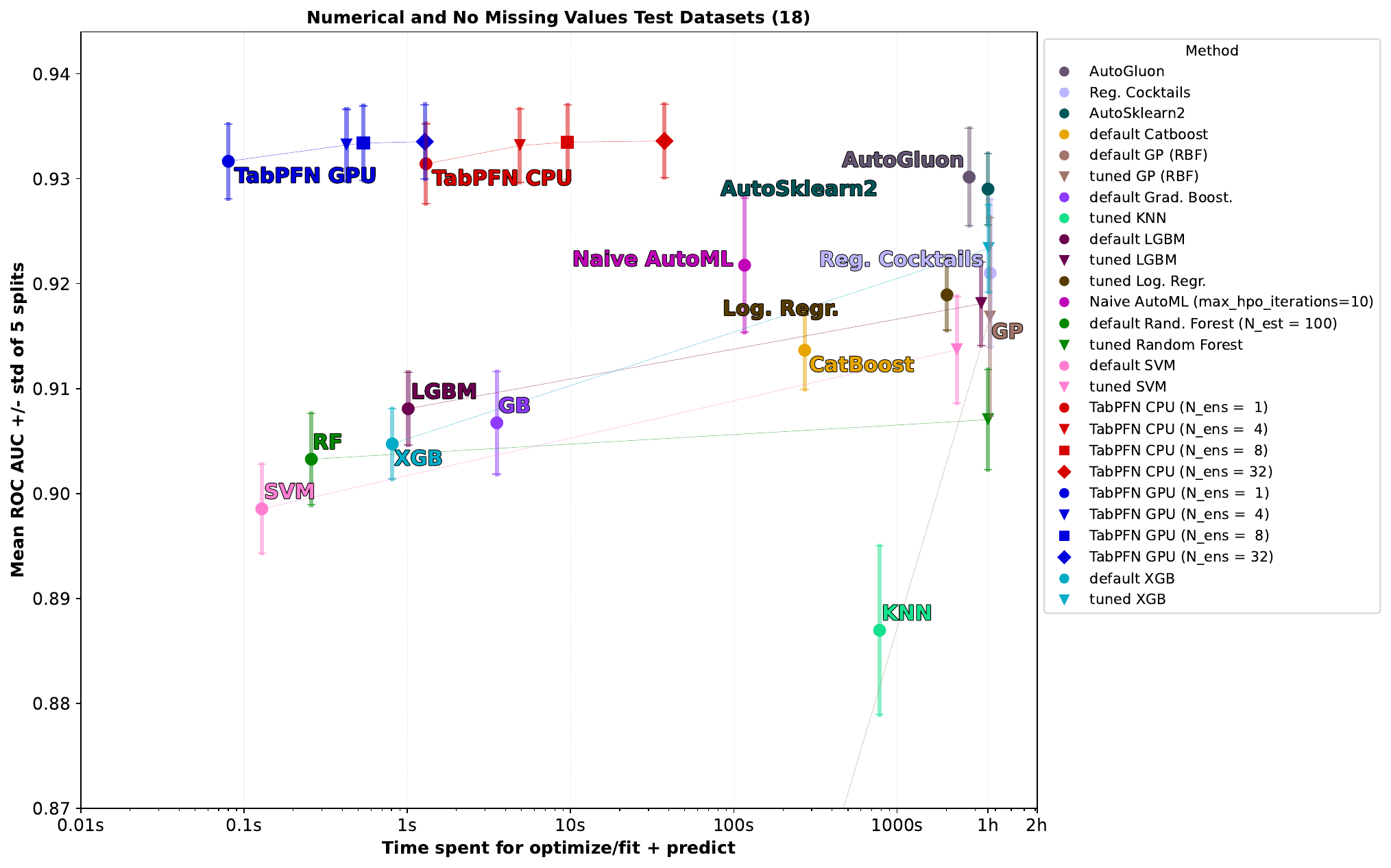}%numerical_test_datasets_opt_metric_roc_auc_opt_time_3600.0_metric_roc.pdf}
     \end{subfigure}
     \begin{subfigure}[h]{1.0\textwidth}
         \includegraphics[width=1\columnwidth]{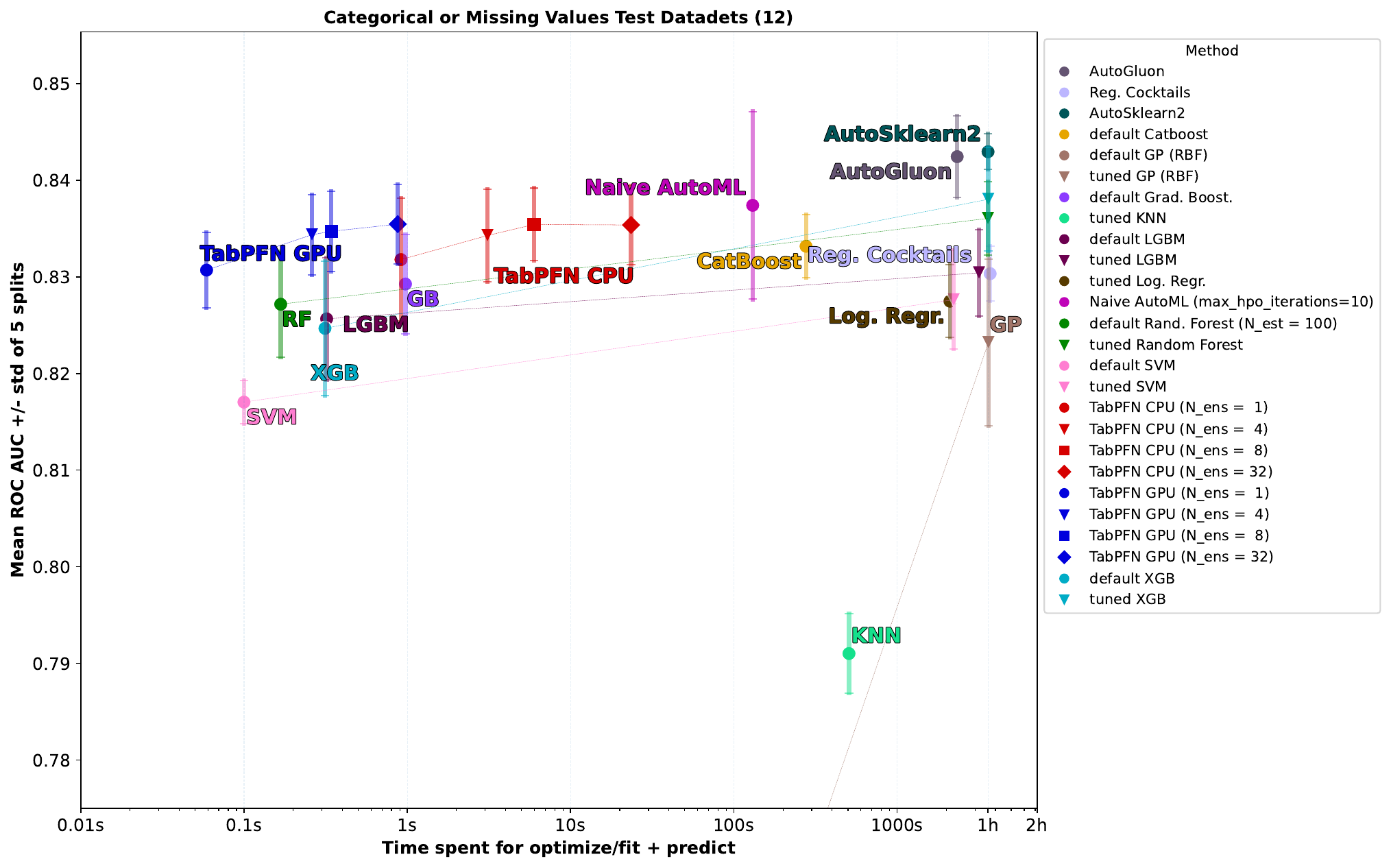}%test_datasets_w__categorical_features_opt_metric_roc_auc_opt_time_3600.0_metric_roc.pdf}
     \end{subfigure}
     \caption{ROC AUC comparison on the \openmlcc{} Benchmark. Baselines were tuned for one hour or until 10000 configurations were exhausted (Log. Reg and KNN).}
     \label{fig:new_benchmark_test}
\end{figure}

\begin{figure}[h]
     \centering
     \begin{subfigure}[h]{1.0\textwidth}
         \includegraphics[width=1\columnwidth]{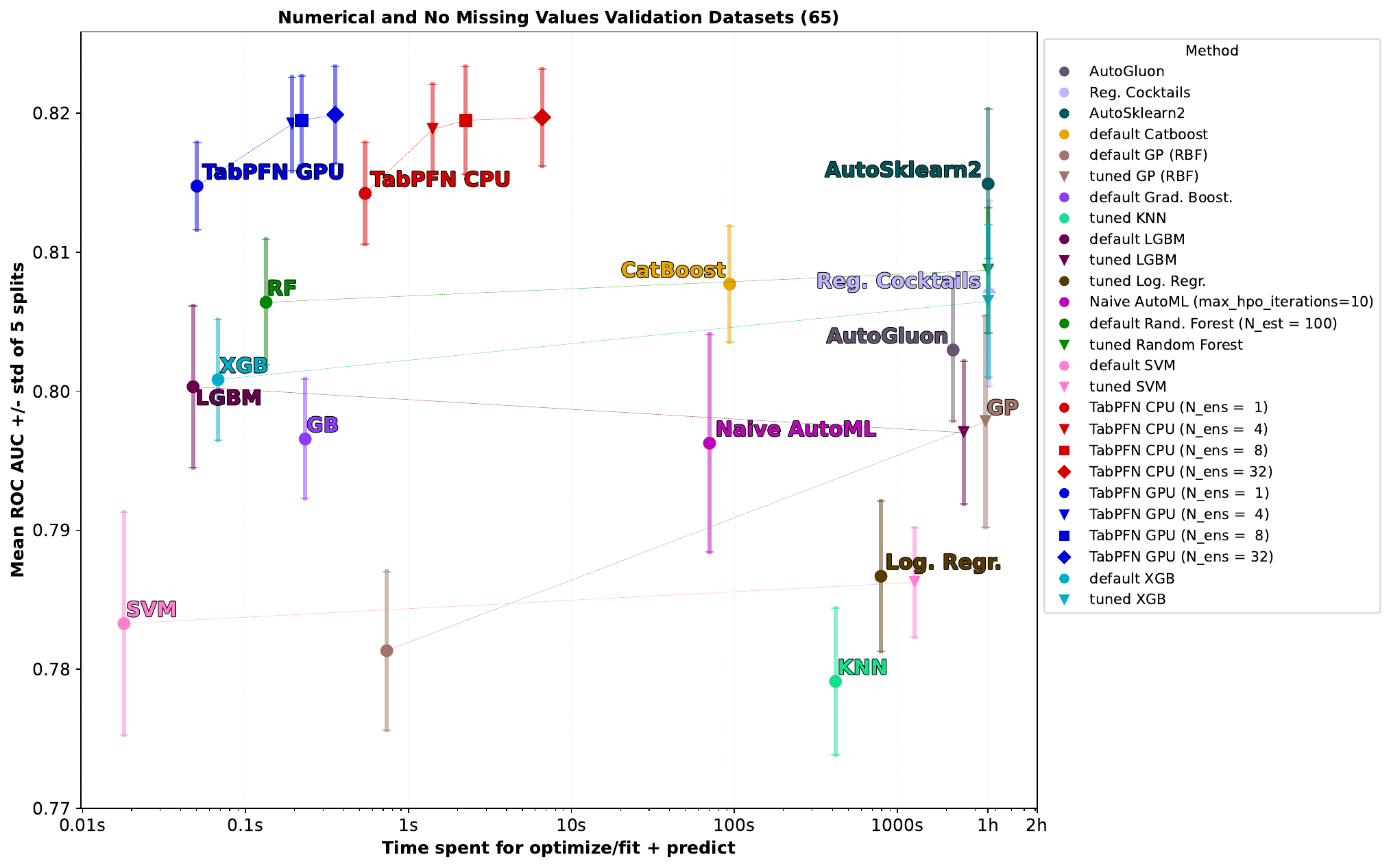}%numerical_validation_datasets_opt_metric_roc_auc_opt_time_3600.0_metric_roc.pdf}
     \end{subfigure}
     \centering
     \begin{subfigure}[h]{1.0\textwidth}
         \includegraphics[width=1\columnwidth]{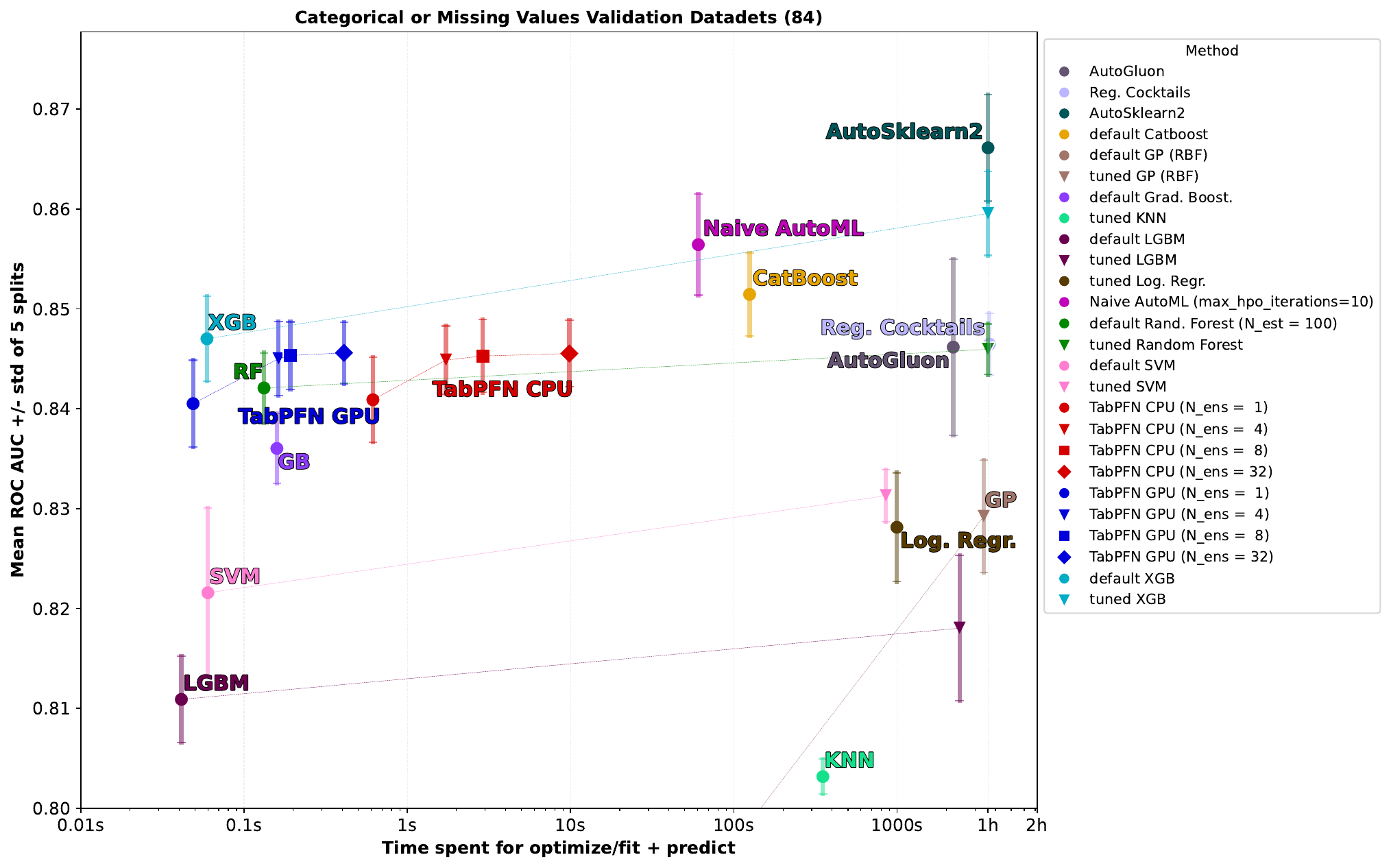}%validation_datasets_w__categorical_features_opt_metric_roc_auc_opt_time_3600.0_metric_roc.pdf}
     \end{subfigure}
     \caption{ROC AUC comparison on 149 validation datasets (see Table \ref{table:reference_datasets_table}). Baselines were tuned for one hour or until $10\,000$ configurations were exhausted (Log. Reg and KNN).}
     \label{fig:new_benchmark_valid}
\end{figure}

Figure \ref{fig:new_benchmark_test} shows results for our 30 test datasets (separated into the 18 purely numerical datasets without missing values we focus on in the main paper, and the 12 remaining ones), and Figure \ref{fig:new_benchmark_valid} shows results for the 149 validation datasets (again split into numerical datasets without missing values and the remainder). These figures demonstrate that on the purely numerical datasets without missing values, TabPFN Pareto-dominates all other methods in aggregate results across datasets, in terms of tradeoffs between ROC and time spent. For categorical datasets and datasets with missing values, it still performs well, but not nearly as well as for the numerical case without missing values we targeted.

\begin{figure}[h]
    \centering
    \includegraphics[width=0.95\columnwidth]{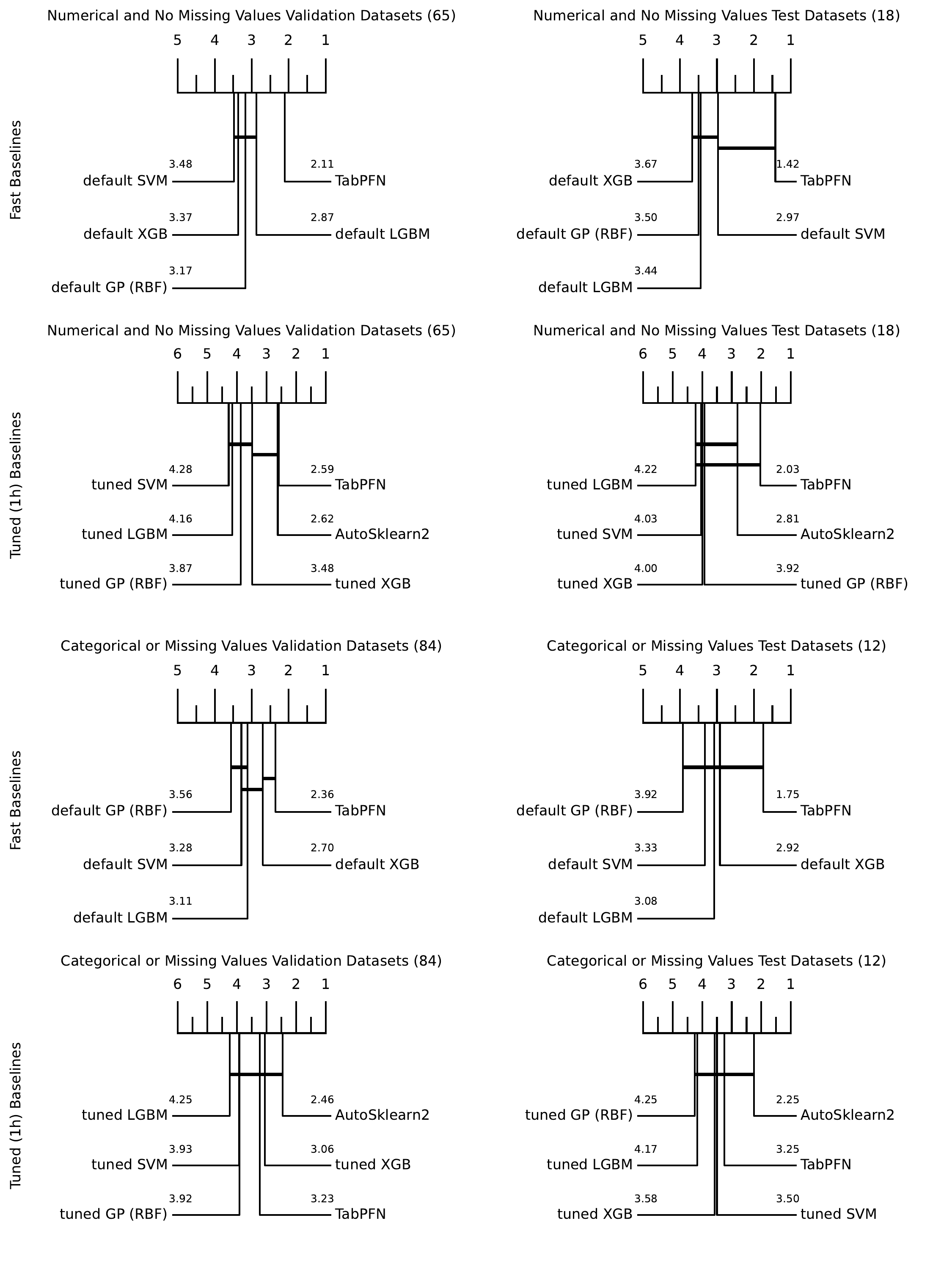}
    \caption{Critical difference plots on average ranks with a Wilcoxon significance analysis.
    We show plots for both our test set (\openmlcc{}) and our validation set.
    We split each into a subset of purely numerical datasets without missing values and the rest.
    We compare to two sets of baselines.
    i) \emph{Fast Baselines} that finish tuning, training and prediction in less than 30 seconds on average (even the TabPFN on CPU is within this bound).
    ii) \emph{Tuned Baselines} with one hour of budget for tuning, training and prediction (TabPFN still requires less than 30 seconds on average with the same hardware).
    We note that in the fast run regime, on numerical datasets without missing values, TabPFN statistically significantly outperforms all other methods.
%    For the tuned versions, the plots do not show a statistically significant difference because of their multiple testing correction. In individual 1-vs-1 comparisons, however, on numerical datasets without missing values, TabPFN (ensemble of 32) performs statistically significantly better than XGBoost.
    }
    \label{fig:cd_plots}
\end{figure}

We also asses performance using statistical tests. We note that when evaluating a new classifier, performance on one dataset is just one data point. In order to apply statistical tests, we need to perform analyses across datasets. The standard method for this are critical difference diagrams as introduced by \citet{demsar-06a}. We use the Wilcoxon rank test and correct for multiple testing using Holm–Bonferroni method. We use a a significance level $alpha = 0.05$ for our critical difference diagrams.
Figure \ref{fig:cd_plots} performs these statistical tests to compare methods in two regimes: fast runs that require up to 30 seconds and tuned runs that require up to an hour. We note again that TabPFN is much stronger on purely numerical datasets without missing features. In this case, for the short runs, it statistically significantly outperforms all other baselines. For long runs, it also statistically significantly outperforms all other methods except the AutoML frameworks in 1-vs-1 comparisons; however, this is not visible in the figure because of the multiple testing correction applied due to the multiple pairwise tests.

\clearpage

While the analysis above aggregates across datasets, variation across datasets is large and TabPFN by no means works best on \emph{all} datasets. All classifiers have worst and best cases, and we believe that it is important to complement our reports of TabPFN's strong aggregate performance with the examples we encountered for which the current version does \emph{not} do well.
We do show results for all datasets one-by-one at \url{https://github.com/automl/TabPFNResults/blob/main/individual_plots.pdf}.

\begin{figure}[h]
     \centering
     \begin{subfigure}[h]{0.69\textwidth}
         \includegraphics[width=1\columnwidth]{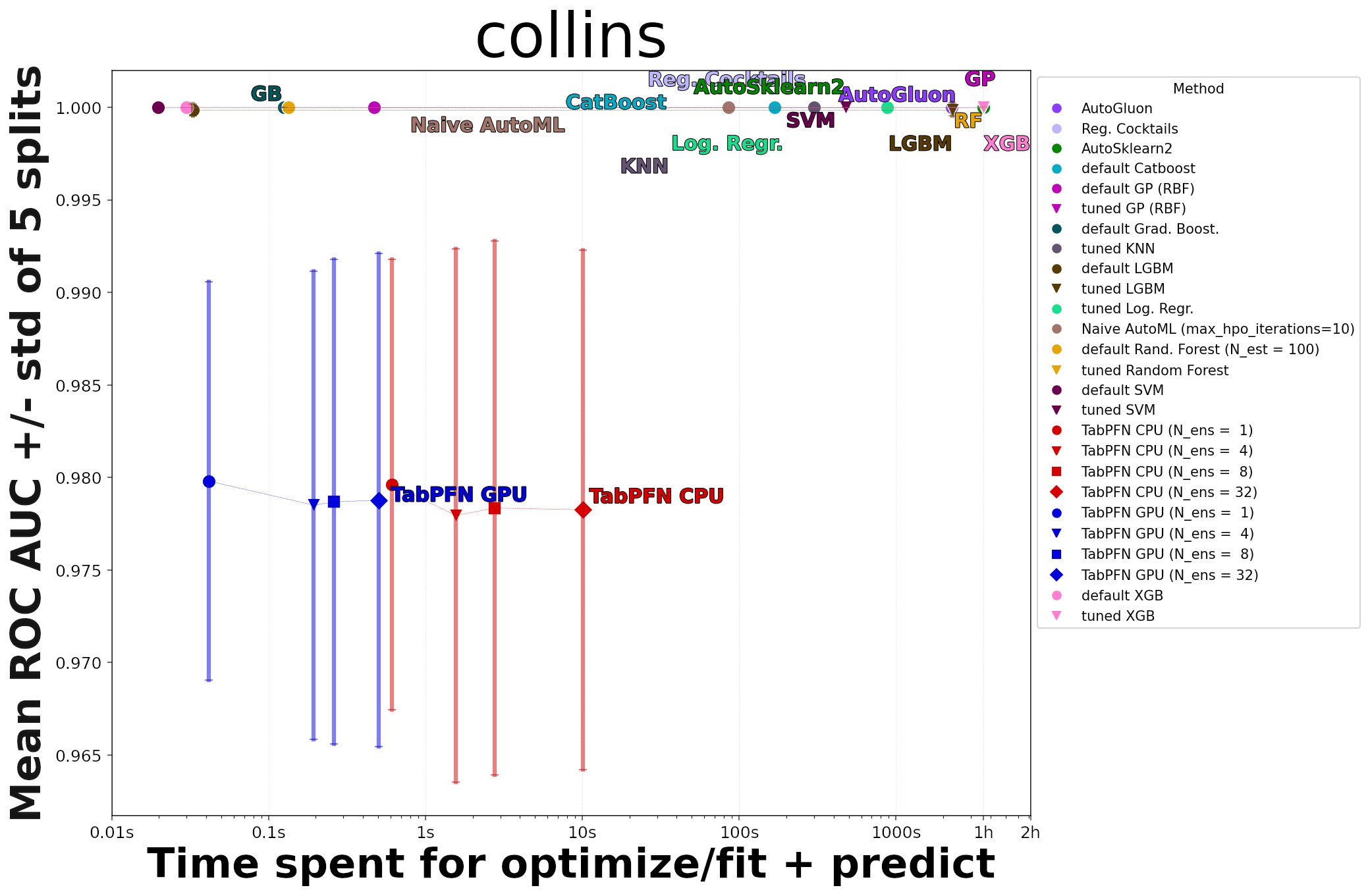}
     \end{subfigure}
     \caption{Worst result for TabPFN: \texttt{collins}\label{fig:collins}, due to uninformative features. We did not focus on uninformative features in creating TabPFN. With just the 5 most important features (as judged by a random forest), TabPFN also achieves an accuracy of 1.0.}
     \label{fig:worst_results}
\end{figure}

The dataset with the worst relative performance for TabPFN is \texttt{collins}, with results shown in Figure \ref{fig:collins}. While most other methods obtain a ROC AUC close to, or identical to 1.0, TabPFN only achieves around 98\%. We found this to be caused by uninformative features, which we have already shown to affect TabPFN's performance negatively in Section \ref{sec:app:robustness_uninformative}. With just the 5 most important features (as judged by a random forest), TabPFN also achieves an accuracy of 1.0. (In the aggregate analysis above, we of course count the original, poor performance of TabPFN, to not cheat.)

\begin{figure}[h]
     \centering
     \begin{subfigure}[h]{0.49\textwidth}
         \includegraphics[width=1\columnwidth]{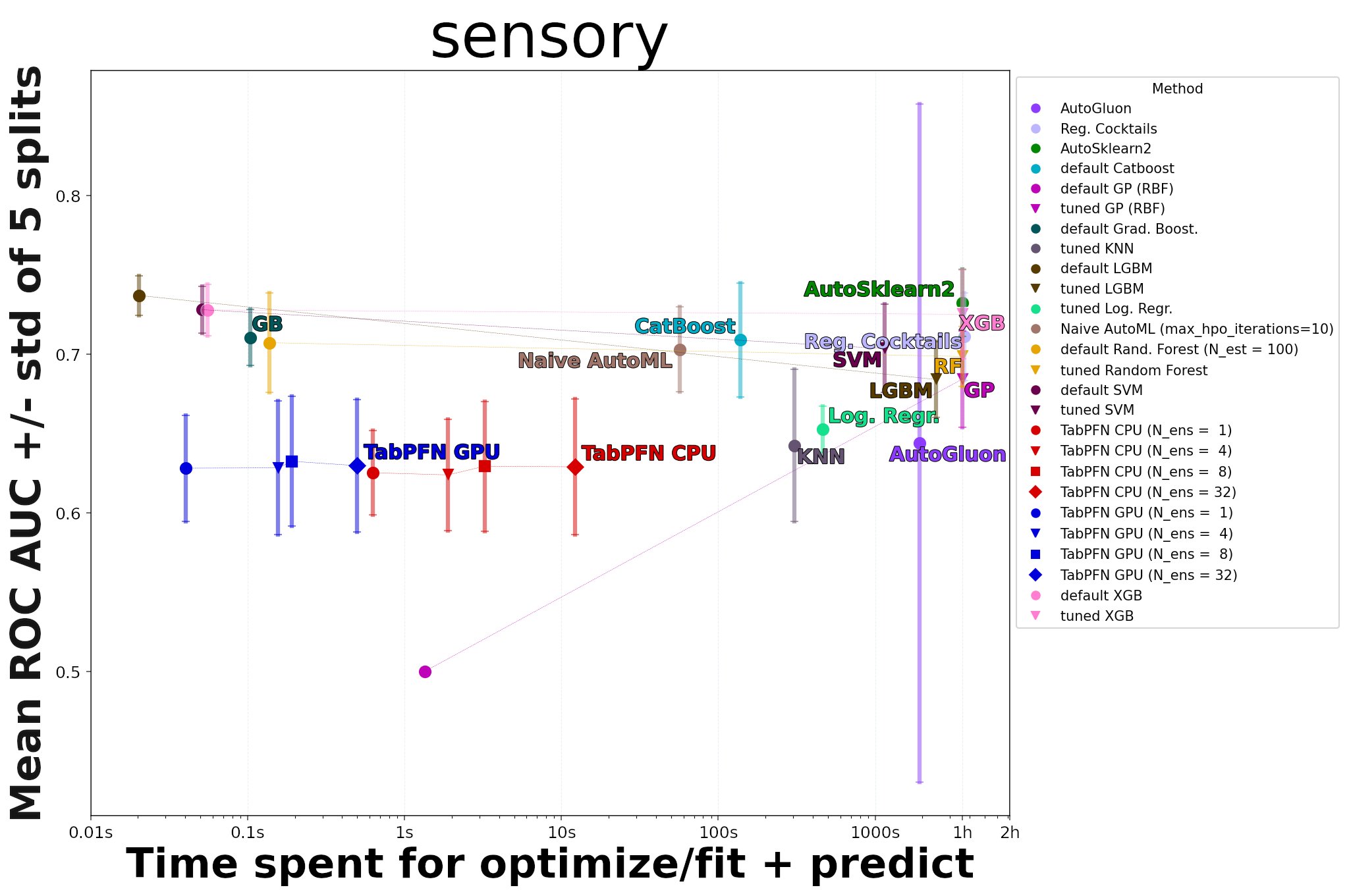}
         \caption{\texttt{sensory}, purely categorical \label{fig:sensory}}
     \end{subfigure}\hfill
     \begin{subfigure}[h]{0.49\textwidth}
         \includegraphics[width=1\columnwidth]{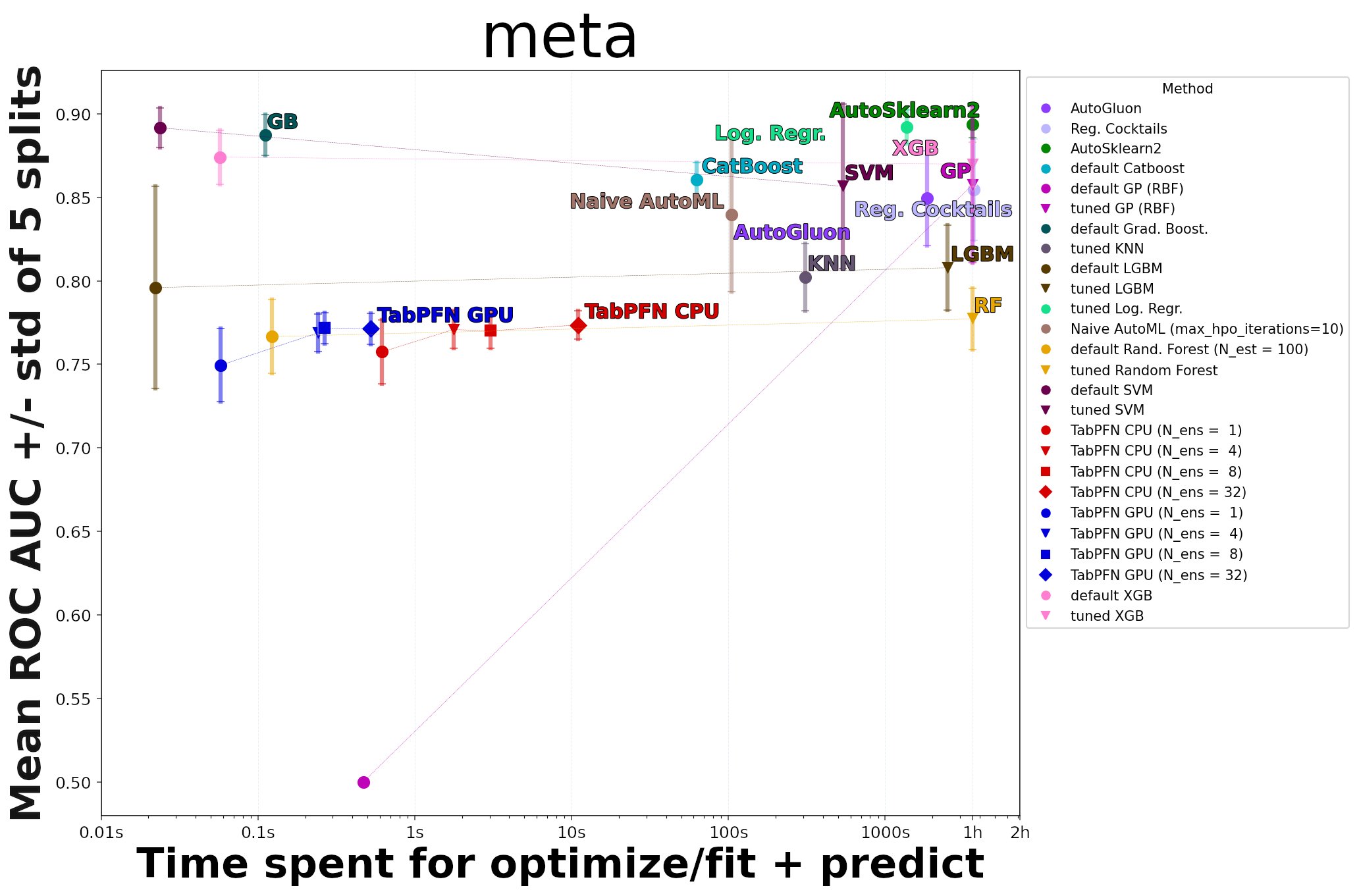}
         \caption{\texttt{meta}, many missing values\label{fig:meta}}
     \end{subfigure}
     \caption{Further poor results for TabPFN, on datasets with categorical features and missing values, which we did not focus on in creating TabPFN.}
     \label{fig:poor_results}
\end{figure}

TabPFN also works poorly on purely categorical datasets (e.g., dataset \texttt{sensory}, Figure \ref{fig:sensory}) or with many missing values (e.g., dataset \texttt{meta}, Figure \ref{fig:meta}).

\begin{figure}[h]
     \centering
     \begin{subfigure}[h]{0.49\textwidth}
         \includegraphics[width=1\columnwidth]{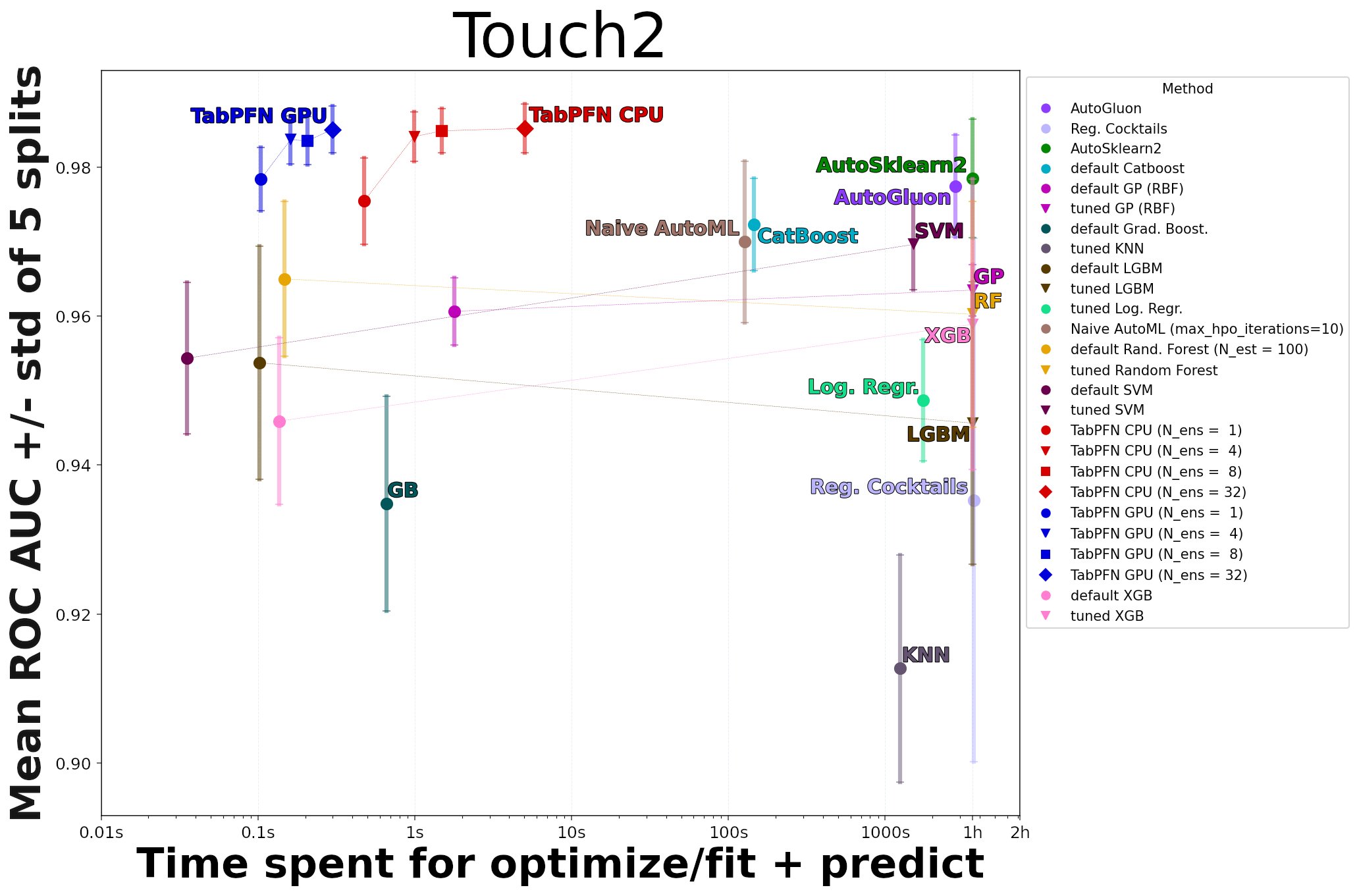}
         \caption{\texttt{Touch-2}, purely categorical \label{fig:touch}}
     \end{subfigure}
     \begin{subfigure}[h]{0.49\textwidth}
         \includegraphics[width=1\columnwidth]{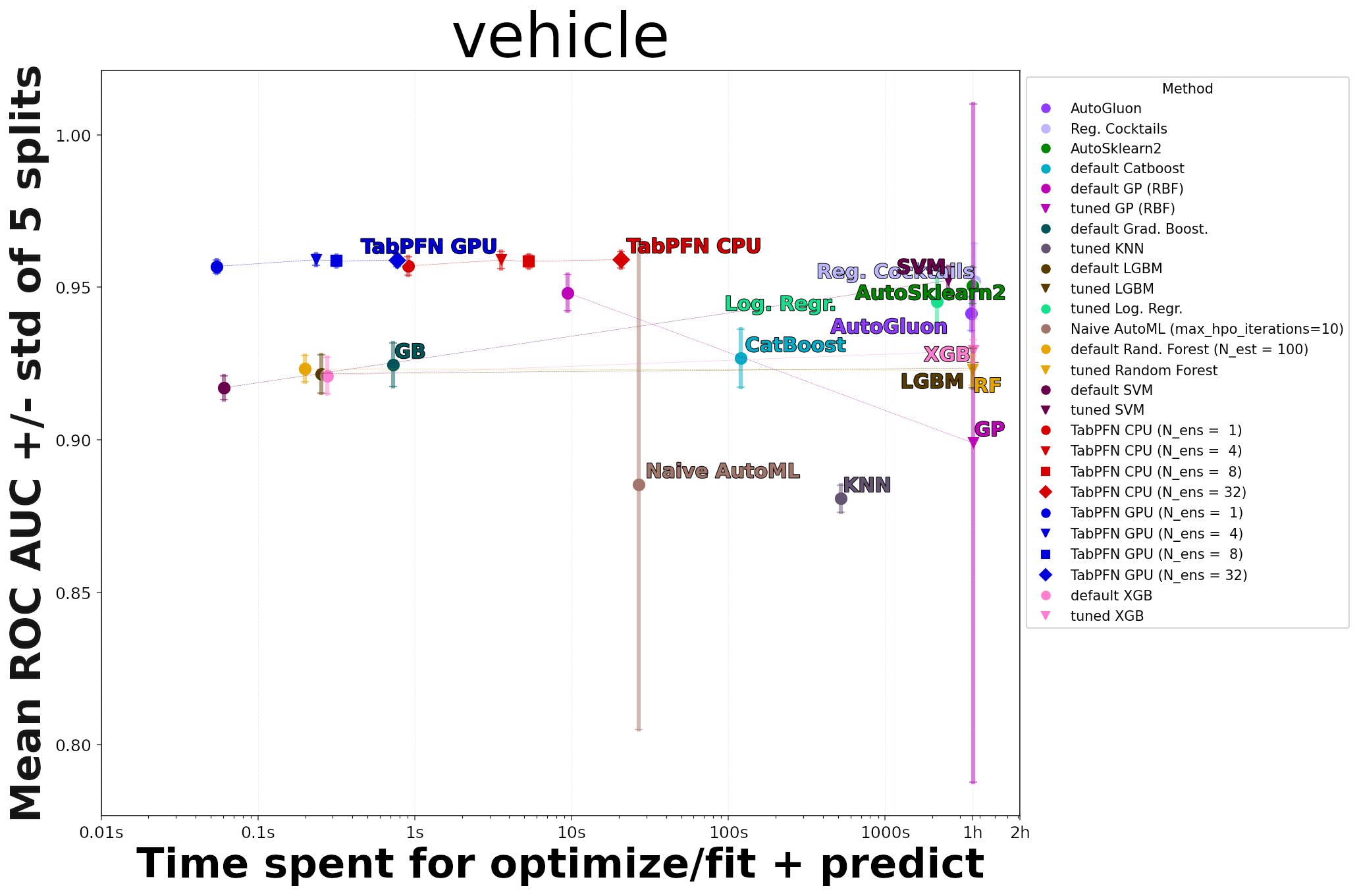}
         \caption{\texttt{vehicle}\label{fig:vehicle}}
     \end{subfigure}
     \caption{Strong results for TabPFN for individual datasets with numerical features and no missing values.}
     \label{fig:good_results}
\end{figure}

TabPFN works best for purely numerical datasets, with no missing values, e.g., datasets \texttt{Touch2} and \texttt{vehicle}, see Figure \ref{fig:good_results}.

\begin{figure}[h]
     \centering
     \begin{subfigure}[h]{0.49\textwidth}
         \includegraphics[width=1\columnwidth]{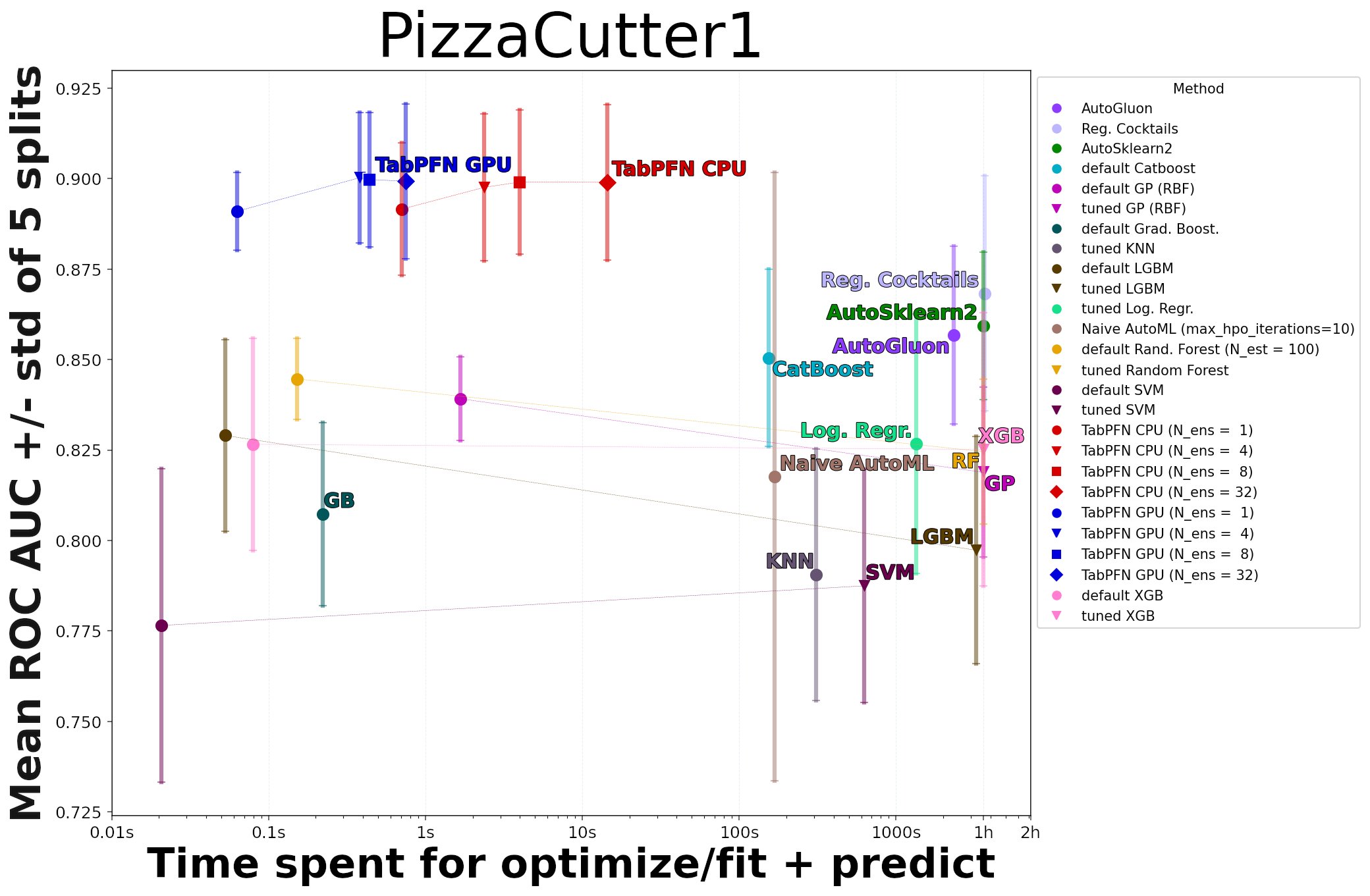}
         \caption{\texttt{PizzaCutter1} \label{fig:PizzaCutter}}
     \end{subfigure}
     \begin{subfigure}[h]{0.49\textwidth}
         \includegraphics[width=1\columnwidth]{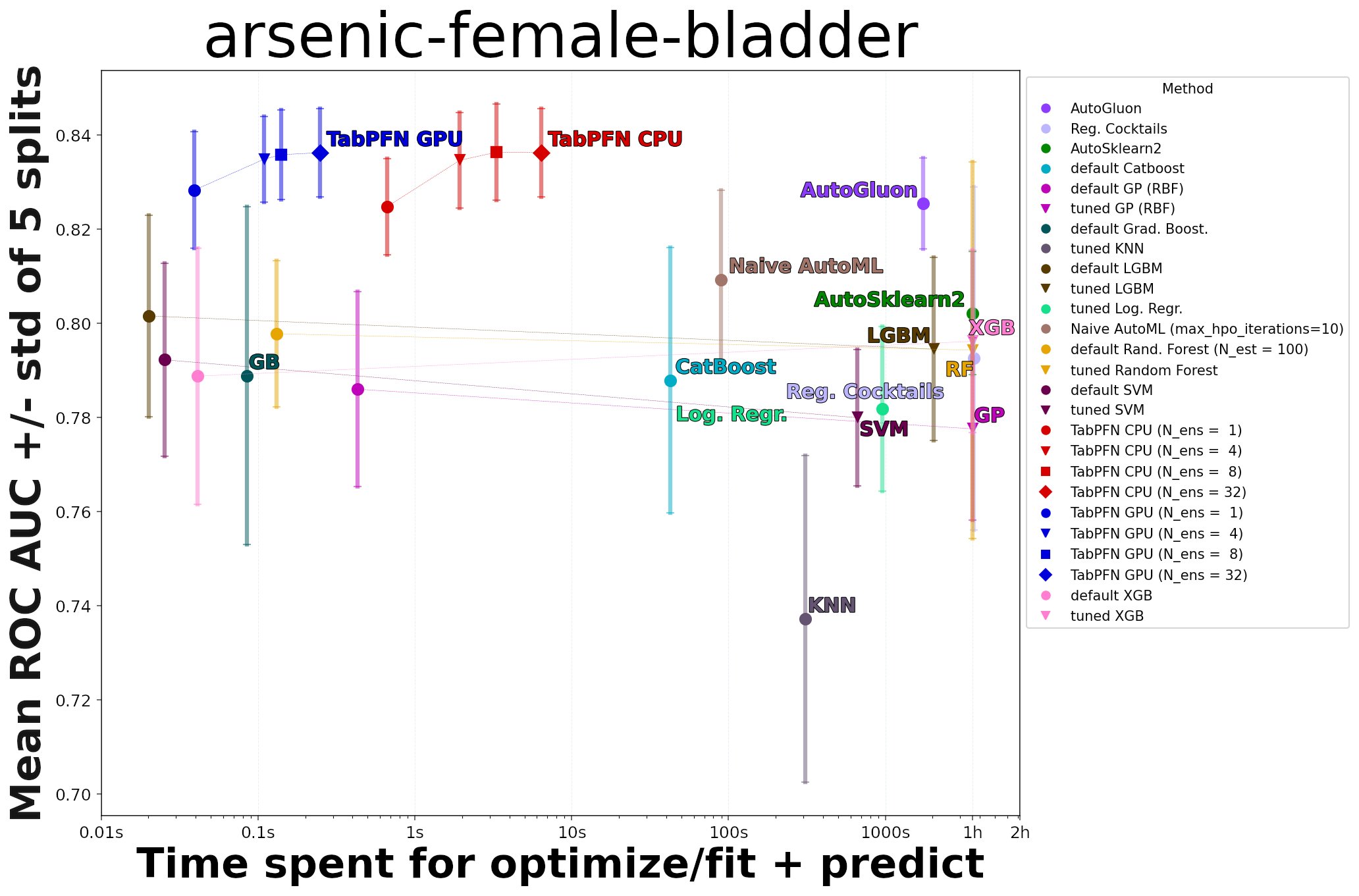}
         \caption{\texttt{arsenic-female-bladder}\label{fig:arsenic-female-bladder}}
     \end{subfigure}
     \caption{TabPFN can still yield strong results when hyperparameter optimization does not help (potentially due to overfitting).}
     \label{fig:good_results-tuning}
\end{figure}

TabPFN still works well in several datasets where hyperparameter optimization does not help in the baselines (maybe due to overfitting), e.g. \texttt{Pizza-cutter1} and \texttt{arsenic-female-bladder}, see Figure \ref{fig:good_results-tuning}. We attribute this to being Bayesian, having meta-learned not to overfit on small datasets with a simplicity prior using principles from causality. 

\begin{figure}[h]
     \centering
     \begin{subfigure}[h]{0.49\textwidth}
         \includegraphics[width=1\columnwidth]{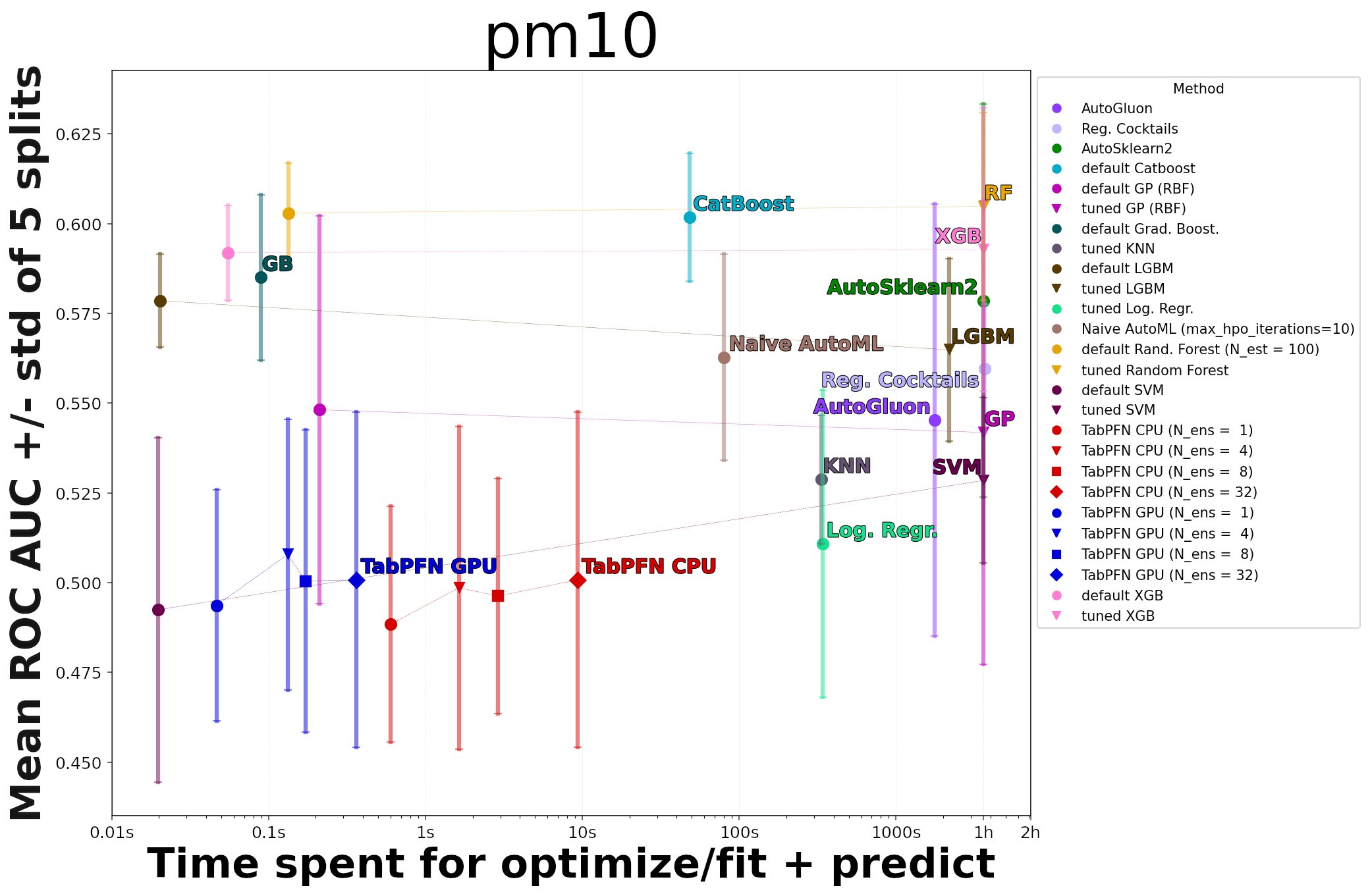}
         \caption{\texttt{pm10}, purely numerical, but still poor performance \label{fig:pm}}
     \end{subfigure}
     \begin{subfigure}[h]{0.49\textwidth}
         \includegraphics[width=1\columnwidth]{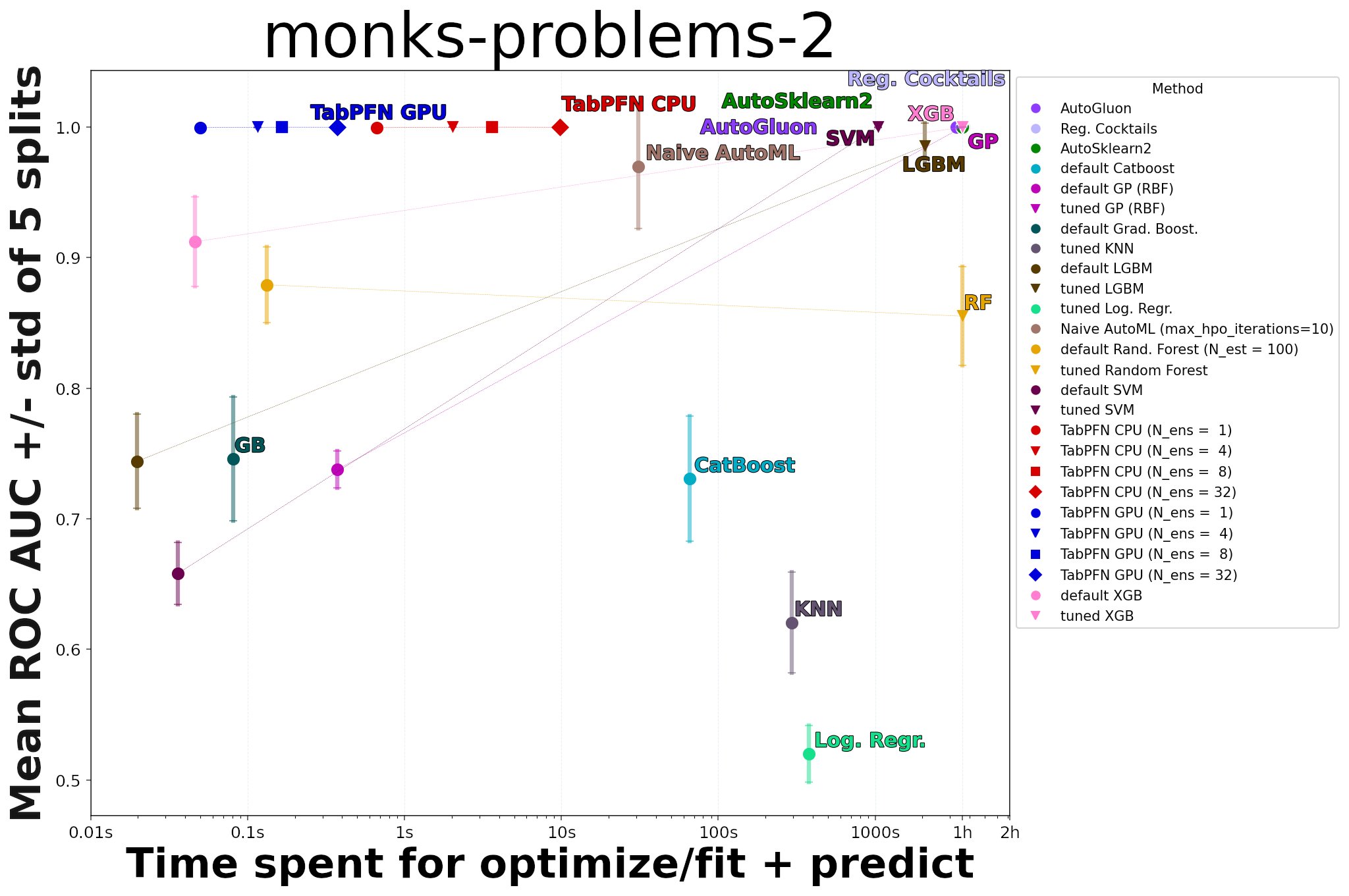}
         \caption{\texttt{monks-problem2}, categorical, but still strong performance\label{fig:monks-problem2}}
     \end{subfigure}
     \caption{TabPFN can also yield poor performance on numerical datasets (\texttt{pm}) and  strong performance on categorical datasets (\texttt{monks-problem2}.}
     \label{fig:strange_results}
\end{figure}

It is \emph{not} the case, however, that TabPFN works great on all numerical datasets without missing values and poorly on all categorical datasets. For example, TabPFN works poorly on the numerical dataset \texttt{pm10} and great on the categorical dataset \texttt{monks-problem2}, see Figure \ref{fig:strange_results}. It is therefore necessary to look beyond individual datasets to obtain the entire picture. 

%In this section, we describe additional methods included in our prior. %First, we describe mixing in Gaussian Process (GP) priors as another way to diversify the hypotheses approximated by \methodname{}. Next, we describe refinements to our prior for the usage on tabular data.

\section{Details of the \methodname{} Prior}
\label{app:a prior for tabular data}
%In this section, we describe additional methods included in our prior. %First, we describe mixing in Gaussian Process (GP) priors as another way to diversify the hypotheses approximated by \methodname{}. Next, we describe refinements to our prior for the usage on tabular data.
\subsection{SCM Prior}
\label{sec:details_scm_prior}
\paragraph{The Sampling Algorithm}
We instantiate a subfamily of DAGs that can be efficiently sampled from by starting with a MLP architecture and dropping weights from it.
That is, to sample a dataset with $k$ features and $n$ samples from our prior we perform the following steps for each dataset: \\
\hphantom{~~~~}(1) We sample the number of MLP layers $l \sim p(l)$ and nodes $h \sim p(h)$ and sample a graph $\mathcal{G}$($Z$, $E$) structured like an $l$-layered MLP with hidden size $h$.\\%, $p(l)$ and $p(h)$ are hyperparameters.\\
\hphantom{~~~~}(2) We sample weights for each Edge $E_{ij}$ % between $i$ and $j$
as $W_{i,j} \sim p_w(\cdot)$.\\
\hphantom{~~~~}(3) We drop a random set of edges $e \in E$ to yield a random DAG.\\
\hphantom{~~~~}(4) We sample a set of $k$ feature nodes $N_x$  and a label node $N_y$ from the nodes $Z$.\\
\hphantom{~~~~}(5) We sample the noise distributions $p(\epsilon) \sim p(p(\epsilon))$ from a meta-distribution. %, $p(p(\epsilon))$ is a hyperparameter.
This yields an SCM, with all $f_i$'s instantiated as random affine mappings followed by an activation. Each $z_i$ corresponds to a sparsely connected neuron in the MLP.

With the above parameters fixed, we perform the following steps for each member of the dataset:\\
\hphantom{~~~~}(1) We sample noise variables $\epsilon_i$ from their specific distributions.\\
\hphantom{~~~~}(2) We compute the value of all $z \in Z$ with $z_i = a((\sum_{j \in \text{PA}_{\mathcal{G}(i)}} E_{ij} z_j) + \epsilon_i)$.\\
\hphantom{~~~~}(3) We retrieve the values at the feature nodes $N_x$ and the output node $N_y$ and return them.

We sample one activation function $a$ per dataset from $\{Tanh, Leaky ReLU, ELU, Identity\}$ \citep{nair2010rectified}. The sampling scheme for the number of layers $p(l)$ and nodes $p(h)$ is designed to follow a discretized noisy log-normal distribution, $p(\epsilon)$ is a noisy log-normal distribution and the dropout rate follows a beta distribution.
The full information can be found in Table \ref{tab:diffable_hps}.

\subsection{Tabular Data Refinements}
\label{sec:tabular data refinement}
% How do these priors relate to BIC or other criteria that trade off size and quality?Where is this used later on? I can't find the simplicity in the prior.Why does this not cite a single paper about dataset generation? There is plenty of work on learning datasets distributions and sampling from them and that should be recognized. I'd be surprised if no one has yet done sampling from priors as you do, so this should be discussed.

Tabular datasets comprise a range of peculiarities, e.g.\ feature types can be numerical, ordinal, or
categorical and feature values can be missing, leading to sparse features. We seek to reflect these peculiarities in the design of our prior as described in the following sections. 

\subsubsection{Preprocessing} \label{app:preprocessing} During prior-fitting, input data is normalized to zero mean and unit variance, and we apply the same step when evaluating on real data. Since tabular data frequently contains exponentially scaled data, which might not be present during prior-fitting we apply power scaling during inference \citep{yeo2000new}. Thus, during inference on real tabular datasets the features more closely match those seen during prior-fitting. We use only training samples for calculating z-statistics, power transforms and all other preprocessing. We take this preprocessing time into account when reporting the inference time of our method.

\subsubsection{Correlated Features}
\begin{figure}[t]
     \centering
         \centering
            \includegraphics[width=0.4\columnwidth]{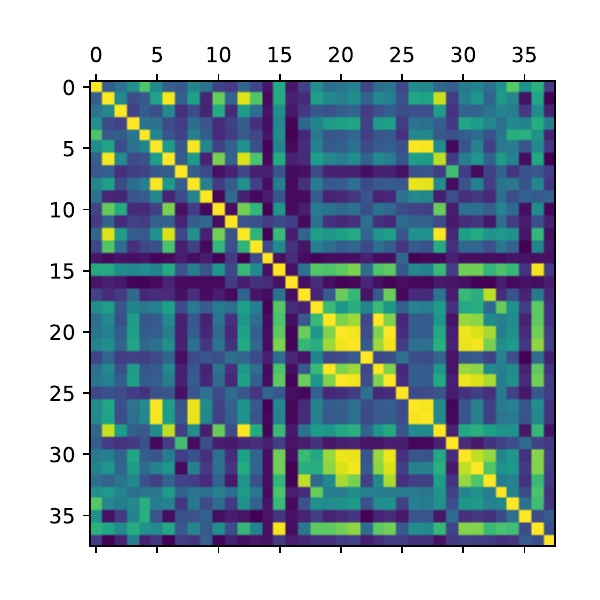}
%         \caption{}
         \centering
            \includegraphics[width=0.4\columnwidth]{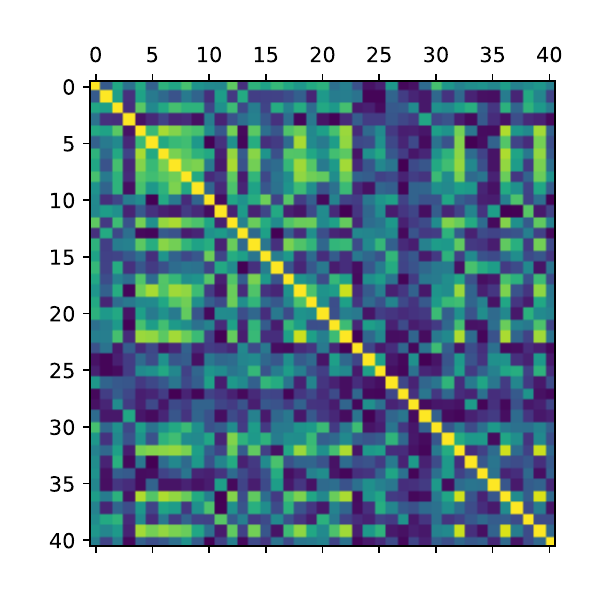}
     \vspace*{-0.2cm}
     \caption{Feature correlation matrices for a real-world (``PC4 Software defect prediction'', left) and a synthetic (right) dataset, where brighter colors indicate higher correlation. 
     %The real-world data comes from the ``PC4 Software defect prediction'' dataset, while the synthetic data is generated by our causal prior, detailed information on feature ordering in Appendix \ref{sec:tabular data refinement}. 
     %Both seem to have similarly structured correlations.
     }
     \label{fig:feature correlation example}
\end{figure}

Feature correlation in tabular data varies between datasets and ranges from independent to highly correlated. This poses problems to classical deep learning methods \citep{nnsfortabular}.
% In a structural causal model, nodes can be connected to a similar set of input nodes and their values will thus be a linear combination of these features before the application of noise and an activation function.
When considering a large space of SCMs, correlated features of varying degrees naturally arise in our priors.
Furthermore, in real-world tabular data, the ordering of features is often unstructured, however adjacent features are often more highly correlated than others. We use ``Blockwise feature sampling'' to reflect the correlation structure between ordered features. Our generation method of SCMs naturally provides a way to do this. The first step in generating our SCMs is generating a unidirectional layered network structure in which nodes in one layer can only receive inputs from the preceding layer. Thus, features in the same layer tend to be more highly correlated. We use this by sampling adjacent nodes in the layered network structure in blocks and using these ordered blocks in our set of features. In Figure~\ref{fig:feature correlation example}, we visualize the correlations of such a generated dataset (right) and compare them to a real-world dataset (left), demonstrating that our prior yields correlation structures similar to those of real datasets.
% Also inspired by \citep{simon1991architecture} we drop out connectivity in blocks, .. maybe in appendix but this is a detail I believe

\subsubsection{Generating irregular functions} \label{app:irregular_funcs}

In real-world data, some features are consistently more important than others. While a random network weight initialization leads to slightly different feature importances, the average effect of input features regresses to the mean when the hidden dimensionality increases. We amplify differences by sampling a weight parameter for each input feature and multiplying all outgoing weights by this factor.
In the prior, we randomly sparsify connections of the graph. Thus hidden variables and the output node are influenced by fewer parameters, yielding more irregular patterns, as a larger number of parameters once again regresses to the mean. We also extend sparsification to blocks of variables, leading to some groups of variables interacting more strongly.
We also extend the way noise variables are sampled. Instead of sampling Gaussian noise at each node from the same distribution, we first sample separate noise means and standard deviations for each node and then sample from this distribution.
Also, we generate non-uniformly distributed input data x, as observed in real-world data: We sample the input variables x (which are propagated through our network), from a mix of distributions, namely the Gaussian, Zipfian and Multivariate Distribution.

\subsubsection{Nan Handling}
We do not have special nan handling built into our model.
We replace nan values with zero at test time.

\subsubsection{Categorical Features}
Tabular data often includes not only numeric features but also discrete categorical ones. While categorical features should technically not be ordered, in practice, they sometimes are, i.e., the categories represent binned degrees of some underlying variable. We introduce categorical features by picking a random fraction $p_{cat}$ (a hyperparameter) of categorical features per dataset. Analogous to transforming numeric class labels to discrete multiclass labels, we convert dense features to discrete ones. Also analogous to multiclass labels, we pick a shuffling fraction of categorical features $p_{scat}$ where we reshuffle categories. For details, see Section \ref{sec:multi-class prediction}.
During prior-fitting, we use a probability for categorical features of $20\%$.

\subsubsection{Differences to Prior work on PFNs for Tabular Data}
\label{app:comparison_muller}
Prior work by \cite{muller-iclr22a} has demonstrated tabular data classification using PFNs, but was limited to 30 training samples, balanced binary-classification and 60 features. Here we summarize the most important changes, to this prior work:
\begin{enumerate}[i), itemsep=-0.1em]
    \item The PFNs for tabular data described by \cite{muller-iclr22a} can only handle balanced binary datasets. In Section \ref{sec:multi-class prediction}, we show how to extend the prior to handle imbalanced classes and multi-class classification problems.
    \item  We studied pre-processing techniques for the TabPFN, which was not done at all before. This includes outlier removal and power scaling during inference \citep{yeo2000new}. 
    During training, we also rotate feature indices (based on the assumptions that it should not be informative whether a feature is listed in the $i$-th or the $j$-th column, but that features may be listed in groups and that there may be more interactions between features with similar indices\footnote{To avoid misunderstandings and give an example, a rotation of column indices by 2 positions would change the columns of the $X$ matrix from $[x_1, x_2, x_3, x_4]$ to $[x_3, x_4, x_1, x_2]$. This is \emph{not} a rotation of the $X$ matrix, just of its column indices.}) and class labels (based on the same assumptions).
    \item We construct ensembles over multiple pre-processings. Specifically, ensemble members differ in the feature column rotation, the class label rotation, and the use of a power transform. We only include each combination of the $k$ possible feature column rotations (for $k$ features), $j$ possible class label rotations (for $j$-ary classification problems) and 2 choices of transformation (power transform, none). Even if a number of ensemble members is specified that is larger than $2kj$, we only use $2kj$ ensemble members.
    \item We changed the transformer architecture to make it faster. We shrank attention matrix sizes from $(n+m)^2$ to $n^2 + n*m$, for $n$ training points and m inference points.
    \item We introduce a novel SCM prior. We compare the SCM, the improved BNN prior (which is more heavily based on the BNN setup of \ref{app:comparison_muller}) and SCM + BNN in Table \ref{tab:scm bnn ablation}. This improvesperformance by 2\% in this smaller scale setup, which is a bigger difference than between the performance of the final TabPFN and all baselines besides KNN and SAINT.
\end{enumerate}

% \subsubsection{Fixing prior bias}
% Furthermore the current technical implementation of our sampling setup leads to detrimental biases in the expected class distribution and significance of features based on their position. Biases over the class distribution can be attributed to sampling lower class labels more often than high class labels (e.g. classes one and two will be present in all datasets). Even when a class with a low class label is never observed, observing a higher class value means that this lower class label will be present in the test set due to our algorithm.
% We seek to alleviate these biases by performing class and feature shifts during inference, i.e. we randomly rotate the label and feature ordering. When the same data is predicted across all class and feature rotations these biases even out. This however is costly and so we randomly sample shifts balancing accuracy and speed - the losses in speed and accuracy introduced by this are affecting our results. Fixes to this are possible and will push performance even further.
\section{Details of the Prior-Data Fitted Network Algorithm} 
\label{app:pfn}
Algorithm \ref{alg:prior-fitting} describes the training method proposed by \citet{muller-iclr22a} for PFNs.

\begin{algorithm}[H]
\SetAlgoLined
\SetKwInOut{Input}{Input}
\SetKwInOut{Output}{Output}
\Input{A prior distribution over datasets $p(D)$, from which samples can be drawn and the number of samples $K$ to draw}
\Output{A model $q_\theta$ that will approximate the PPD}
 Initialize the neural network $q_\theta$\;
 % $p(x_1,\dots,x_{n+m},y_1,\dots,y_{n+m})$\;
 \For{$j\gets1$ \KwTo $K$}{
  Sample $D \cup \{(x_i,y_i)\}_{i=1}^m\sim p(D)$\; %$x_1,\dots,x_{n+m},y_1,\dots,y_{n+m} \sim p(\dataset{})$\;
  Compute stochastic loss approximation $\bar{\ell}_\theta$ = $\sum_{i=1}^{m}(-\log %q_\theta(y_{n+i}|x_{n+i},x_1,\dots,x_{n},y_1,\dots,y_{n}))$\;
  q_\theta(y_i|x_i, D))$\;
  Update parameters $\theta$ with stochastic gradient descent on $\nabla_\theta \bar{\ell}_\theta$\;
 }
 \caption{Prior-fitting of a PFN \citep{muller-iclr22a}}
 \label{alg:prior-fitting}
\end{algorithm}
\section{Setup of our method}
\label{sec:our_setup}

\subsection{Transformer Hyperparameters}\label{app:transformer_hypers}
We considered only PFN Transformers with $12$ layers, embeddings size $512$, hidden size $1024$ in feed-forward layers, and $4$-head attention. We used the Adam optimizer \citep{kingma-iclr15a} with linear-warmup and cosine annealing \citep{loshchilov-iclr17a}. For each training we tested a set of 3 learning rates, $\{.001,.0003,.0001\}$, and used the one with the lowest final training loss. The resulting model contains  25.82 M parameters.

\subsection{PFN Architecture Adaptations}
\label{app:pfn architecture change}
\textbf{Attention Adaption}
The original PFN architecture \citep{muller-iclr22a} uses a single multi-head self-attention module \citep{vaswani-neurips17a} to compute the attention between all the training examples, as well as, the attention from validation examples to training examples.
We replaced this, with two modules that share weights, one which computes self-attention among the training examples and the other that only compute cross-attention from validation examples to training examples.
Conceptually, this is equivalent to the original architecture, except that we're using a slightly different self-attention mask than the original architecture, which allowed all examples to attend to itself (the diagonal is 1), as in this example:
\begin{align}
%M_{3,2} =
\begin{bmatrix}
1& 1& 1& 0& 0\\
1& 1& 1& 0& 0\\
1& 1& 1& 0& 0\\
1& 1& 1& 1& 0\\
1& 1& 1& 0& 1\\
\end{bmatrix}.
\end{align}
For validation examples, we remove the attention to themselves. In terms of the example above:
\begin{align}
%M_{3,2} =
\begin{bmatrix}
1& 1& 1& 0& 0\\
1& 1& 1& 0& 0\\
1& 1& 1& 0& 0\\
1& 1& 1& \textcolor{red}{0}& 0\\
1& 1& 1& 0& \textcolor{red}{0}\\
\end{bmatrix}.
\end{align}
Information about the state of the current position does still flow through the residual branch, though.

\textbf{Flexible Encoder}
Datasets have unequal numbers of input dimensions (features), while PFNs use an encoder layer that accepts fixed dimensional inputs. Here we explain how datasets with different numbers of dimensions can be modelled with a single PFN:
We draw the number of dimensions of a dataset during training uniformly at random up to $100$.
Our encoder changes to accomodate this training and inference with different numbers of features by zero-padding datasets where
the number of features $k$ is smaller than the maximum number of features $K$ and scaling these features by $\frac{K}{k}$, s.t.\ the magnitude stays the same.

\subsection{\methodname{} Training}\label{app:training_details}
We trained our final model for $18\,000$ steps with a batch size of $512$ datasets. That is our \methodname{} is trained on $9\,216\,000$ synthetically generated datasets. This training takes 20 hours on 8 GPUs (Nvidia RTX 2080 Ti). Each dataset had a fixed size of $1024$ and we split it into training and validation uniformly at random. We generally saw that learning curves tended to flatten after around $10$ million datasets and were generally very noisy. Likely, this is because our prior generates a wide variety of different datasets.

\subsection{Prior Hyperparameters}
\label{app:prior_hparams}
The hyperparameters of our prior were chosen based on simplicity and our observations on the validation datasets (such as their class distributions or feature correlation strengths). Also, during algorithm development, we evaluated our models on this set of datasets to decide if our developed methods were correct and working. Since our prior hyperparameters specify distributions and not definite values, they can be chosen over a wide range and resemble the intervals chosen for a random hyperparameter search. The prior distributions we used are given in Table \ref{tab:diffable_hps}. %\ref{tab:diffable_hps}. Hyperparameters that specify sampling distributions other than uniform are made up of one or more underlying uniform distributions. Choice HPs are made up of one hyperparameter per choice, except for the first choice. Beta distributions are made up of two parameters $\alpha$ and $\beta$. Truncated Normals are made up of a mean and a standard deviation. The $Min$ value is added to the HP to give a lower bound. If Rounding is enabled, Hps will be rounded to the next value after sampling. The model receives the values of each underlying parameter making up these distributions during \priortuning{}.

\begin{table}[htbp]
\small
    \centering
    \caption{Overview of our prior hyperparameter distribution. For many features we use a Log Uniform distribution with truncated normal noise, which we refer to as $\text{TNLU}(h|\check{\mu}, \hat{\mu}, min, round)$. We sample from it by first sampling mean $\mu$ and standard deviation $\sigma$ from $\mu, \sigma \sim \text{LogUniform}(\check{\mu},\hat{\mu})$ and then sampling from the resulting truncated normal distribution $v \sim \text{TruncNormal}(\mu, \sigma^2, a=0, b=inf)$.
    $v$ is rounded to the closest integer, if $round$ is set. The final sampled value then is $h = v + min$.}
    \begin{tabular}{llllll}
    \toprule
    {} & Sampling &\\% Minimum & Maximum & & \\
    {} & distribution $p(\psi)$ &  &  & & \\
    % \midrule
    % GP sampling likelihood &                         Uniform &     0.5 &     8.0 & & \\
    %\toprule
    %{} &  & Min $\alpha$ and $\beta$ & Max $\alpha$ and $\beta$ & %Scale HP & \\
    \midrule
    MLP weight dropout &\multicolumn{3}{l}{$0.9 \cdot \text{Beta}(a,b)$, where $a,b \sim \text{Uniform}(0.1,5.0)$}\\
    \toprule
    {} &  & Choices &  &  &  \\
    \midrule
    Sample SCM vs BNN &                     Uniform Choice &                                      \{True, False\} & & & \\
    Share Noise mean for nodes &                     Uniform Choice &                                      \{True, False\} & & & \\
    Input feature scaling enabled &                     Uniform Choice &                                      \{True, False\} & & & \\
    Sample y from last MLP layer         &                     Uniform Choice &                                      \{True, False\} & & &  \\
    MLP Activation Functions                   &                     Uniform Choice &  \multicolumn{4}{l}{\{Tanh, Leaky ReLU, ELU, Identity\}}  \\
    Blockwise Dropout                       &                     Uniform Choice &                                      \{True, False\} & & &  \\
    Keep SCM feature order                  &                     Uniform Choice &                                      \{True, False\} & & &  \\
    Sample feature nodes blockwise          &                     Uniform Choice &                                      \{True, False\} & & &  \\
    %GP noise                                &                     Uniform Choice &              \multicolumn{2}{l}{\{1e-05, 0.0001, 0.01\}}\\
    \toprule
    {} &  & Max Mean $\hat{\mu}$ & Min Mean $\check{\mu}$ & $round$ & $min$ \\
    \midrule
    MLP \#layers                 &      TNLU &        6 &        1 &        True &           2 \\
    MLP \#hidden nodes per layer &      TNLU &      130 &        5 &        True &           4 \\
    Gaussian Noise Std.          &      TNLU &      0.3 &   0.0001 &       False &         0.0 \\
    MLP Weights Std.             &      TNLU &     10.0 &     0.01 &       False &         0.0 \\
    SCM \#nodes at layer 1        &     TNLU &       12 &        1 &        True &           1 \\
    %GP outputscale               &      Trunc. Normal &     10.0 &  0.00001 &       False &           0 \\
    %GP lengthscale               &      Trunc. Normal &     10.0 &  0.00001 &       False &           0 \\
    \bottomrule
    \end{tabular}
    \label{tab:diffable_hps}
\end{table}

%\section{Reproducibility}
%Code, ..
% ---------------------------------------------
\section{Details for Tabular Experiments}\label{app:expdetails}
% ---------------------------------------------

Here we provide additional details for the experiments conducted in Section~\ref{sec:experiments} in the main paper. 

% ---------------------------------------------
\subsection{Hardware Setup} \label{app:ssec:hardware}
% ---------------------------------------------
All evaluations, including the baselines, ran on a compute cluster equipped with Intel(R) Xeon(R) Gold 6242 CPU @ 2.80GHz using $1$ CPU with up to $6$GB RAM. For evaluation using our \methodname{}, we additionally use an RTX 2080 Ti.

% ---------------------------------------------
\subsection{Baselines}\label{app:ssec:spaces}
% ---------------------------------------------
We provide the search space used to tune our baselines in Table~\ref{app:tab:spaces}. For \textit{CatBoost} and \textit{XGBoost}, we used the same ranges as~\citet{nnsfortabular2} with the following exception: For \textit{CatBoost} we removed the hyperparameter max\_size since we could not find it in the official documentation.
To be maximally fair to XGBoost, we also tried the search space of quadruple Kaggle grandmaster Bojan Tunguz \citep{xgboost_search_space_tunguz}, which we adapted slightly by using softmax instead of logistic, as we are in the multi-class setting. XGBoost with this search space performed worse for all considered time budgets than the search space by \citet{nnsfortabular2}.
The search spaces for the KNN, GP and Logistic Regression baselines were designed from scratch and we used the respective implementation from \textit{scikit-learn}~\citep{scikit-learn}. For \textit{CatBoost} and \textit{AutoSklearn}, we pass the position of categorical features to the classifier (\textit{AutoGluon} automatically detects categorical feature columns). We normalize inputs for Logistic Regression, GP and KNN to the range [0, 1] using MinMax Scaling.
% Hyperparameters taken from https://arxiv.org/pdf/2106.03253.pdf. Catboost max_depth couldn't be found.

\begin{figure}[h]
     \centering
     \hspace*{-1.1cm}
     \begin{subfigure}[h]{1.0\textwidth}
         \includegraphics[width=1\columnwidth]{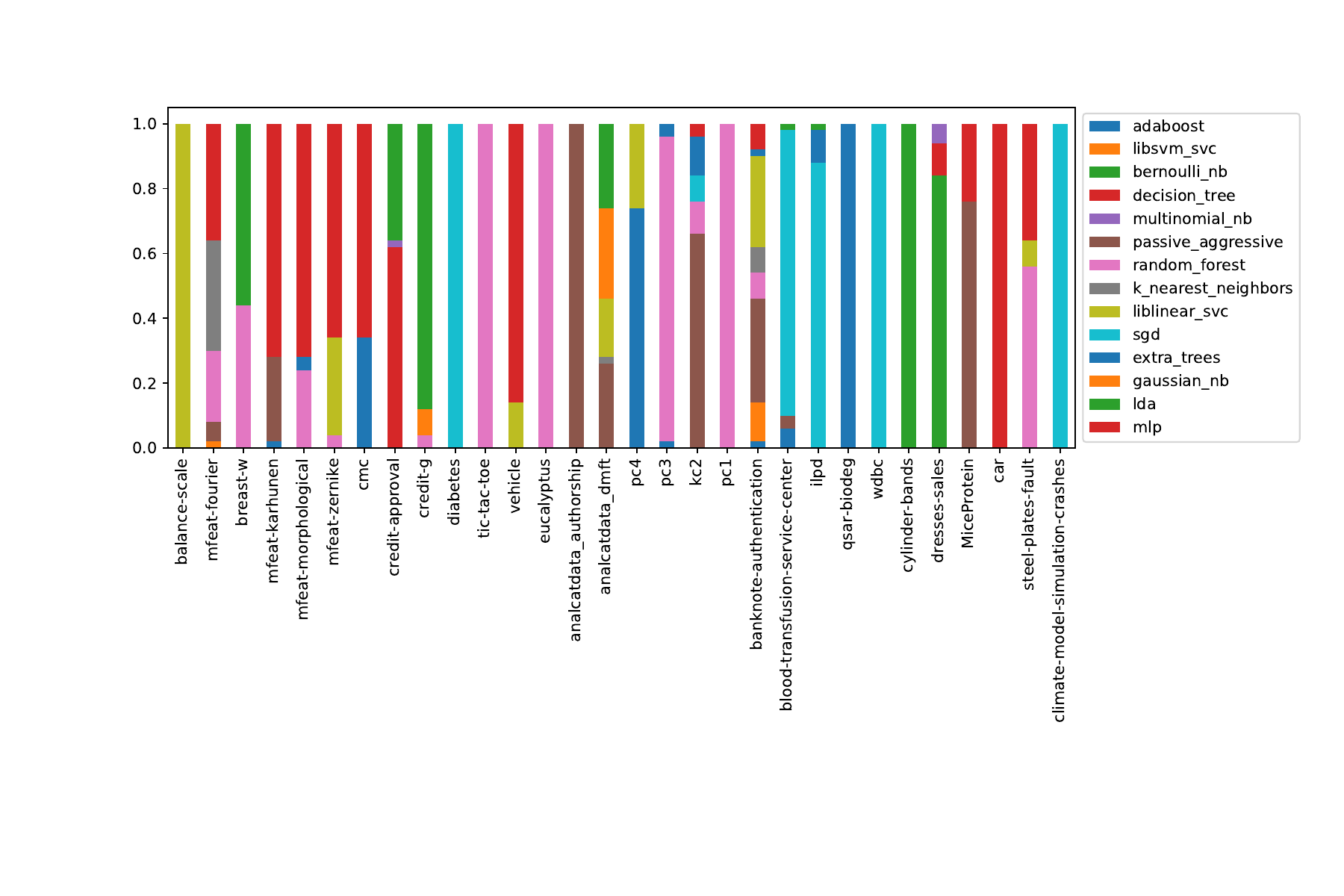}
     \end{subfigure}
     \vspace*{-1.5cm}
     \caption{Ensemble weights of classifiers used in AutoSklearn baseline for each dataset. Ensemble weights are averaged across 5 splits for each dataset in the \openmlcc{} Benchmark after one hour of training.}
    \label{fig:autosklearn_models}
\end{figure}

\begin{table}[htbp]
\caption{Hyperparameter spaces for baselines. All, except LightGBM, adapted from~\citet{nnsfortabular2}.}
\label{app:tab:spaces}
\centering
\begin{tabular}{lllll}
\toprule
baseline &  name & type & log & range \\
\midrule
\multirow{3}{*}{LogReg} & penalty & cat & (l1, l2, none) & - \\
    & max\_iter & int & [50, 500] & - \\
    & fit\_intercept & cat & (True, False) & - \\
    & C & float & [$e^{-5}$, 5] & - \\
\midrule
\multirow{1}{*}{KNN} & n\_neighbors & int & [1, 16] & - \\
\midrule
\multirow{2}{*}{GP} & params\_y\_scale & float & [0.05, 5.0] & yes \\
    & params\_length\_scale & float & [0.1, 1.0] & yes \\
\midrule
\multirow{6}{*}{CatBoost} & learning\_rate & float & [$e^{-5}$, 1] & yes \\
    & random\_strength & int & [1, 20] & - \\
    & l2\_leaf\_reg & float & [1, 10] & yes \\
    & bagging\_temperature & float & [0, 1.0] & yes \\
    & leaf\_estimation\_iterations & int & [1, 20] & - \\
    & iterations & int & [100, 4000] & - \\
\midrule
\multirow{10}{*}{XGBoost} & learning\_rate & float & [$e^{-7}$, 1] & yes \\
    & max\_depth & int & [1, 10] & - \\
    & subsample & float & [0.2, 1] & - \\
    & colsample\_bytree & float & [0.2, 1] & - \\
    & colsample\_bylevel & float & [0.2, 1] & - \\
    & min\_child\_weight & float & [$e^{-16}$, $e^5$] & yes \\
    & alpha & float & [$e^{-16}$, $e^2$] & yes \\
    & lambda & float & [$e^{-16}$, $e^2$] & yes \\
    & gamma & float & [$e^{-16}$, $e^2$] & yes \\
    & n\_estimators & int & [100, 4000] & - \\
\midrule
\multirow{10}{*}{LightGBM} & num\_leaves & int & [5, 50] & yes \\
    & max\_depth & int & [3, 20] & yes \\
    & learning\_rate & float & [$e^{-3}$, 1] & - \\
    & n\_estimators & int & 50, 2000 & - \\
    & min\_child\_weight & float & [$e^{-5}$, $e^4$] & yes \\
    & reg\_alpha & float & [0, 1e-1, 1, 2, 5, 7, 10, 50, 100] & yes \\
    & reg\_lambda & float & [0, 1e-1, 1, 5, 10, 20, 50, 100] & yes \\
    & subsample & float & [0.2, 0.8] & - \\
\bottomrule
\end{tabular}
\end{table}
\iffalse
\pgfplotsset{
    mystylepalette3/.style={
        height=6cm,
        width=\textwidth,
    },
}
\begin{figure}[h]
     \centering
     \begin{subfigure}[t]{0.32\textwidth}
         \centering
         %\hspace*{-1.9cm}
         \input{figures/tabular_results/roc_over_time_autogluon_roc}
         \label{fig:aucovertime_autogluon}
     \end{subfigure}
     \begin{subfigure}[t]{0.32\textwidth}
         \centering
         %\hspace*{-0.2cm}
         \input{figures/tabular_results/roc_wins_tabular_autogluon_roc}
         \label{fig:winsovertime_autogluon}
     \end{subfigure}
     \begin{subfigure}[t]{0.32\textwidth}
         \centering
         %\hspace*{1.1cm}
         \input{figures/tabular_results/roc_raks_tabular_autogluon_roc}
         \label{fig:ranksovertime_autogluon}
     \end{subfigure}
     \vspace{-.6cm}
     \begin{subfigure}[b]{1.0\textwidth}
         \centering
         \input{figures/tabular_results/legend_tabular}
         \label{fig:legendovertime_autogluon}
     \end{subfigure}
     \vspace{.3cm}
     \caption{ROC AUC performance over time, as reported in Figure \ref{fig:perfovertime}. But with Autogluon fails.
     }
     \label{fig:perfovertime_autogluon}
\end{figure}
\fi
% ---------------------------------------------
\subsection{Used Datasets}\label{app:ssec:datasets}
% ---------------------------------------------

To construct and evaluate our method, we used the following four sets of datasets. 
%Our meta-test set (see Table~\ref{table:test_datasets_table}) is a subset of the \openmlcc{} benchmark suite~\citep{bischl-neuripsdbt21a}, whereas our meta-validation set (see Table~\ref{table:reference_datasets_table} and~\ref{table:reference_datasets_table_2}), comprises 150 datasets from \url{OpenML.org}~\citep{vanschoren-sigkdd14a}. These are licensed under the BSD 3-Clause license.

First, our meta-test set (see Table~\ref{table:test_datasets_table}) comprises all datasets in the \openmlcc{} benchmark suite~\citep{bischl-neuripsdbt21a}(available at \url{OpenML.org}) with at most $2\,000$ samples, $100$ features and $10$ classes, which leaves us with $30$ datasets that represent small, tabular datasets.

Second, our meta-validation set (see Tables~\ref{table:reference_datasets_table} and~\ref{table:reference_datasets_table_2}) comprises 150 datasets from \url{OpenML.org}~\citep{vanschoren-sigkdd14a}. For this, we considered all datasets on \url{OpenML.org} and applied the following filtering procedure: We dropped all datasets that are in the meta-test set and all datasets with more than $2\,000$ samples, $100$ features or $10$ classes. We also manually checked for overlaps and removed datasets where the number of features, classes and samples was identical to a dataset in the meta-test set. Furthermore, we manually dropped FOREX (since it is a time series dataset) and artificially-created datasets, such as the Univ and Friedman datasets. The remaining meta-validation set then contains $150$ datasets. This meta-validation set was used to guide the development of our prior hyperparameters as described in Appendix \ref{app:prior_hparams}.

Third, a subset of 5 datasets from the \openmlautoml{} Benchmark, comprises all datasets from the \openmlautoml{} Benchmark with at most $1\,111$ samples, $100$ features and $10$ classes. This is given in Table \ref{table:automl_small_datasets}. Due to the 10-fold cross-validation in the \openmlautoml{} Benchmark this is identical to the setup for our meta-test and meta-validation datasets above, where we used up to 2\,000 samples split 50-50 into training and test.

Fourth, our meta-generalization set, described in Appendix \ref{app:ssec:abl_generalization}, comprises 18 larger datasets from the OpenML AutoML Benchmark.

%filtering out datasets with more than 2,000 samples, 100 features or 10 classes. Of the 72 datasets included in the OpenML CC-18 Benchmark, 30 datasets remain that we use for benchmarking. The list of datasets used can be found in Appendix \ref{}. For each dataset we generate 5 splits of which we use 50\% as training data and 50\% for testing.
\subsection{Model Generalization}\label{app:ssec:abl_generalization}
For testing the \methodname{} performance on longer sequences, as described in Section \ref{fig:model generalization}, we used a set of 18 datasets from the OpenML AutoML Benchmark that contain at least $10\,000$ samples. The list of datasets used can be found in Table \ref{table:openml_datasets_table}. % TODO: After review cite Automl openml benchmark
For this evaluation, datasets with more than 100 features are limited to the first 100 features. When more than 10 classes are contained in the datasets, samples with any but the first 10 classes are discarded. 

\subsection{Details on Time Comparisons}
\label{app:time_comparison_discussion}
Time comparisons refer to combined fitting, tuning and prediction; see Table \ref{app:table:tabular_results_table} for the times split into tuning/fitting and prediction.
The time taken for each baseline and \methodname{} does not include the one-time cost of development of each method.
Thus, for AutoML baselines meta-learning cost was not included (e.g. Auto-Sklearn pipelines meta-learning, which involved running hyper-parameter search for 24 hours on 140 datasets (= 3360 CPU hours) \citep{feurer-arxiv20v2}). For GBDT methods the manual time that went into defining suitable hyperparameter spaces and the manual crafting of algorithms that perform well on tabular datasets is hard to measure and was not included.
For \methodname{} the prior-fitting phase (which is part of our algorithm development: i.e. developing ideas, writing code, trying ideas) is not included.
This is fair to all methods, as these costs are not on the user side and are amortized over time.

\begin{table}[bt]
    \caption{Datasets used for the evaluation. These include all 30 datasets in the \openmlcc{} benchmark suite with at most $2\,000$ samples, $100$ features and $10$ classes.}
    \label{table:test_datasets_table}
    \centering
    \tiny
    \begin{tabular}{@{\hskip 0mm}l@{\hskip 0mm}rrrrrrr}
\toprule
 Name & \#Feat. &  \#Cat. &  \#Inst. &  \#Class. &  \#NaNs &  Minor. Class Size & OpenML ID \\
\midrule
 balance-scale & 5 & 1 & 625 & 3 &   0 & 49 &    11 \\
 mfeat-fourier & 77 & 1 & 2000 & 10 &   0 & 200 &    14 \\
 breast-w & 10 & 1 & 699 & 2 &  16 & 241 &    15 \\
 mfeat-karhunen & 65 & 1 & 2000 & 10 &   0 & 200 &    16 \\
 mfeat-morphological & 7 & 1 & 2000 & 10 &   0 & 200 &    18 \\
 mfeat-zernike & 48 & 1 & 2000 & 10 &   0 & 200 &    22 \\
 cmc & 10 & 8 & 1473 & 3 &   0 & 333 &    23 \\
 credit-approval & 16 &    10 & 690 & 2 &  67 & 307 &    29 \\
 credit-g & 21 &    14 & 1000 & 2 &   0 & 300 &    31 \\
 diabetes & 9 & 1 & 768 & 2 &   0 & 268 &    37 \\
 tic-tac-toe & 10 &    10 & 958 & 2 &   0 & 332 &    50 \\
 vehicle & 19 & 1 & 846 & 4 &   0 & 199 &    54 \\
 eucalyptus & 20 & 6 & 736 & 5 & 448 & 105 &   188 \\
 analcatdata\_auth... & 71 & 1 & 841 & 4 &   0 & 55 &   458 \\
 analcatdata\_dmft & 5 & 5 & 797 & 6 &   0 & 123 &   469 \\
 pc4 & 38 & 1 & 1458 & 2 &   0 & 178 &  1049 \\
 pc3 & 38 & 1 & 1563 & 2 &   0 & 160 &  1050 \\
 kc2 & 22 & 1 & 522 & 2 &   0 & 107 &  1063 \\
 pc1 & 22 & 1 & 1109 & 2 &   0 & 77 &  1068 \\
 banknote-authenti... & 5 & 1 & 1372 & 2 &   0 & 610 &  1462 \\
blood-transfusion-... & 5 & 1 & 748 & 2 &   0 & 178 &  1464 \\
ilpd & 11 & 2 & 583 & 2 &   0 & 167 &  1480 \\
qsar-biodeg & 42 & 1 & 1055 & 2 &   0 & 356 &  1494 \\
wdbc & 31 & 1 & 569 & 2 &   0 & 212 &  1510 \\
cylinder-bands & 40 &    22 & 540 & 2 & 999 & 228 &  6332 \\
dresses-sales & 13 &    12 & 500 & 2 & 835 & 210 & 23381 \\
MiceProtein & 82 & 5 & 1080 & 8 &    1396 & 105 & 40966 \\
car & 7 & 7 & 1728 & 4 &   0 & 65 & 40975 \\
steel-plates-fault & 28 & 1 & 1941 & 7 &   0 & 55 & 40982 \\
climate-model-simu... & 21 & 1 & 540 & 2 &   0 & 46 & 40994 \\
\bottomrule
\end{tabular}

\end{table}

\begin{table}[bt]
    \centering
    \caption{Meta-Datasets used for developing the prior.}
    \tiny
    \begin{tabular}{@{\hskip 0mm}l@{\hskip 0mm}rrrrrrr}
\toprule
 Name & \#Feat. &  \#Cat. &  \#Inst. &  \#Class. &  \#NaNs &  Minor. Class Size & OpenML ID \\
\midrule
 breast-cancer & 10 & 10 & 286 & 2 & 9 & 85 & 13 \\
 colic & 27 & 20 & 368 & 2 & 1927 & 136 & 25 \\
 dermatology & 35 & 34 & 366 & 6 & 8 & 20 & 35 \\
 sonar & 61 & 1 & 208 & 2 & 0 & 97 & 40 \\
 glass & 10 & 1 & 214 & 6 & 0 & 9 & 41 \\
 haberman & 4 & 2 & 306 & 2 & 0 & 81 & 43 \\
 tae & 6 & 3 & 151 & 3 & 0 & 49 & 48 \\
 heart-c & 14 & 8 & 303 & 2 & 7 & 138 & 49 \\
 heart-h & 14 & 8 & 294 & 2 & 782 & 106 & 51 \\
 heart-statlog & 14 & 1 & 270 & 2 & 0 & 120 & 53 \\
 hepatitis & 20 & 14 & 155 & 2 & 167 & 32 & 55 \\
 vote & 17 & 17 & 435 & 2 & 392 & 168 & 56 \\
 ionosphere & 35 & 1 & 351 & 2 & 0 & 126 & 59 \\
 iris & 5 & 1 & 150 & 3 & 0 & 50 & 61 \\
 wine & 14 & 1 & 178 & 3 & 0 & 48 & 187 \\
 flags & 29 & 27 & 194 & 8 & 0 & 4 & 285 \\
 hayes-roth & 5 & 1 & 160 & 3 & 0 & 31 & 329 \\
 monks-problems-1 & 7 & 7 & 556 & 2 & 0 & 278 & 333 \\
 monks-problems-2 & 7 & 7 & 601 & 2 & 0 & 206 & 334 \\
 monks-problems-3 & 7 & 7 & 554 & 2 & 0 & 266 & 335 \\
 SPECT & 23 & 23 & 267 & 2 & 0 & 55 & 336 \\
 SPECTF & 45 & 1 & 349 & 2 & 0 & 95 & 337 \\
 grub-damage & 9 & 7 & 155 & 4 & 0 & 19 & 338 \\
 synthetic\_control & 61 & 1 & 600 & 6 & 0 & 100 & 377 \\
 prnn\_crabs & 8 & 2 & 200 & 2 & 0 & 100 & 446 \\
 analcatdata\_lawsuit & 5 & 2 & 264 & 2 & 0 & 19 & 450 \\
 irish & 6 & 4 & 500 & 2 & 32 & 222 & 451 \\
 analcatdata\_broadwaymult & 8 & 5 & 285 & 7 & 27 & 21 & 452 \\
 analcatdata\_reviewer & 8 & 8 & 379 & 4 & 1418 & 54 & 460 \\
 backache & 32 & 27 & 180 & 2 & 0 & 25 & 463 \\
 prnn\_synth & 3 & 1 & 250 & 2 & 0 & 125 & 464 \\
 schizo & 15 & 3 & 340 & 2 & 834 & 163 & 466 \\
 profb & 10 & 5 & 672 & 2 & 1200 & 224 & 470 \\
 analcatdata\_germangss & 6 & 5 & 400 & 4 & 0 & 100 & 475 \\
 biomed & 9 & 2 & 209 & 2 & 15 & 75 & 481 \\
 rmftsa\_sleepdata & 3 & 1 & 1024 & 4 & 0 & 94 & 679 \\
 diggle\_table\_a2 & 9 & 1 & 310 & 9 & 0 & 18 & 694 \\
 rmftsa\_ladata & 11 & 1 & 508 & 2 & 0 & 222 & 717 \\
 pwLinear & 11 & 1 & 200 & 2 & 0 & 97 & 721 \\
 analcatdata\_vineyard & 4 & 2 & 468 & 2 & 0 & 208 & 724 \\
 machine\_cpu & 7 & 1 & 209 & 2 & 0 & 56 & 733 \\
 pharynx & 11 & 10 & 195 & 2 & 2 & 74 & 738 \\
 auto\_price & 16 & 2 & 159 & 2 & 0 & 54 & 745 \\
 servo & 5 & 5 & 167 & 2 & 0 & 38 & 747 \\
 analcatdata\_wildcat & 6 & 3 & 163 & 2 & 0 & 47 & 748 \\
 pm10 & 8 & 1 & 500 & 2 & 0 & 246 & 750 \\
 wisconsin & 33 & 1 & 194 & 2 & 0 & 90 & 753 \\
 autoPrice & 16 & 1 & 159 & 2 & 0 & 54 & 756 \\
 meta & 22 & 3 & 528 & 2 & 504 & 54 & 757 \\
 analcatdata\_apnea3 & 4 & 3 & 450 & 2 & 0 & 55 & 764 \\
 analcatdata\_apnea2 & 4 & 3 & 475 & 2 & 0 & 64 & 765 \\
 analcatdata\_apnea1 & 4 & 3 & 475 & 2 & 0 & 61 & 767 \\
 disclosure\_x\_bias & 4 & 1 & 662 & 2 & 0 & 317 & 774 \\
 bodyfat & 15 & 1 & 252 & 2 & 0 & 124 & 778 \\
 cleveland & 14 & 8 & 303 & 2 & 6 & 139 & 786 \\
 triazines & 61 & 1 & 186 & 2 & 0 & 77 & 788 \\
 disclosure\_x\_tampered & 4 & 1 & 662 & 2 & 0 & 327 & 795 \\
 cpu & 8 & 2 & 209 & 2 & 0 & 53 & 796 \\
 cholesterol & 14 & 8 & 303 & 2 & 6 & 137 & 798 \\
 chscase\_funds & 3 & 1 & 185 & 2 & 0 & 87 & 801 \\
 pbcseq & 19 & 7 & 1945 & 2 & 1133 & 972 & 802 \\
 pbc & 19 & 9 & 418 & 2 & 1239 & 188 & 810 \\
 rmftsa\_ctoarrivals & 3 & 2 & 264 & 2 & 0 & 101 & 811 \\
 chscase\_vine2 & 3 & 1 & 468 & 2 & 0 & 212 & 814 \\
 chatfield\_4 & 13 & 1 & 235 & 2 & 0 & 93 & 820 \\
 boston\_corrected & 21 & 4 & 506 & 2 & 0 & 223 & 825 \\
 sensory & 12 & 12 & 576 & 2 & 0 & 239 & 826 \\
 disclosure\_x\_noise & 4 & 1 & 662 & 2 & 0 & 329 & 827 \\
 autoMpg & 8 & 4 & 398 & 2 & 6 & 189 & 831 \\
 kdd\_el\_nino-small & 9 & 3 & 782 & 2 & 466 & 274 & 839 \\
 autoHorse & 26 & 9 & 205 & 2 & 57 & 83 & 840 \\
 stock & 10 & 1 & 950 & 2 & 0 & 462 & 841 \\
 breastTumor & 10 & 9 & 286 & 2 & 9 & 120 & 844 \\
 analcatdata\_gsssexsurvey & 10 & 6 & 159 & 2 & 6 & 35 & 852 \\
 boston & 14 & 2 & 506 & 2 & 0 & 209 & 853 \\
 fishcatch & 8 & 3 & 158 & 2 & 87 & 63 & 854 \\
 vinnie & 3 & 1 & 380 & 2 & 0 & 185 & 860 \\
 mu284 & 11 & 1 & 284 & 2 & 0 & 142 & 880 \\
 no2 & 8 & 1 & 500 & 2 & 0 & 249 & 886 \\
 chscase\_geyser1 & 3 & 1 & 222 & 2 & 0 & 88 & 895 \\
 chscase\_census6 & 7 & 1 & 400 & 2 & 0 & 165 & 900 \\
 chscase\_census5 & 8 & 1 & 400 & 2 & 0 & 193 & 906 \\
 chscase\_census4 & 8 & 1 & 400 & 2 & 0 & 194 & 907 \\
 chscase\_census3 & 8 & 1 & 400 & 2 & 0 & 192 & 908 \\
 chscase\_census2 & 8 & 1 & 400 & 2 & 0 & 197 & 909 \\
 plasma\_retinol & 14 & 4 & 315 & 2 & 0 & 133 & 915 \\
 visualizing\_galaxy & 5 & 1 & 323 & 2 & 0 & 148 & 925 \\
 colleges\_usnews & 34 & 2 & 1302 & 2 & 7830 & 614 & 930 \\
\bottomrule
\end{tabular}
    \label{table:reference_datasets_table}
\end{table}

\begin{table}[bt]
    \centering
    \caption{Meta-Datasets used for developing the prior (continued).}
    \tiny
    \begin{tabular}{@{\hskip 0mm}l@{\hskip 0mm}rrrrrrr}
\toprule
 Name & \#Feat. &  \#Cat. &  \#Inst. &  \#Class. &  \#NaNs &  Minor. Class Size & OpenML ID \\
\midrule
 disclosure\_z & 4 & 1 & 662 & 2 & 0 & 314 & 931 \\
 socmob & 6 & 5 & 1156 & 2 & 0 & 256 & 934 \\
 chscase\_whale & 9 & 1 & 228 & 2 & 20 & 111 & 939 \\
 water-treatment & 37 & 16 & 527 & 2 & 542 & 80 & 940 \\
 lowbwt & 10 & 8 & 189 & 2 & 0 & 90 & 941 \\
 arsenic-female-bladder & 5 & 2 & 559 & 2 & 0 & 80 & 949 \\
 analcatdata\_halloffame & 17 & 2 & 1340 & 2 & 20 & 125 & 966 \\
 analcatdata\_birthday & 4 & 3 & 365 & 2 & 30 & 53 & 968 \\
 analcatdata\_draft & 5 & 3 & 366 & 2 & 1 & 32 & 984 \\
 collins & 23 & 3 & 500 & 2 & 0 & 80 & 987 \\
 prnn\_fglass & 10 & 1 & 214 & 2 & 0 & 76 & 996 \\
 jEdit\_4.2\_4.3 & 9 & 1 & 369 & 2 & 0 & 165 & 1048 \\
 mc2 & 40 & 1 & 161 & 2 & 0 & 52 & 1054 \\
 mw1 & 38 & 1 & 403 & 2 & 0 & 31 & 1071 \\
 jEdit\_4.0\_4.2 & 9 & 1 & 274 & 2 & 0 & 134 & 1073 \\
 PopularKids & 11 & 5 & 478 & 3 & 0 & 90 & 1100 \\
 teachingAssistant & 7 & 5 & 151 & 3 & 0 & 49 & 1115 \\
 lungcancer\_GSE31210 & 24 & 3 & 226 & 2 & 0 & 35 & 1412 \\
 MegaWatt1 & 38 & 1 & 253 & 2 & 0 & 27 & 1442 \\
 PizzaCutter1 & 38 & 1 & 661 & 2 & 0 & 52 & 1443 \\
 PizzaCutter3 & 38 & 1 & 1043 & 2 & 0 & 127 & 1444 \\
 CostaMadre1 & 38 & 1 & 296 & 2 & 0 & 38 & 1446 \\
 CastMetal1 & 38 & 1 & 327 & 2 & 0 & 42 & 1447 \\
 KnuggetChase3 & 40 & 1 & 194 & 2 & 0 & 36 & 1448 \\
 PieChart1 & 38 & 1 & 705 & 2 & 0 & 61 & 1451 \\
 PieChart3 & 38 & 1 & 1077 & 2 & 0 & 134 & 1453 \\
 parkinsons & 23 & 1 & 195 & 2 & 0 & 48 & 1488 \\
 planning-relax & 13 & 1 & 182 & 2 & 0 & 52 & 1490 \\
 qualitative-bankruptcy & 7 & 7 & 250 & 2 & 0 & 107 & 1495 \\
 sa-heart & 10 & 2 & 462 & 2 & 0 & 160 & 1498 \\
 seeds & 8 & 1 & 210 & 3 & 0 & 70 & 1499 \\
 thoracic-surgery & 17 & 14 & 470 & 2 & 0 & 70 & 1506 \\
 user-knowledge & 6 & 1 & 403 & 5 & 0 & 24 & 1508 \\
 wholesale-customers & 9 & 2 & 440 & 2 & 0 & 142 & 1511 \\
 heart-long-beach & 14 & 1 & 200 & 5 & 0 & 10 & 1512 \\
 robot-failures-lp5 & 91 & 1 & 164 & 5 & 0 & 21 & 1520 \\
 vertebra-column & 7 & 1 & 310 & 3 & 0 & 60 & 1523 \\
Smartphone-Based... & 68 & 2 & 180 & 6 & 0 & 30 & 4153 \\
 breast-cancer-... & 10 & 10 & 277 & 2 & 0 & 81 & 23499 \\
 LED-display-... & 8 & 1 & 500 & 10 & 0 & 37 & 40496 \\
 GAMETES\_Epistasis... & 21 & 21 & 1600 & 2 & 0 & 800 & 40646 \\
 calendarDOW & 33 & 21 & 399 & 5 & 0 & 44 & 40663 \\
 corral & 7 & 7 & 160 & 2 & 0 & 70 & 40669 \\
 mofn-3-7-10 & 11 & 11 & 1324 & 2 & 0 & 292 & 40680 \\
 thyroid-new & 6 & 1 & 215 & 3 & 0 & 30 & 40682 \\
 solar-flare & 13 & 13 & 315 & 5 & 0 & 21 & 40686 \\
 threeOf9 & 10 & 10 & 512 & 2 & 0 & 238 & 40690 \\
 xd6 & 10 & 10 & 973 & 2 & 0 & 322 & 40693 \\
 tokyo1 & 45 & 3 & 959 & 2 & 0 & 346 & 40705 \\
 parity5\_plus\_5 & 11 & 11 & 1124 & 2 & 0 & 557 & 40706 \\
 cleve & 14 & 9 & 303 & 2 & 0 & 138 & 40710 \\
 cleveland-nominal & 8 & 8 & 303 & 5 & 0 & 13 & 40711 \\
 Australian & 15 & 9 & 690 & 2 & 0 & 307 & 40981 \\
 DiabeticMellitus & 98 & 1 & 281 & 2 & 2 & 99 & 41430 \\
 conference\_attendance & 7 & 7 & 246 & 2 & 0 & 31 & 41538 \\
 CPMP-2015-... & 23 & 1 & 527 & 4 & 0 & 78 & 41919 \\
 TuningSVMs & 81 & 1 & 156 & 2 & 0 & 54 & 41976 \\
 regime\_alimentaire & 20 & 17 & 202 & 2 & 17 & 41 & 42172 \\
 iris-example & 5 & 1 & 150 & 3 & 0 & 50 & 42261 \\
 Touch2 & 11 & 1 & 265 & 8 & 0 & 27 & 42544 \\
 penguins & 7 & 3 & 344 & 3 & 18 & 68 & 42585 \\
 titanic & 8 & 5 & 891 & 2 & 689 & 342 & 42638 \\
\bottomrule
\end{tabular}
    \label{table:reference_datasets_table_2}
\end{table}

\begin{table}[bt]
    \caption{Datasets used for the evaluation in the \openmlautoml{} Benchmark. These include all datasets with at most $1\,111$ samples, $100$ features and $10$ classes.}
    \label{table:automl_small_datasets}
    \centering
    \tiny
    \begin{tabular}{lrrrrrrr}
\toprule
                                  Name &  \#Feat. &  \#Cat. &  \#Inst. &  Class Size &   \#NaNs &  Minor. Class Size &  OpenML ID \\
\midrule
                              credit-g &      21 &     14 &    1000 &           2 &       0 &                300 &         31 \\
                               vehicle &      19 &      1 &     846 &           4 &       0 &                199 &         54 \\
                                  wine &      14 &      1 &     178 &           2 &       0 &                 71 &        973 \\
      blood-transfusion-service-center &       5 &      1 &     748 &           2 &       0 &                178 &       1464 \\
                            Australian &      15 &      9 &     690 &           2 &       0 &                307 &      40981 \\
\bottomrule
\end{tabular}
\end{table}

\begin{table}[bt]
    \centering
    \caption{Evaluation datasets for model generalization experiments.}
    \tiny
    \begin{tabular}{lrrrrrrr}
\toprule
 Name & \#Feat. &  \#Cat. &  \#Inst. &  \#Class. &  \#NaNs &  Minor. Class Size & OpenML ID \\
\midrule
                    KDDCup09\_appetency &         231 &                      39 &        50000 &          2 & 8024152 &                  890 &  1111 \\
                              airlines &           8 &                       5 &       539383 &          2 &       0 &               240264 &  1169 \\
                        bank-marketing &          17 &                      10 &        45211 &          2 &       0 &                 5289 &  1461 \\
                                 nomao &         119 &                      30 &        34465 &          2 &       0 &                 9844 &  1486 \\
                                 adult &          15 &                       9 &        48842 &          2 &    6465 &                11687 &  1590 \\
                             covertype &          55 &                      45 &       581012 &          7 &       0 &                 2747 &  1596 \\
                           numerai28.6 &          22 &                       1 &        96320 &          2 &       0 &                47662 & 23517 \\
                             connect-4 &          43 &                      43 &        67557 &          3 &       0 &                 6449 & 40668 \\
jungle\_chess\_2pcs\... &           7 &                       1 &        44819 &          3 &       0 &                 4335 & 41027 \\
                            APSFailure &         171 &                       1 &        76000 &          2 & 1078695 &                 1375 & 41138 \\
                                albert &          79 &                      53 &       425240 &          2 & 2734000 &               212620 & 41147 \\
                             MiniBooNE &          51 &                       1 &       130064 &          2 &       0 &                36499 & 41150 \\
                             guillermo &        4297 &                       1 &        20000 &          2 &       0 &                 8003 & 41159 \\
                              riccardo &        4297 &                       1 &        20000 &          2 &       0 &                 5000 & 41161 \\
                               volkert &         181 &                       1 &        58310 &         10 &       0 &                 1361 & 41166 \\
                                dionis &          61 &                       1 &       416188 &        355 &       0 &                  878 & 41167 \\
                                jannis &          55 &                       1 &        83733 &          4 &       0 &                 1687 & 41168 \\
                                helena &          28 &                       1 &        65196 &        100 &       0 &                  111 & 41169 \\
\bottomrule
\end{tabular}
    \label{table:openml_datasets_table}
\end{table}

\end{document}